%% file: msu-vsrb.tex
\documentclass[runningheads]{llncs}
\usepackage{graphicx}

\usepackage{amsmath}
\usepackage{amssymb}
\usepackage{booktabs}
\usepackage{multirow}
\usepackage{pgfplots}
\usepackage{tikz}
\usepackage{scalefnt}
\usepackage{comment}
\usepackage{amsmath,amssymb} 
\usepackage{color}
\usepackage{caption}
\usepackage{subcaption}
\usepackage{lmodern}
\pgfplotsset{compat=1.17}

\usepackage[pagebackref,breaklinks,colorlinks]{hyperref}

\usepackage[capitalize]{cleveref}
\crefname{section}{Sec.}{Secs.}
\Crefname{section}{Section}{Sections}
\Crefname{table}{Table}{Tables}
\crefname{table}{Tab.}{Tabs.}

\newcommand{\etal}{et al.}

\newsavebox{\bigimage}

\usepackage[accsupp]{axessibility}  

\usepackage[width=122mm,left=12mm,paperwidth=146mm,height=193mm,top=12mm,paperheight=217mm]{geometry}

\begin{document}
\pagestyle{headings}
\mainmatter
\def\ECCVSubNumber{5728}  

\title{Towards True Detail Restoration for Super-Resolution: \\ A Benchmark and a Quality Metric}

\titlerunning{Towards True Detail Restoration for Super-Resolution}
%
\author{Eugene Lyapustin \and
Anastasia Kirillova \and
Viacheslav Meshchaninov \and \\
Evgeney Zimin \and
Nikolai Karetin \and
Dmitriy Vatolin
}
%
\authorrunning{E. Lyapustin et al.}
%
\institute{Lomonosov Moscow State University, Moscow, Russia
\email{\{evgeny.lyapustin,anastasia.kirillova,vyacheslav.meshchaninov,\\evgeney.zimin,nikolai.karetin,dmitriy\}@graphics.cs.msu.ru}}
\maketitle

\begin{abstract}
    Super-resolution (SR) has become a widely researched topic in recent years. SR methods can improve overall image and video quality and create new possibilities for further content analysis. But the SR mainstream focuses primarily on increasing the naturalness of the resulting image despite potentially losing context accuracy. Such methods may produce an incorrect digit, character, face, or other structural object even though they otherwise yield good visual quality. Incorrect detail restoration can cause errors when detecting and identifying objects both manually and automatically.
    
    To analyze the detail-restoration capabilities of image and video SR models, we developed a benchmark based on our own video dataset, which contains complex patterns that SR models generally fail to correctly restore. We assessed 32 recent SR models using our benchmark and compared their ability to preserve scene context. We also conducted a crowd-sourced comparison of restored details and developed an objective assessment metric that outperforms other quality metrics by correlation with subjective scores for this task. In conclusion, we provide a deep analysis of benchmark results that yields insights for future SR-based work.
\end{abstract}

\section{Introduction}
\label{sec:intro}
\begin{figure}
    \begin{center}
  \sbox{\bigimage}{%
    \input{img/runtime_to_quality.pgf}%
  }%
  \usebox{\bigimage}\hfill
  \begin{minipage}[b][\ht\bigimage][c]{0.5\linewidth}
      \begin{subfigure}{0.2\linewidth}
      \includegraphics[width=\linewidth]{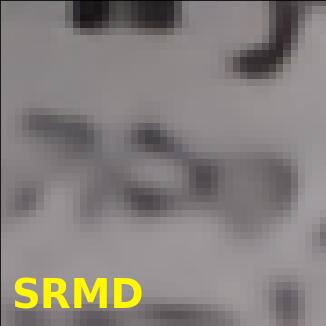}
  \label{fig:crops-a}
      \end{subfigure}
      \begin{subfigure}{0.2\linewidth}
      \includegraphics[width=\linewidth]{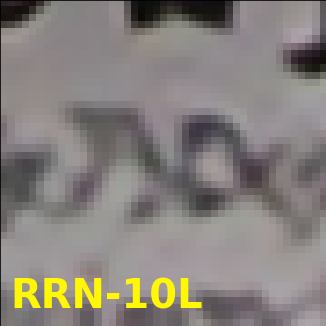}
  \label{fig:crops-b}
      \end{subfigure}
      \begin{subfigure}{0.2\linewidth}
      \includegraphics[width=\linewidth]{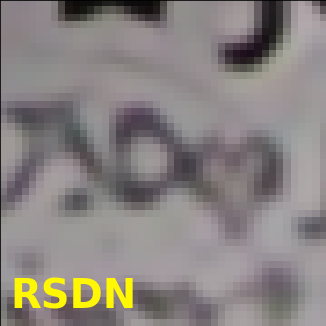}
  \label{fig:crops-c}
      \end{subfigure}
      \begin{subfigure}{0.2\linewidth}
      \includegraphics[width=\linewidth]{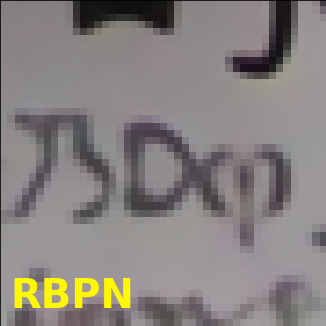}
  \label{fig:crops-d}
      \end{subfigure}
      \begin{subfigure}{0.2\linewidth}
      \includegraphics[width=\linewidth]{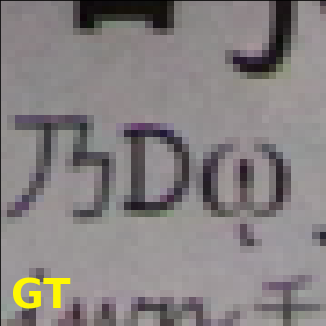}
  \label{fig:crops-e}
      \end{subfigure}
      \begin{subfigure}{0.2\linewidth}
      \includegraphics[width=\linewidth]{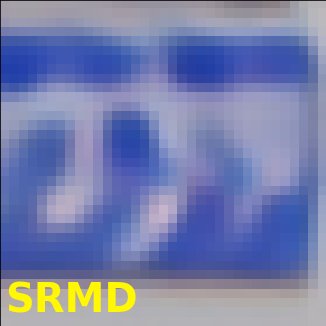}
  \label{fig:crops2-a}
      \end{subfigure}
      \begin{subfigure}{0.2\linewidth}
      \includegraphics[width=\linewidth]{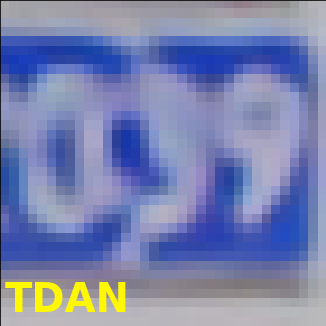}
  \label{fig:crops2-b}
      \end{subfigure}
      \begin{subfigure}{0.2\linewidth}
      \includegraphics[width=\linewidth]{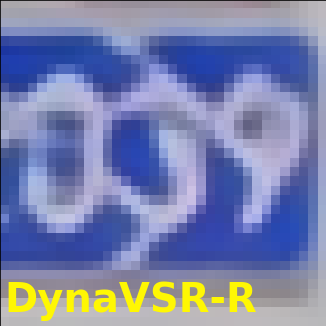}
  \label{fig:crops2-c}
      \end{subfigure}
      \begin{subfigure}{0.2\linewidth}
      \includegraphics[width=\linewidth]{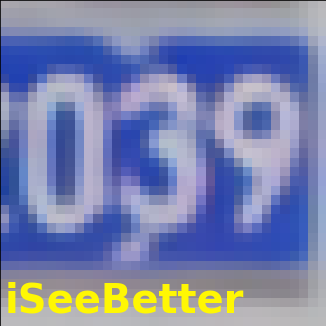}
  \label{fig:crops2-d}
      \end{subfigure}
      \begin{subfigure}{0.2\linewidth}
      \includegraphics[width=\linewidth]{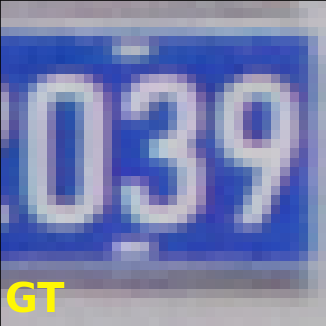}
  \label{fig:crops2-e}
      \end{subfigure}
      \vspace{0pt}%
      \end{minipage}
    \end{center}
    \vspace*{-2\baselineskip}
    \caption{Left: ranking of subjective score and runtime among 25 Super-Resolution models. Every model was tested on the same video sequence of 100 frames of $480\times320$ resolution. Subjective score is calculated using the Bradly-Terry model. Dotted line shows Pareto front. Right: Visual comparison between the ground truth (GT) images and the results of super-resolution models.}
\end{figure}

Super-resolution (SR) involves increasing the spatial resolution of images and videos. Potential uses include image and video enhancement, restoration, and compression \cite{10.1145/3458306.3458874} through increases in perceptual quality and through more-accurate identification and recognition of objects that are unclear in the original (lower-resolution) image. The topic is under considerable research, with new works appearing monthly.

Although image-based methods, by design, lack additional information when attempting to super-resolve images on the basis of spatial information and learned distributions, video- and burst-based methods employ a sequence of images, allowing them to restore details using temporal information. Because digital images and videos are discrete, the scene is undersampled, but neighboring frames can help fill in the gap when upscaling. Even small movements caused by camera tremor may yield enough information to accurately increase the resolution 2--3 times, as demonstrated in a Google Pixel 3 camera \cite{10.1145/3306346.3323024}.

\begin{figure}[t]
  \centering
  \begin{subfigure}{0.2\linewidth}
  \includegraphics[width=\linewidth]{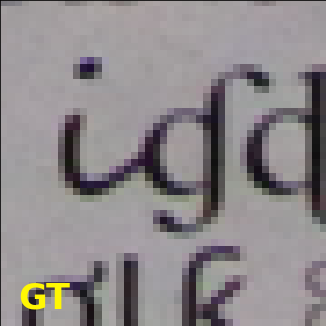}
  \end{subfigure} \hspace*{-0.3em}
  \begin{subfigure}{0.2\linewidth}
  \includegraphics[width=\linewidth]{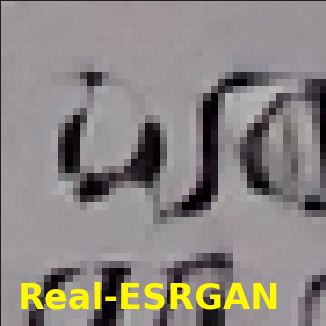}
  \end{subfigure} \hspace*{-0.3em}
  \begin{subfigure}{0.2\linewidth}
  \includegraphics[width=\linewidth]{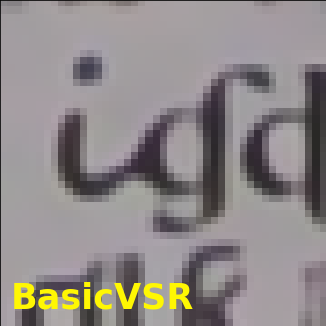}
  \end{subfigure}
  \caption{Example  of characters changing after video upscaling: two characters from source frame (GT, leftmost) merge into a single one for one method (Real-ESRGAN, central) but not for the other (BasicSR, rightmost).}
  \label{fig:example_character}
\end{figure}

Recent advances in SR take advantage of deep learning \cite{haris2019recurrent,chan2021basicvsr,liang2022vrt}, and new works regularly set higher quality standards for existing datasets \cite{Son_2021_CVPR}.
Most of these works target subjective beauty in the resulting image, which is often fine when an appealing image is helpful or necessary. But no such works to our knowledge have analyzed how those models will restore actual details and context from the original scene. For example, recent GAN methods likely produce results similar to the distribution of their training dataset rather than restoring reference details~\cite{pipal}. But preserving context is necessary for video surveillance, dashboard cameras, and other tasks that require scene interpretation. An incorrectly restored digit or character (\cref{fig:example_character}) can lead to incorrect conclusions when analyzing images or videos. Likewise, an improperly reproduced face will cause misidentification during both manual and automatic classification.
But, as far as we know, there are no existing benchmarks for SR detail restoration.

Analyzing detail restoration requires objective quality metrics that can automatically detect misrepresented details compared with ground truth and that can validate SR models. The most common image-quality metrics, PSNR and SSIM~\cite{wang2004image}, are unable to detect mistaken details.
Prior works reveal that both PSNR and SSIM, on which existing benchmarks rely heavily, are unsuitable for assessing SR methods~\cite{pipal,ding2020iqa}.
An objective-quality metric for detail restoration should consider how well object structure is restored relative to ground-truth imagery. 
Because some works use edge maps to prevent distortion of super-resolved faces~\cite{KIM202111}, we argue that a similar method will help in assessing SR models and find promising SR approaches.

With detail restoration in mind, our main contributions are the following:
\begin{enumerate}
    \item We developed a comprehensive benchmark\footnote{\url{https://videoprocessing.ai/vsr/}} for restoring details when performing super-resolution. It includes objective and subjective assessments.
    \item We gathered and analyzed 32 SR models and provided a deep analysis as well as insights for future work in this area.
    \item We developed a new quality metric, ERQAv2.0\footnote{\url{https://github.com/msu-video-group/erqa}}, to assess detail restoration; it outperforms other quality metrics by correlation with subjective scores for SR assessment.
\end{enumerate}

\section{Related work}
\label{sec:related}

\paragraph{Super-resolution methods.}
The SR problem gets a lot of attention recently with new different approaches pushing state-of-the-art visual quality and performance.

Earlier deep learning methods like ESPCN~\cite{shi2016real} extracted feature maps in the low-resolution space and used a sub-pixel convolution layer which learned an array of upscaling filters to produce high-resolution output.
Several approaches were proposed that predict degradation kernels for super-resolution~\cite{ji2020real,lee2021dynavsr}.

Crucial part for video super-resolution is usage of temporal information. DBVSR~\cite{pan2021deep}, for example, estimates a motion blur for the particular video and compensates the motion between frames explicitly. SOF-VSR~\cite{Wang2018accv} explicitly compensates motion by high-resolution optical flow, estimated from the low-resolution video in a coarse-to-fine manner.
Several works~\cite{jo2018deep,ying2020deformable,tian2020tdan} use a deformable 3D convolution to compensate the motion between frames.
LGFN~\cite{su2020local} extends deformable convolution approach with decreased multi-dilation convolution units (DMDCUs) for  explicit frames alignment.
Another approach to dealing with temporal information is using recurrent architectures.
Harris~\etal~\cite{haris2019recurrent} treated each context frame as a separate source of information. These sources are combined in an iterative refinement framework which is aided with explicit inter-frame motion estimation.
Isobe~\etal~\cite{isobe2020video} divided the input into structure and detail components which are fed to a recurrent unit composed of several proposed two-stream structure-detail blocks.

Generative adversarial networks (GAN) became very popular in deep learning and particularly in super-resolution.
Wang~\etal~\cite{wang2018esrgan} has improved network architecture, adversarial loss and perceptual loss of SRGAN~\cite{Ledig_2017_CVPR}. They use RRDB without batch normalization as the basic network building unit and let the discriminator predict relative realness instead of the absolute value. They also use loss features before activation.
Later it was extended~\cite{wang2021realesrgan} by a high-order degradation modeling process to better simulate complex real-world degradations.
iSeeBetter \cite{chadha2020iseebetter} combines GAN and recurrent approaches. It extracts spatial and temporal information from the current and neighboring frames. Recurrent back-projection network is used as generator and the discriminator is the same as for SRGAN~\cite{Ledig_2017_CVPR}.

Some works propose other unique approaches.
TGA~\cite{isobe2020video1} reorganizes input sequence into several groups of subsequences with different frame rates.  Groups allow to extract spatio-temporal information in a hierarchical manner. They are followed by an intra-group and inter-group fusion modules. TMNet~\cite{xu2021temporal} is trained for space-time video super-resolution. The temporal information is integrated by deformable convolution with the multi-frame input.

\paragraph{Super-resolution quality metrics.}
Various video quality metrics are used to measure the performance of super-resolution methods. 

PSNR is a traditional quality metric. It uses mean square error and maximum of pixel values to calculate peak to noise ratio. SSIM~\cite{wang2004image} is another traditional quality metric which calculates average, variance and covariance pixel values on windows of an image. Despite the poor correlation of these metrics with subjective score, they are still commonly used in super-resolution papers.

Recently, deep learning approaches to video quality metrics are gaining popularity. LPIPS~\cite{zhang2018perceptual} is a metric which utilizes deep features of various neural networks (for example, VGG trained on ImageNet) for image comparison. It performs well for its general task and now it is gaining popularity among researchers. Another example is DISTS~\cite{ding2020iqa}, which uses a convolutional neural network to extract texture representations from images and then combines spatial features of these texture representations with the traditional structure similarity.

Some metrics take more unique approaches. SFSN~\cite{zhou2021image} considers super-resolved images in a two-dimensional space of structural fidelity versus statistical naturalness to account for the behavior of specific super-resolution approaches, such as generative adversarial networks. ERQA~\cite{kirillova2021erqa} is a metric which validates super-resolved images in terms of real detail restoration. In order to achieve this, object edges are detected to be matched with their counterparts on the reference image as edge fidelity is crucial for truthful restoration.

While most super-resolution methods are evaluated by full-reference metrics, they require ground-truth images, which are not always available in practice. NeuralSBS \cite{Khrulkov_2021_CVPR} is a no-reference metric designed specifically for the super-resolution task. This metric uses Siamese neural network trained on aligned image pairs. Labels for the training are pick rates for each image in a pair. After training model is able to predict which one of two super-resolved images will be preferred by humans.

\paragraph{Super-resolution benchmarks.}
A few super-resolution benchmarks were organized by companies and conferences with the purposes to compare diverse algorithms and to find the state-of-the-art for the task.

Vimeo 90k \cite{xue2019video} is a large high-quality video dataset for low-level video processing. Authors use this dataset to evaluate their TOFlow algorithm in three video processing tasks: frame interpolation, video denoising and super-resolution. PSNR and SSIM are used for evaluation.

NTIRE 2021 Challenge on Video Super-resolution \cite{Son_2021_CVPR} presents evaluation results of quality restoration competition on full (track 1) and half (track 2) framerate, 247 and 223 participants have registered, respectively. This challenge uses REDS \cite{Nah_2019_CVPR_Workshops} dataset with 30000 images divided in 100-frame sequences. PSNR and SSIM are used to score participants with additional LPIPS values available.

RealVSR \cite{Yang_2021_ICCV} dataset consists of 500 paired LR-HR videos captured using the multi-camera system of iPhone 11 Pro Max. To combat misalignment and color differences between the images caused by using two separate cameras, Laplacian pyramid with different loss functions is used. PSNR, SSIM, NIQE~\cite{6353522} and BRISQUE \cite{6272356} are used for evaluation.
    
\paragraph{Super-resolution assessment.}
There is a number of datasets which contain super-resolved images aligned with some subjective scores.

QADS \cite{8640853} includes 20 reference images and 980 super-resolved images by 21 image SR methods. Mean opinion score for these images were then collected from 100 participants. 

SBS180K \cite{Khrulkov_2021_CVPR} consists of aligned pairs of images with human pick rates of each image in a pair. Images are evenly extracted from large video database. It was used for training NeuralSBS \cite{Khrulkov_2021_CVPR} metric.
  
PIPAL \cite{pipal} dataset consists of 250 image patches from two high-quality image datasets DIV2K \cite{Agustsson_2017_CVPR_Workshops} and Flickr2K \cite{timofte2017ntire}. Selected areas are hard to restore due to the presence of high-frequency textures. More than 1.13 million human judgments are collected to assign subjective scores for PIPAL images. PIPAL dataset includes the results of GAN-based methods, which are missing in previous datasets.
\section{Benchmark}
\label{sec:benchmark}

In this section, we describe our dataset and the benchmark.

\begin{figure}
\centering
\begin{minipage}{0.4\textwidth}
  \centering
  \vspace*{1.5\baselineskip}
   \includegraphics[width=0.9\linewidth]{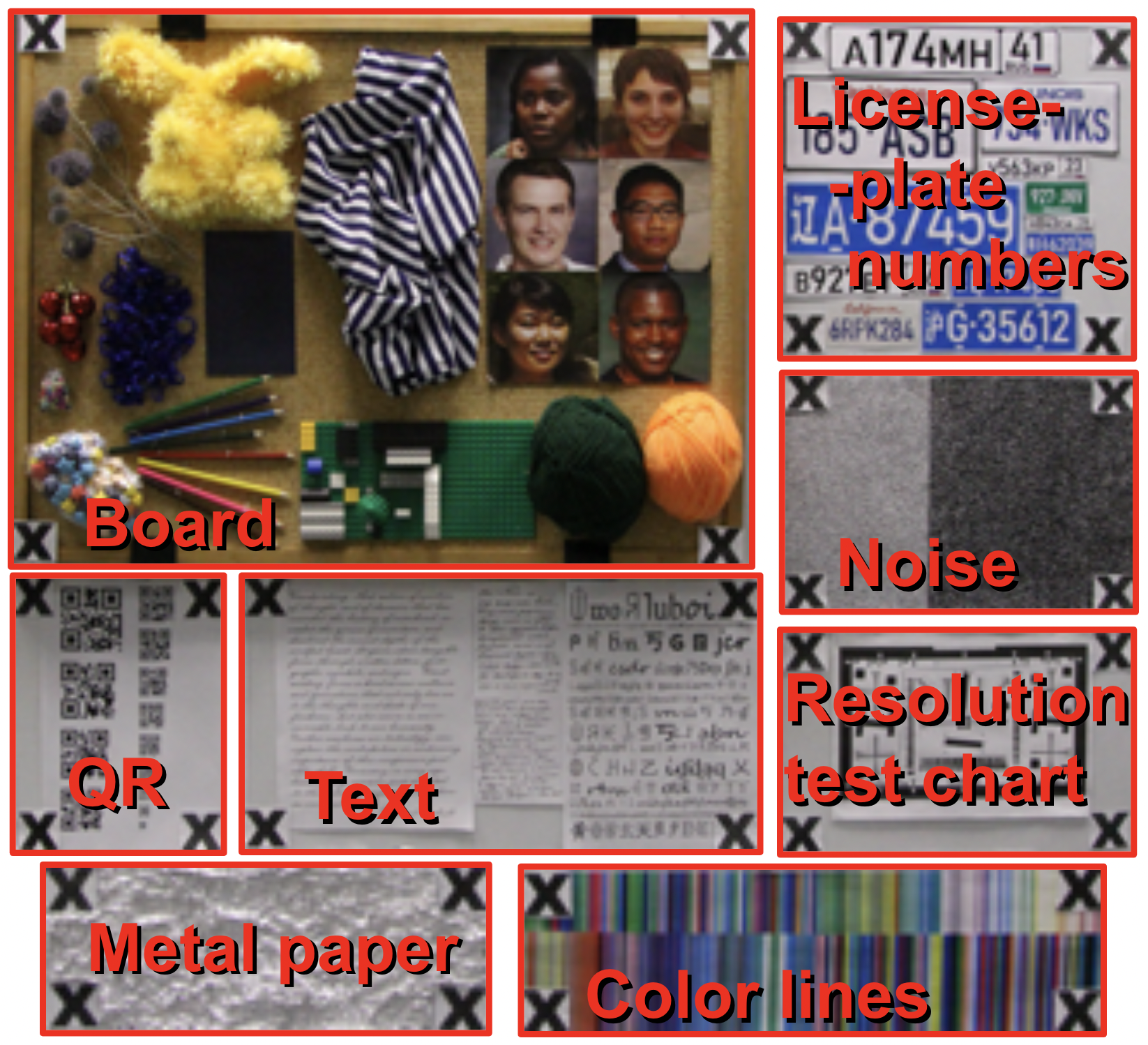}
   \caption{Parts of the dataset scene for the evaluation of different restoration aspects.}
   \label{fig:scene}
\end{minipage}\hfill
\begin{minipage}{0.55\textwidth}
    \centering
    \includegraphics[width=0.95\linewidth]{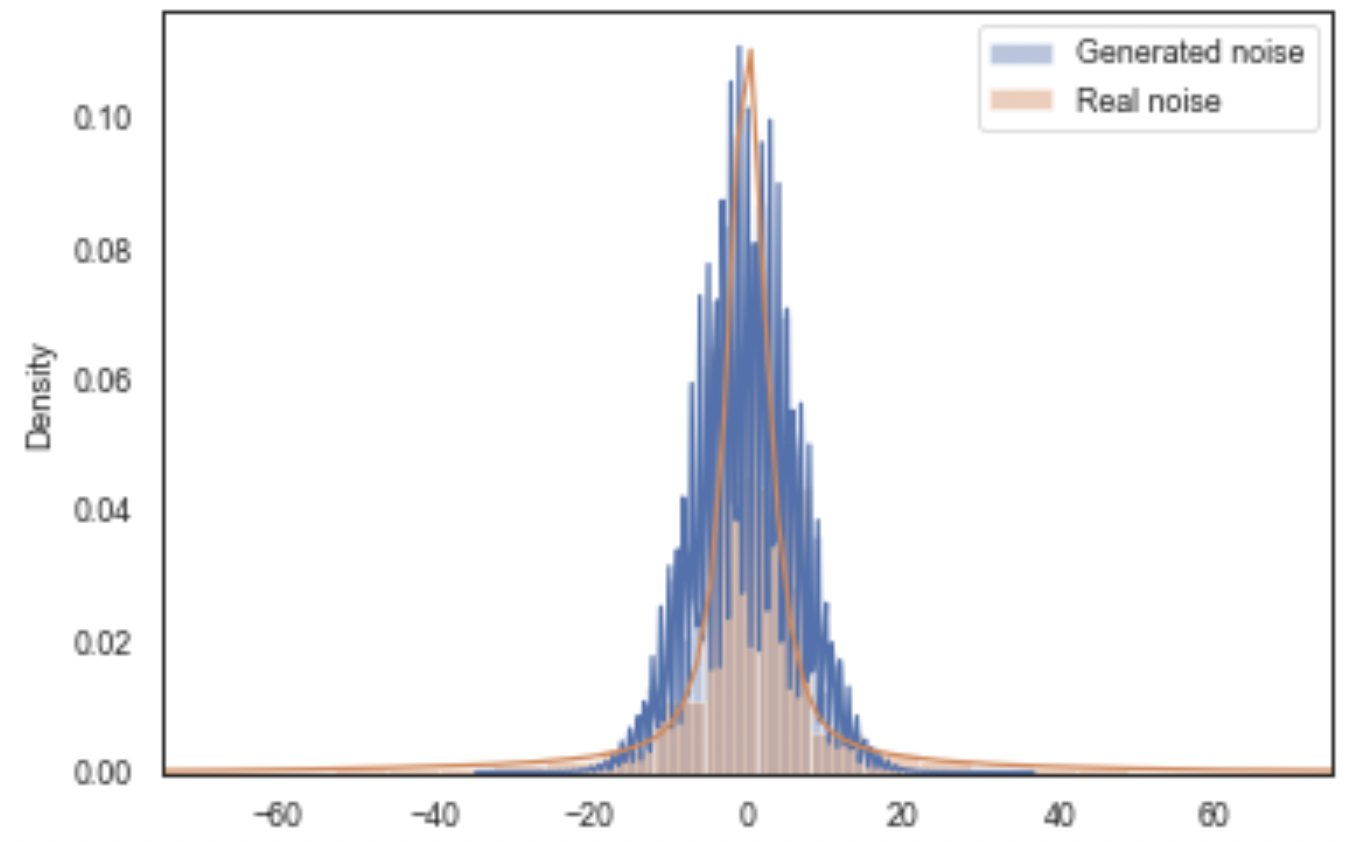}
    \caption{Comparison of real and generated noise distribution.}
    \label{fig:noise_distribution}
\end{minipage}
\end{figure}

\subsection{Scene}
\label{subsec:scene}

The scene for our dataset consists of eight patterns separated by X marks, which we used later for processing.
Each pattern aims to challenge SR models and, preferably, represent something from everyday life.
Scene can be seen in \cref{fig:scene}.
Here are descriptions for every pattern:
\begin{itemize}
    \item \textbf{Board} is a set of several objects placed on a cork board.
  Objects present volumetric and textural complexity for SR models.
  Moreover, there are 6 face images generated using StyleGAN2~\cite{Karras_2020_CVPR} for face restoration assessment.
    \item \textbf{License-plate numbers} include several car plates from different regions: USA, China and Russia.
  It is becoming common to use deep learning for investigations where it is crucial for SR methods to work reliably or explicitly limited.
    \item \textbf{QR} codes can be used to approximate information loss after consecutive image degradation and restoration.
  There are several QR codes in scene with varying size and similar coded content complexity.
    \item \textbf{Text} is common thing to see in images and videos.
  Incorrect text restoration may cause corruption of sentences and overall scene context.
  There are multiple characters of different fonts and a set of handwritten text. 
    \item \textbf{Noise} pattern provides a unique ``footprint'' for each SR method and can be used to detect overfitting.
    \item \textbf{Resolution test chart} is a commonly used for camera calibration and analysis. The chart used in a scene is compliant with ISO 12233 standard.
    \item \textbf{Metal paper} is a crumpled foil which provide unique light reflections for each video frame that complicates overall restoration.
    \item \textbf{Color lines} add complexity to restoration because of high-frequency color changes.
\end{itemize}

\subsection{Dataset preparation}
\label{subsec:tech}

\paragraph{Capture.}
We shot the videos using a Canon EOS 7D camera in daylight conditions to ensure naturalness. The camera settings aimed to minimize blur and obtain the appropriate brightness: the ISO was 4000 and the aperture 400.
Each video is a sequence of 100 photos taken in burst mode with a maximum resolution of $5184\times3456$ because the camera video resolution and compression quality are limited. We refer to these videos as \textit{source videos}.
\paragraph{High- and low-resolution frames.}
To prevent undesirable motion blur and noise in the ground-truth frames, and to avoid running afoul of memory restrictions when applying models to high-resolution images, we downscaled the source videos to $1920\times1080$ using bicubic interpolation. We refer to these downscaled videos as the \textit{ground-truth} or \textit{high-resolution (HR)} data. The \textit{low-resolution (LR)} frames are the HR frames further downscaled by $4\times$ to $480\times270$.
\paragraph{Realistic noise.}
Because real-world imagery generally contains noise, we generated additional LR videos using noise from the source  videos. We use a noise model from CBDNet~\cite{Guo_2019_CVPR}. To make it natural, we conducted an experiment to measure the camera’s noise parameters. We recorded 100 frames using the Canon EOS 7D on a tripod, calculated the average frame, and subtracted it from the actual frames to obtain noise samples. Our next step was to tune the noise-model parameters to match the real noise distribution in the camera. The results were  $\sigma_s = 0.001$ and $\sigma_c = 0.035$ (see \cref{fig:noise_distribution}).

\subsection{Subjective comparison}
\label{subsec:subjective}

To calculate accurate ranks for detail restoration we conduct crowd-source subjective comparison on \url{Subjectify.us} service. From each video we extract a few video-crops with the most difficult patterns for detail-restoration. For comparison we randomly show each participant a pair of frame patches from two super-resolution models and ask them to choose the patch on which details are restored better, contrary to the most beautiful one. Each pair of patches was shown to 10--15 participants until confidence interval stopped changing. We decided to work with patches, because it is very difficult to visually compare small details in a full frame. Each participant compared 25 pairs total with 3 of them for verification to filter out untruthful participants. 1400 participants total participated in comparison and their answer were used for final subjective scores calculation with Bradley-Terry model~\cite{bradley1952rank}.


\section{Quality Metric}
\label{sec:metric}


Inspired by ERQA metric~\cite{kirillova2021erqa}, we propose sequential refinement of restored boundaries ERQAv2.0, following the naming from original authors (ERQAv1.0 and ERQAv1.1).
Unlike ERQAv1.1, however, our approach is using a lower-level technique instead of an explicit edge detector.

First, we compute the gradients of the ground-truth and input frames using the simple convolutional kernels  $[-0.5, 0, 0.5]$ and $[-0.5, 0, 0.5]^T$.
Since outliers may degrade the results, we filter out gradients with magnitude greater than the 85th percentile of the frame-gradient distribution.
Experimental observations show that employing such filter, increases the algorithm's performance in both quality and speed.
Next, we consider gradients in the ground-truth and input frames matched, if the cosine of the angle between them exceeds $0.85$. 

\begin{figure}
\begin{minipage}{0.45\textwidth}
\centering
\vspace*{-1.5\baselineskip}
\includegraphics[width=1.0\linewidth]{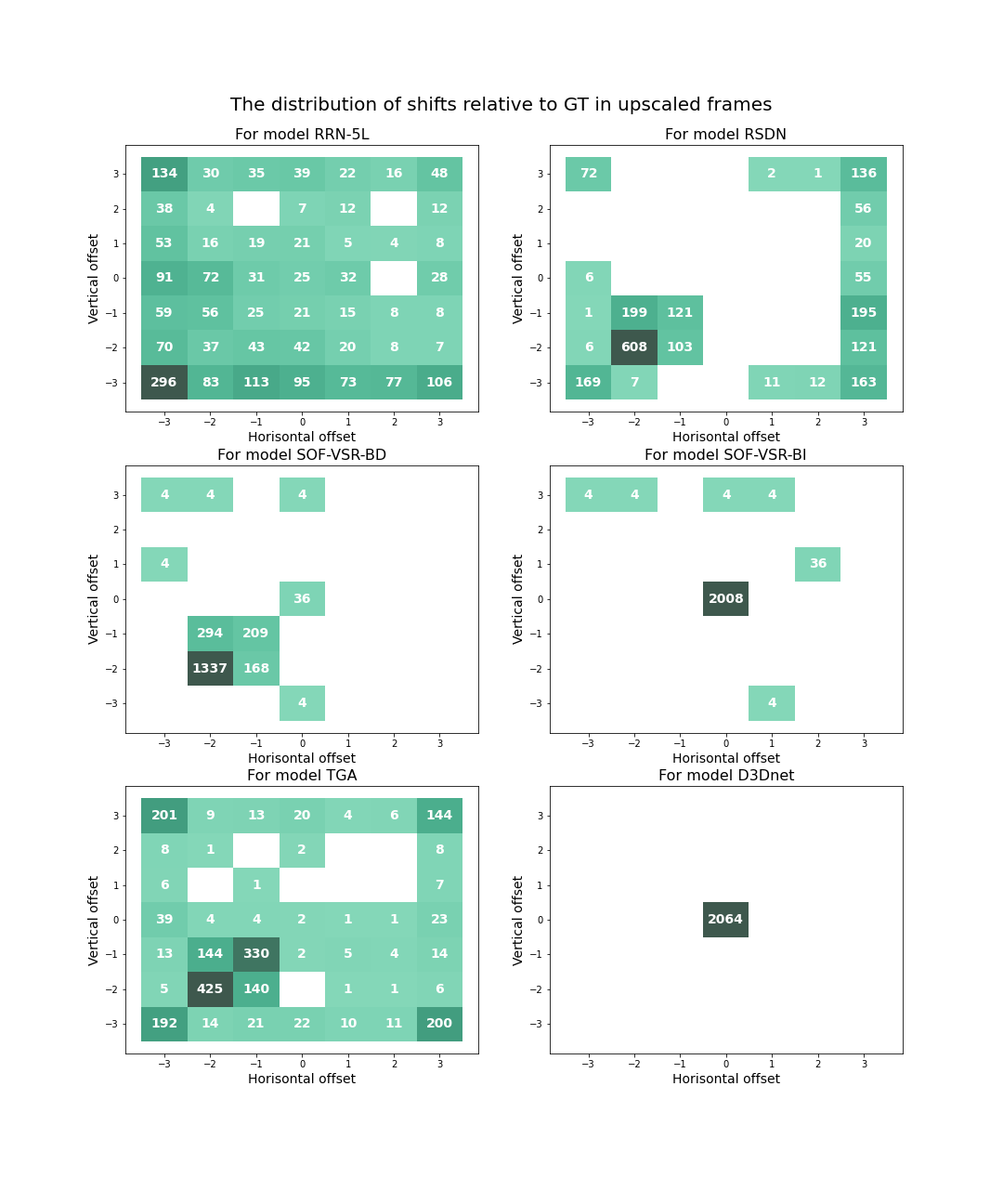}
\vspace*{-3\baselineskip}
\caption{The distribution of global pixel shifts of some methods relative to ground-truth. Some models produce non-constant shifts which can be explained by training on a different downsampling method.}
\label{fig:models_shifts_2}
\end{minipage}\hfill
\begin{minipage}{0.45\textwidth}
    \begin{center}
    \vspace*{-3\baselineskip}
    \scalebox{0.65}{\input{metric2.pgf}}
    \end{center}
    \caption{Heat map of Super-Resolution artifacts showcasing the difference in text restoration.}
    \label{fig:erqa-heatmap}
\end{minipage}
\end{figure}

As in Kirillova~\etal~\cite{kirillova2021erqa} was shown, some SR methods can generate frames with small pixel shifts relative to ground-truth (see \cref{fig:models_shifts_2}). Authors of original ERQA addressed the issue by grid-searching pixel shifts to find one that maximizes PSNR.
Our experiments shown that pixel shifts may vary across frame so we developed an iterative approach. ERQAv2.0 first finds measure of similarity --- the number of matching gradients between the input and ground truth frames with 5-pixel radius.
Next we sort the shifts in descending order by this measure of similarity.
We then iteratively take the first $N$ shifts, find matches between gradient maps, add these matches to the true positive mask and remove them from the gradient maps of the input and ground-truth frames. 
We found that the true positive mask stops changing after $N = 35$. 

Our approach yields three gradient maps: true positive, false positive and false negative.
False positive and false negative masks are the rest of the gradient maps of the input and ground-truth frames accordingly.
Then we compute the $F_\beta$-score with $\beta=0.5$.
Super-resolution methods tend to generate wide edges which means high false positive rate, so we use $\beta=0.5$.
Experimental results are presented in \cref{sec:experiments}. 

The metric can automatically assess detail-restoration quality and, on the basis of an error map, find frame areas that contain artifacts produced by SR, as shown in \cref{fig:erqa-heatmap}.

\section{Experiments}
\label{sec:experiments}

In this section we show evaluation results of SR methods and quality metrics on our dataset.

\begin{table}
  \centering
  \small
  \begin{tabular}{ c|l|ccccc|c} 
    \toprule
    Rank & Model &  Subjective $\uparrow$ & ERQAv2.0 $\uparrow$ & PSNR $\uparrow$ & SSIM $\uparrow$ & LPIPS $\downarrow$ & FPS $\uparrow $ \\ 
    \midrule
    1 & VRT~\cite{liang2022vrt}  & \textbf{7.627} & \textbf{0.851}  & \textbf{31.669} & \textbf{0.902} & \textbf{0.241} & 0.166\\
 2 & BasicVSR~\cite{chan2021basicvsr} &  \textbf{7.186} & \textbf{0.846} & \textbf{31.443} & \textbf{0.900} & \textbf{0.240} & \textbf{2.128}\\
   3 & RBPN~\cite{haris2019recurrent} & 7.068 & 0.841 & 31.407 & 0.899 & 0.260 & 0.043\\
4 & DBVSR~\cite{pan2021deep} & 6.947 & 0.835 & 31.071 & 0.894 & 0.274 & 0.241\\
5 & iSeeBetter~\cite{chadha2020iseebetter} & 6.809 & 0.839 & 31.104 & 0.896 & 0.259 & 0.044\\
6 & LGFN~\cite{su2020local} & 6.505 & 0.831 & 31.291 & 0.898 & 0.275 & 0.667\\
7 & DynaVSR-R~\cite{lee2021dynavsr} & 6.135 & 0.802 & 28.37 & 0.865 & 0.274 & 0.177\\
8 & TMNet~\cite{xu2021temporal} & 6.000 & 0.821 & 30.364 & 0.885 & 0.270 & 1.136\\
9 & COMISR~\cite{li2021comisr}  & 5.636 & 0.794 & 26.708 & 0.840 & 0.271 & 1.613 \\
10 & RSDN~\cite{isobe2020video} & 5.565 & 0.764 & 25.321 & 0.826 & 0.333 & \textbf{1.961}\\
    \bottomrule
  \end{tabular}
  \vspace*{0.5\baselineskip}
  \caption{A comparison of top-10 Super-Resolution methods by subjective score and objective metrics.}
  \label{tab:vsr_results}
\end{table}

\subsection{Super-resolution models}

For our experiments we chose models so that various approaches were included in our benchmark and compared with each other: alignment by optical flow \cite{wang2020deep,pan2021deep} and 3D convolutions \cite{jo2018deep,ying2020deformable}, deformable convolutions \cite{tian2020tdan,su2020local}, recurrent models \cite{isobe2020revisiting,haris2019recurrent,isobe2020video}, meta-learning \cite{lee2021dynavsr}, and generative adversarial networks \cite{chadha2020iseebetter,wang2021realesrgan}. We mostly used the latest methods, because they generally show better results.

We evaluated 32 different Super-resolution models  by applying them to our dataset on a server with Intel(R) Xeon(R) Silver 4216 CPU and NVIDIA Titan RTX GPU. The execution time is calculated as a full runtime divided by the number of frames in a sequence with $480\times320$ resolution. FPS is an inverse value to the execution time. You can see top-10 leaderboard in \cref{tab:vsr_results} and full leaderboard in supplementary materials.

\subsection{Quality metrics}

On top of super-resolution results on our dataset we calculated several quality metrics that are usually used for super-resolution assessment. You can see metrics correlation with subjective scores in \cref{tab:metrics_correlation}.

\begin{figure}
\centering
\begin{minipage}{0.4\textwidth}
  \centering
    \vspace*{-2.3\baselineskip}
    \begin{tabular}{ c|c|c}
    \toprule
    Metric & PLCC & SRCC \\
    \midrule
    ERQAv2.0 & \textbf{0.899} & \textbf{0.805} \\
    ERQAv1.1 & \textbf{0.878} & \textbf{0.787} \\
    LPIPS & 0.835 & 0.738 \\
    DISTS & 0.828 & 0.719  \\
    SSIM &  0.670 & 0.575 \\
    MS-SSIM & 0.599 & 0.548 \\
    SFSN & 0.474 & 0.512 \\
    PSNR & 0.431 & 0.471 \\
    VMAF & 0.264 & 0.459 \\
    NeuralSBS & 0.066 & 0.029 \\
    \bottomrule
    \end{tabular}
     \vspace*{2.3\baselineskip}
    \captionof{table}{Mean Pearson (PLCC) and Spearman (SRCC) correlation coefficients of used metrics with subjective comparison.}
    \label{tab:metrics_correlation}
\end{minipage}\hfill
\begin{minipage}{0.55\textwidth}
    \centering
    \scalebox{0.85}{\input{img/clustering.pgf}}
    \caption{Clustering of algorithms by a few objective metrics (ERQAv2.0, LPIPS, PSNR, and SSIM). Centers of clusters were chosen as representative methods for the analysis of subjective comparison. The mean values of ERQAv2.0 and LPIPS are chozen for axes on this plot.}
    \label{fig:representative_algorithms}
\end{minipage}
\end{figure}

\subsection{Analysis of subjective-comparison method}
\label{subsec:subjective-exp}

As subjective assessment is expensive and time consuming with a number of comparisons grows quadratic to the number of compared methods. Therefore we sampled a small number of SR methods to analyze several hypotheses. Conducted a subjective evaluation of all models on tests with hand tremor motion and bicubic degradation (\cref{tab:vsr_results}), we were able to choose a few metrics that have high correlation with subjective assessment (\cref{tab:metrics_correlation}). Then we used two traditional metrics (PSNR and SSIM) and two metrics more consistent with subjective assessment (ERQAv1.1~\cite{kirillova2021erqa} and LPIPS \cite{zhang2018perceptual}). We also selected three patches from the scene for this experiment so scores between every patch are distinct. We used 12 features (4 different metrics for 3 patches) to group 21 SR models using KMedoids with 6 clusters. We consider the resulting six group centers to be representative methods for later subjective experiments. (\cref{fig:representative_algorithms}).


Finally, we conducted four subjective comparisons of representative methods under different test settings:
\begin{enumerate}
\setlength{\itemsep}{0pt}
\item Hand tremor and Bicubic Interpolation (BI)
\item Parallel motion and BI
\item Rotation and BI
\item Hand tremor and Blur Downsampling (BD) --- direct downsampling after Gaussian smoothing
\end{enumerate}

The experiment showed that subjective scores on tests with different types of motion correlate highly in both the Pearson and Spearman coefficients (\cref{tab:correlation_of_subjective}) and are greatly consistent with each other. So, it is not necessary to conduct a subjective assessment on all types of camera motion. On the contrary, it is important to collect subjective scores for models on both BI and BD degradation as the results strongly differ. For the top two models, the subjective score on a test with BD degradation differs from the score on the same test with BI degradation. Finally, we conduct subjective comparison on both BI and BD degradation.

\begin{table}
\centering
\begin{minipage}{0.4\textwidth}
  \centering
    \begin{tabular}{ l|c c } 
    \toprule
    \multirow{2}{*}{Test} & \multicolumn{2}{c}{Hand tremor + BI} \\
     & SRCC & PLCC \\
    \midrule
    Parallel motion + BI & 0.943 & 0.946	\\
    Rotation + BI & 1.000 & 0.955 \\
    Hand tremor + BD & 0.943 & 0.685 \\
    \bottomrule
  \end{tabular}
    \vspace*{1.5\baselineskip}
    \caption{Pearson (PLCC) and Spearman (SRCC) correlation coefficients between subjective scores on Hand tremor + BI test and other tests. The correlation was calculated on 6 representative algorithms.}
  \label{tab:correlation_of_subjective}
\end{minipage}\hfill
\begin{minipage}{0.55\textwidth}
    \centering
    \vspace*{-3.5\baselineskip}
    \begin{tabular}{l|cc}
    \toprule
    Metric & SRCC & PLCC \\
    \midrule
  ERQAv2.0 & 0.721 & 0.797 \\
  ERQAv1.1 & 0.548 & 0.677 \\
  PSNR & 0.868 & 0.861 \\
  SSIM & 0.877 & 0.941 \\
  NeuralSBS & 0.879 & 0.645 \\
  SFSN & 0.546 & 0.662 \\
    \bottomrule
  \end{tabular}
  \caption{Pearson (PLCC) and Spearman (SRCC) correlation coefficients of metrics between tests with BI and BD input degradation types.}
  \label{tab:metric_correlation_BIandBD}
\end{minipage}
\end{table}


\subsection{Analysis of values of objective metrics for VSR models}
\label{subsec:vsr_results}

While there is no one true way to downsample images for SR models training, the two most popular ones are Bicubic Interpolation (BI) and Blur Downsampling (BD) --- direct downsampling after Gaussian smoothing. As we considered in \cref{subsec:subjective-exp}, subjective scores differ between different degradations. ERQAv1.1 and ERQAv2.0 also exhibit low correlation between tests with BI and BD (\cref{tab:metric_correlation_BIandBD}), which shows that all models vary in detail restoration quality between BI and BD degradation. ERQA values on BI almost always exceed values of the same metric on BD, because Gaussian smoothing generally reduce the quality of edge restoration. PSNR and SSIM exhibit high correlation between BI and BD, which proves that these metrics are less sensitive to changes of the input degradation. NeuralSBS achieves a high SRCC between BI and BD, but its BI value are always greater than its BD value. On the other hand, SFSN values for BI are less than or equal to its values for BD.

We notice that models generally yield better results with degradation which they used during training which strongly suggest that SR models may overfit to to the specific downsampling method. So a model's rank can change greatly depending on the input degradation and a model produce much better result on the same degradation type it was trained on (\cref{tab:rank_on_BIandBD}), but some stable models, such as LGFN~\cite{su2020local}, provide similar results on both tests. To our knowledge, some authors use several types of degradation chosen randomly to neglect such overfitting~\cite{Zhang_2021_ICCV}.

\begin{figure}
\centering
\begin{minipage}{0.45\textwidth}
    \centering
     \begin{center}
  \scalebox{0.65}{\input{img/difference_parallel_rotation.pgf}}
    \end{center}
    \vspace*{-1\baselineskip}
    \caption{Average difference of ERQAv1.1 values, calculated on tests with parallel motion and rotation.}
    \label{fig:difference_parallel_rotation}
\end{minipage}\hfill
\begin{minipage}{0.45\textwidth}
  \begin{center}
  \scalebox{0.65}{\input{img/difference_noise_and_without.pgf}}
    \end{center}
    \vspace*{-1\baselineskip}    
    \caption{Average difference of ERQAv2.0 values calculated on tests with and without noise.}
    \label{fig:difference_noise_and_without}
\end{minipage}

\end{figure}

In \cref{subsec:subjective-exp} we showed that subjective scores on different types of motion are highly correlated. The same is true for metric values, but the more complicated the motion, the greater challenge for a given model. Thus, we considered how these values change with motion type for each model. We compared results using clean data to exclude noise effects. ERQAv2.0 values for parallel camera motion are generally higher than those for rotation. PSNR and SSIM indicate a much smaller difference between motion types, but for nearly all models, the metric values for hand tremor average slightly better than those for parallel motion, and those for parallel motion average slightly better than those for rotation. The dependence between metric values for rotation and parallel motion is more difficult because different models work better on different motion types (\cref{fig:difference_parallel_rotation}). We showed that the ranking can change depending on camera motion (\cref{tab:rank_on_different_motion}), but these changes are much smaller than they are with input-degradation type (\cref{tab:rank_on_BIandBD}). For example, iSeeBetter~\cite{chadha2020iseebetter} declines considerably when forceful motion is added, whereas other top models are less sensitive to changes in camera motion. DBVSR~\cite{pan2021deep} and ESRGAN~\cite{wang2018esrgan} appear more stable to motion because their metric values decrease, but that decrease is less than for other top models (RBPN~\cite{haris2019recurrent}, LGFN~\cite{su2020local}, and iSeeBetter~\cite{chadha2020iseebetter}).

\begin{table}
\centering
\begin{minipage}{0.49\textwidth}
\centering
    \begin{tabular}{l|cccc}
    \toprule
    Model & BI & BD & Mean & Trained \\
    \midrule
    DynaVSR-R & 10 & 1 & 1 & BD \\
    Real-ESRGAN & 9 & 2 & 2 & both \\
    LGFN & 5 & 6 & 3 & BI \\
    DBVSR & 1 & 11 & 4 & BI \\
    RBPN & 2 & 10 & 5 & BI \\
    iSeeBetter & 4 & 8 & 6 & BI \\
     HCFlow & 6 & 7 & 7 & --- \\
    RealSR & 8 & 9 & 8 & --- \\
    DynaVSR-V & 18 & 3 & 9 & BD \\
    ESRGAN & 3 & 16 & 10 & BI \\
     SwinIR & 16 & 4 & 11 & --- \\
     TMNet & 7 & 14 & 12 & BI \\
     Real-ESRnet & 21 & 5 & 13 & both \\
    \bottomrule
\end{tabular}
  \caption{Ranking of models by ERQAv2.0 on BI and BD input downsampling. }
  \label{tab:rank_on_BIandBD}
\end{minipage}\hfill
\begin{minipage}{0.49\textwidth}
\centering
    \begin{tabular}{l|ccc}
    \toprule
    Model &  w/o noise &  w/ noise &  Mean \\
    \midrule
    Real-ESRGAN &       8 &    1 &     1 \\
    RealSR &       9 &    3 &     2 \\
    SwinIR &      15 &    2 &     3 \\
    RBPN &       2 &    9 &     4 \\
    DBVSR &       1 &   12 &     5 \\
    iSeeBetter &       4 &    8 &     6 \\
    Real-ESRnet &      21 &    4 &     7 \\
    LGFN &       5 &   19 &    10 \\
    GFPGAN &      25 &    5 &    19 \\
    ESRGAN &       3 &   25 &    23 \\
    \bottomrule
    \end{tabular}
  \caption{Ranking of models by ERQAv2.0 on tests with and without noise, along with ranking by mean ERQAv2.0 on both tests. Models with a rank not worse than 5 on any test (with or without noise, or by mean ERQAv2.0) were visualized. Table is sorted by mean ERQAv2.0}
  \label{tab:rank_with_and_without_noise}
\end{minipage}
\end{table}

We also compared tests with and without noise. Generally, metric values for tests with noise are lower than those without noise (\cref{fig:difference_noise_and_without}), but all metrics correlate poorly between these types of tests (\cref{tab:metric_correlation_with_and_without_noise}). We showed that although models generally works worse with noise input data, rank of models can change substantially when noise is added to the input video (\cref{tab:rank_with_and_without_noise}), because metric values for some models decrease more significantly than for others. Real-ESRGAN~\cite{wang2021realesrgan}, D3Dnet~\cite{ying2020deformable}, and SOF-VSR-BI are more stable to noise than others, while RBPN~\cite{haris2019recurrent}, DBVSR~\cite{pan2021deep}, and ESRGAN~\cite{wang2018esrgan} are superior to other models on clean data but fail to perform well with noise.

\begin{table}
\centering
\begin{minipage}{0.65\textwidth}
  \centering
  \begin{tabular}{l|cccc}
    \toprule
    Model & Motion 1 &  Motion 2 &  Motion 3 & Mean \\
    \midrule
    RBPN &    2 &  1 &   3 &  1 \\
    DBVSR &    4 &  2 &   2 &  2 \\
    ESRGAN &    5 &  4 &   1 &  3 \\
    LGFN &    3 &  3 &   4 &  4 \\
     iSeeBetter &    1 &  5 &   6 &  5 \\
     HCFlow &    6 &  7 &   5 &  6 \\
     DynaVSR-R &    8 &  6 &   7 &  7 \\
     TMNet &    7 &  8 &   8 &  8 \\
     RealSR &   10 &  9 &   9 &  9 \\
    TDAN &    9 & 10 &  11 &   10 \\
    Real-ESRGAN &   14 & 11 &  10 &   11 \\
    \bottomrule
    \end{tabular}
  \caption{Ranking of models by ERQAv2.0 on tests with different camera motion (Motion 1 = Hand tremor, Motion 2 = Parallel motion, Motion 3 = Rotation), along with ranking by mean ERQAv2.0 on these three tests. Models with a rank not worse than 10 on any motion type were visualized. Table is sorted by mean ERQAv2.0.}
  \label{tab:rank_on_different_motion}
\end{minipage}\hfill
\begin{minipage}{0.3\textwidth}
    \begin{tabular}{l|cc}
  \toprule
    Metric & SRCC & PLCC \\
  \midrule
    ERQAv2.0 & 0.555 & 0.501 \\
    ERQAv1.1 & 0.455 & 0.511 \\
    PSNR & 0.705 & 0.541 \\
    SSIM & 0.500 & 0.518 \\
    SFSN & 0.900 & 0.760 \\
  \bottomrule
  \end{tabular}
  \caption{Pearson (PLCC) and Spearman (SRCC) correlation coefficients of metrics between tests with and without noise.}
  \label{tab:metric_correlation_with_and_without_noise}
\end{minipage}
\end{table}









\section{Conclusion}
In this paper we propose a new benchmark for super-resolution detail restoration.
We assess 32 super-resolution models both objectively and subjectively.
We propose a new quality metric ERQAv2.0 that has better correlations with subjective scores than other metrics and more closely resembles human ranking. It was improved and then used to study the impact of input degradations on the behavior of different super-resolution models.

Experiments show that modern SR models may overfit to downsampling method used during training, LGFN~\cite{su2020local} being the exception. Camera movement also affects model behavior but not as much, however performance of some models declines with presence of motion, iSeeBetter~\cite{chadha2020iseebetter} being an example. Adding noise to the image generally lowers metric values and substantially changes model rank: models with poor performance on clean images become relatively better with noise appearance, and on the contrary algorithms achieving high metric values before injection noticeably decrease in performance on noisy images. And in total video-based methods perform better for the detail restoration task than image-based method, confirming the earlier hypotheses.

We hope that our work can help to gain insights into detail restoration for Super-Resolution methods and develop this research area in the future. We will continue to update the benchmark with new promising methods as they appear.

%
%
\bibliographystyle{splncs04}
\bibliography{egbib}
\end{document}

%% file: img/runtime_to_quality.pgf
\begingroup%
\makeatletter%
\begin{pgfpicture}%
\pgfpathrectangle{\pgfpointorigin}{\pgfqpoint{2.385197in}{2.051315in}}%
\pgfusepath{use as bounding box, clip}%
\begin{pgfscope}%
\pgfsetbuttcap%
\pgfsetmiterjoin%
\definecolor{currentfill}{rgb}{1.000000,1.000000,1.000000}%
\pgfsetfillcolor{currentfill}%
\pgfsetlinewidth{0.000000pt}%
\definecolor{currentstroke}{rgb}{1.000000,1.000000,1.000000}%
\pgfsetstrokecolor{currentstroke}%
\pgfsetdash{}{0pt}%
\pgfpathmoveto{\pgfqpoint{0.000000in}{0.000000in}}%
\pgfpathlineto{\pgfqpoint{2.385197in}{0.000000in}}%
\pgfpathlineto{\pgfqpoint{2.385197in}{2.051315in}}%
\pgfpathlineto{\pgfqpoint{0.000000in}{2.051315in}}%
\pgfpathlineto{\pgfqpoint{0.000000in}{0.000000in}}%
\pgfpathclose%
\pgfusepath{fill}%
\end{pgfscope}%
\begin{pgfscope}%
\pgfsetbuttcap%
\pgfsetmiterjoin%
\definecolor{currentfill}{rgb}{1.000000,1.000000,1.000000}%
\pgfsetfillcolor{currentfill}%
\pgfsetlinewidth{0.000000pt}%
\definecolor{currentstroke}{rgb}{0.000000,0.000000,0.000000}%
\pgfsetstrokecolor{currentstroke}%
\pgfsetstrokeopacity{0.000000}%
\pgfsetdash{}{0pt}%
\pgfpathmoveto{\pgfqpoint{0.420988in}{0.487346in}}%
\pgfpathlineto{\pgfqpoint{2.203488in}{0.487346in}}%
\pgfpathlineto{\pgfqpoint{2.203488in}{1.950346in}}%
\pgfpathlineto{\pgfqpoint{0.420988in}{1.950346in}}%
\pgfpathlineto{\pgfqpoint{0.420988in}{0.487346in}}%
\pgfpathclose%
\pgfusepath{fill}%
\end{pgfscope}%
\begin{pgfscope}%
\pgfpathrectangle{\pgfqpoint{0.420988in}{0.487346in}}{\pgfqpoint{1.782500in}{1.463000in}}%
\pgfusepath{clip}%
\pgfsetbuttcap%
\pgfsetroundjoin%
\pgfsetlinewidth{1.003750pt}%
\definecolor{currentstroke}{rgb}{0.600000,0.600000,0.600000}%
\pgfsetstrokecolor{currentstroke}%
\pgfsetdash{{3.700000pt}{1.600000pt}}{0.000000pt}%
\pgfpathmoveto{\pgfqpoint{0.502011in}{0.883158in}}%
\pgfpathlineto{\pgfqpoint{0.773380in}{1.335873in}}%
\pgfpathlineto{\pgfqpoint{0.834256in}{1.777523in}}%
\pgfpathlineto{\pgfqpoint{1.667626in}{1.883846in}}%
\pgfusepath{stroke}%
\end{pgfscope}%
\begin{pgfscope}%
\pgfsetbuttcap%
\pgfsetroundjoin%
\definecolor{currentfill}{rgb}{0.000000,0.000000,0.000000}%
\pgfsetfillcolor{currentfill}%
\pgfsetlinewidth{0.803000pt}%
\definecolor{currentstroke}{rgb}{0.000000,0.000000,0.000000}%
\pgfsetstrokecolor{currentstroke}%
\pgfsetdash{}{0pt}%
\pgfsys@defobject{currentmarker}{\pgfqpoint{0.000000in}{-0.048611in}}{\pgfqpoint{0.000000in}{0.000000in}}{%
\pgfpathmoveto{\pgfqpoint{0.000000in}{0.000000in}}%
\pgfpathlineto{\pgfqpoint{0.000000in}{-0.048611in}}%
\pgfusepath{stroke,fill}%
}%
\begin{pgfscope}%
\pgfsys@transformshift{1.080931in}{0.487346in}%
\pgfsys@useobject{currentmarker}{}%
\end{pgfscope}%
\end{pgfscope}%
\begin{pgfscope}%
\definecolor{textcolor}{rgb}{0.000000,0.000000,0.000000}%
\pgfsetstrokecolor{textcolor}%
\pgfsetfillcolor{textcolor}%
\pgftext[x=1.080931in,y=0.390123in,,top]{\color{textcolor}\rmfamily\fontsize{10.000000}{12.000000}\selectfont \(\displaystyle {10^{2}}\)}%
\end{pgfscope}%
\begin{pgfscope}%
\pgfsetbuttcap%
\pgfsetroundjoin%
\definecolor{currentfill}{rgb}{0.000000,0.000000,0.000000}%
\pgfsetfillcolor{currentfill}%
\pgfsetlinewidth{0.803000pt}%
\definecolor{currentstroke}{rgb}{0.000000,0.000000,0.000000}%
\pgfsetstrokecolor{currentstroke}%
\pgfsetdash{}{0pt}%
\pgfsys@defobject{currentmarker}{\pgfqpoint{0.000000in}{-0.048611in}}{\pgfqpoint{0.000000in}{0.000000in}}{%
\pgfpathmoveto{\pgfqpoint{0.000000in}{0.000000in}}%
\pgfpathlineto{\pgfqpoint{0.000000in}{-0.048611in}}%
\pgfusepath{stroke,fill}%
}%
\begin{pgfscope}%
\pgfsys@transformshift{1.833215in}{0.487346in}%
\pgfsys@useobject{currentmarker}{}%
\end{pgfscope}%
\end{pgfscope}%
\begin{pgfscope}%
\definecolor{textcolor}{rgb}{0.000000,0.000000,0.000000}%
\pgfsetstrokecolor{textcolor}%
\pgfsetfillcolor{textcolor}%
\pgftext[x=1.833215in,y=0.390123in,,top]{\color{textcolor}\rmfamily\fontsize{10.000000}{12.000000}\selectfont \(\displaystyle {10^{3}}\)}%
\end{pgfscope}%
\begin{pgfscope}%
\pgfsetbuttcap%
\pgfsetroundjoin%
\definecolor{currentfill}{rgb}{0.000000,0.000000,0.000000}%
\pgfsetfillcolor{currentfill}%
\pgfsetlinewidth{0.602250pt}%
\definecolor{currentstroke}{rgb}{0.000000,0.000000,0.000000}%
\pgfsetstrokecolor{currentstroke}%
\pgfsetdash{}{0pt}%
\pgfsys@defobject{currentmarker}{\pgfqpoint{0.000000in}{-0.027778in}}{\pgfqpoint{0.000000in}{0.000000in}}{%
\pgfpathmoveto{\pgfqpoint{0.000000in}{0.000000in}}%
\pgfpathlineto{\pgfqpoint{0.000000in}{-0.027778in}}%
\pgfusepath{stroke,fill}%
}%
\begin{pgfscope}%
\pgfsys@transformshift{0.555108in}{0.487346in}%
\pgfsys@useobject{currentmarker}{}%
\end{pgfscope}%
\end{pgfscope}%
\begin{pgfscope}%
\pgfsetbuttcap%
\pgfsetroundjoin%
\definecolor{currentfill}{rgb}{0.000000,0.000000,0.000000}%
\pgfsetfillcolor{currentfill}%
\pgfsetlinewidth{0.602250pt}%
\definecolor{currentstroke}{rgb}{0.000000,0.000000,0.000000}%
\pgfsetstrokecolor{currentstroke}%
\pgfsetdash{}{0pt}%
\pgfsys@defobject{currentmarker}{\pgfqpoint{0.000000in}{-0.027778in}}{\pgfqpoint{0.000000in}{0.000000in}}{%
\pgfpathmoveto{\pgfqpoint{0.000000in}{0.000000in}}%
\pgfpathlineto{\pgfqpoint{0.000000in}{-0.027778in}}%
\pgfusepath{stroke,fill}%
}%
\begin{pgfscope}%
\pgfsys@transformshift{0.687578in}{0.487346in}%
\pgfsys@useobject{currentmarker}{}%
\end{pgfscope}%
\end{pgfscope}%
\begin{pgfscope}%
\pgfsetbuttcap%
\pgfsetroundjoin%
\definecolor{currentfill}{rgb}{0.000000,0.000000,0.000000}%
\pgfsetfillcolor{currentfill}%
\pgfsetlinewidth{0.602250pt}%
\definecolor{currentstroke}{rgb}{0.000000,0.000000,0.000000}%
\pgfsetstrokecolor{currentstroke}%
\pgfsetdash{}{0pt}%
\pgfsys@defobject{currentmarker}{\pgfqpoint{0.000000in}{-0.027778in}}{\pgfqpoint{0.000000in}{0.000000in}}{%
\pgfpathmoveto{\pgfqpoint{0.000000in}{0.000000in}}%
\pgfpathlineto{\pgfqpoint{0.000000in}{-0.027778in}}%
\pgfusepath{stroke,fill}%
}%
\begin{pgfscope}%
\pgfsys@transformshift{0.781568in}{0.487346in}%
\pgfsys@useobject{currentmarker}{}%
\end{pgfscope}%
\end{pgfscope}%
\begin{pgfscope}%
\pgfsetbuttcap%
\pgfsetroundjoin%
\definecolor{currentfill}{rgb}{0.000000,0.000000,0.000000}%
\pgfsetfillcolor{currentfill}%
\pgfsetlinewidth{0.602250pt}%
\definecolor{currentstroke}{rgb}{0.000000,0.000000,0.000000}%
\pgfsetstrokecolor{currentstroke}%
\pgfsetdash{}{0pt}%
\pgfsys@defobject{currentmarker}{\pgfqpoint{0.000000in}{-0.027778in}}{\pgfqpoint{0.000000in}{0.000000in}}{%
\pgfpathmoveto{\pgfqpoint{0.000000in}{0.000000in}}%
\pgfpathlineto{\pgfqpoint{0.000000in}{-0.027778in}}%
\pgfusepath{stroke,fill}%
}%
\begin{pgfscope}%
\pgfsys@transformshift{0.854471in}{0.487346in}%
\pgfsys@useobject{currentmarker}{}%
\end{pgfscope}%
\end{pgfscope}%
\begin{pgfscope}%
\pgfsetbuttcap%
\pgfsetroundjoin%
\definecolor{currentfill}{rgb}{0.000000,0.000000,0.000000}%
\pgfsetfillcolor{currentfill}%
\pgfsetlinewidth{0.602250pt}%
\definecolor{currentstroke}{rgb}{0.000000,0.000000,0.000000}%
\pgfsetstrokecolor{currentstroke}%
\pgfsetdash{}{0pt}%
\pgfsys@defobject{currentmarker}{\pgfqpoint{0.000000in}{-0.027778in}}{\pgfqpoint{0.000000in}{0.000000in}}{%
\pgfpathmoveto{\pgfqpoint{0.000000in}{0.000000in}}%
\pgfpathlineto{\pgfqpoint{0.000000in}{-0.027778in}}%
\pgfusepath{stroke,fill}%
}%
\begin{pgfscope}%
\pgfsys@transformshift{0.914038in}{0.487346in}%
\pgfsys@useobject{currentmarker}{}%
\end{pgfscope}%
\end{pgfscope}%
\begin{pgfscope}%
\pgfsetbuttcap%
\pgfsetroundjoin%
\definecolor{currentfill}{rgb}{0.000000,0.000000,0.000000}%
\pgfsetfillcolor{currentfill}%
\pgfsetlinewidth{0.602250pt}%
\definecolor{currentstroke}{rgb}{0.000000,0.000000,0.000000}%
\pgfsetstrokecolor{currentstroke}%
\pgfsetdash{}{0pt}%
\pgfsys@defobject{currentmarker}{\pgfqpoint{0.000000in}{-0.027778in}}{\pgfqpoint{0.000000in}{0.000000in}}{%
\pgfpathmoveto{\pgfqpoint{0.000000in}{0.000000in}}%
\pgfpathlineto{\pgfqpoint{0.000000in}{-0.027778in}}%
\pgfusepath{stroke,fill}%
}%
\begin{pgfscope}%
\pgfsys@transformshift{0.964401in}{0.487346in}%
\pgfsys@useobject{currentmarker}{}%
\end{pgfscope}%
\end{pgfscope}%
\begin{pgfscope}%
\pgfsetbuttcap%
\pgfsetroundjoin%
\definecolor{currentfill}{rgb}{0.000000,0.000000,0.000000}%
\pgfsetfillcolor{currentfill}%
\pgfsetlinewidth{0.602250pt}%
\definecolor{currentstroke}{rgb}{0.000000,0.000000,0.000000}%
\pgfsetstrokecolor{currentstroke}%
\pgfsetdash{}{0pt}%
\pgfsys@defobject{currentmarker}{\pgfqpoint{0.000000in}{-0.027778in}}{\pgfqpoint{0.000000in}{0.000000in}}{%
\pgfpathmoveto{\pgfqpoint{0.000000in}{0.000000in}}%
\pgfpathlineto{\pgfqpoint{0.000000in}{-0.027778in}}%
\pgfusepath{stroke,fill}%
}%
\begin{pgfscope}%
\pgfsys@transformshift{1.008028in}{0.487346in}%
\pgfsys@useobject{currentmarker}{}%
\end{pgfscope}%
\end{pgfscope}%
\begin{pgfscope}%
\pgfsetbuttcap%
\pgfsetroundjoin%
\definecolor{currentfill}{rgb}{0.000000,0.000000,0.000000}%
\pgfsetfillcolor{currentfill}%
\pgfsetlinewidth{0.602250pt}%
\definecolor{currentstroke}{rgb}{0.000000,0.000000,0.000000}%
\pgfsetstrokecolor{currentstroke}%
\pgfsetdash{}{0pt}%
\pgfsys@defobject{currentmarker}{\pgfqpoint{0.000000in}{-0.027778in}}{\pgfqpoint{0.000000in}{0.000000in}}{%
\pgfpathmoveto{\pgfqpoint{0.000000in}{0.000000in}}%
\pgfpathlineto{\pgfqpoint{0.000000in}{-0.027778in}}%
\pgfusepath{stroke,fill}%
}%
\begin{pgfscope}%
\pgfsys@transformshift{1.046509in}{0.487346in}%
\pgfsys@useobject{currentmarker}{}%
\end{pgfscope}%
\end{pgfscope}%
\begin{pgfscope}%
\pgfsetbuttcap%
\pgfsetroundjoin%
\definecolor{currentfill}{rgb}{0.000000,0.000000,0.000000}%
\pgfsetfillcolor{currentfill}%
\pgfsetlinewidth{0.602250pt}%
\definecolor{currentstroke}{rgb}{0.000000,0.000000,0.000000}%
\pgfsetstrokecolor{currentstroke}%
\pgfsetdash{}{0pt}%
\pgfsys@defobject{currentmarker}{\pgfqpoint{0.000000in}{-0.027778in}}{\pgfqpoint{0.000000in}{0.000000in}}{%
\pgfpathmoveto{\pgfqpoint{0.000000in}{0.000000in}}%
\pgfpathlineto{\pgfqpoint{0.000000in}{-0.027778in}}%
\pgfusepath{stroke,fill}%
}%
\begin{pgfscope}%
\pgfsys@transformshift{1.307391in}{0.487346in}%
\pgfsys@useobject{currentmarker}{}%
\end{pgfscope}%
\end{pgfscope}%
\begin{pgfscope}%
\pgfsetbuttcap%
\pgfsetroundjoin%
\definecolor{currentfill}{rgb}{0.000000,0.000000,0.000000}%
\pgfsetfillcolor{currentfill}%
\pgfsetlinewidth{0.602250pt}%
\definecolor{currentstroke}{rgb}{0.000000,0.000000,0.000000}%
\pgfsetstrokecolor{currentstroke}%
\pgfsetdash{}{0pt}%
\pgfsys@defobject{currentmarker}{\pgfqpoint{0.000000in}{-0.027778in}}{\pgfqpoint{0.000000in}{0.000000in}}{%
\pgfpathmoveto{\pgfqpoint{0.000000in}{0.000000in}}%
\pgfpathlineto{\pgfqpoint{0.000000in}{-0.027778in}}%
\pgfusepath{stroke,fill}%
}%
\begin{pgfscope}%
\pgfsys@transformshift{1.439862in}{0.487346in}%
\pgfsys@useobject{currentmarker}{}%
\end{pgfscope}%
\end{pgfscope}%
\begin{pgfscope}%
\pgfsetbuttcap%
\pgfsetroundjoin%
\definecolor{currentfill}{rgb}{0.000000,0.000000,0.000000}%
\pgfsetfillcolor{currentfill}%
\pgfsetlinewidth{0.602250pt}%
\definecolor{currentstroke}{rgb}{0.000000,0.000000,0.000000}%
\pgfsetstrokecolor{currentstroke}%
\pgfsetdash{}{0pt}%
\pgfsys@defobject{currentmarker}{\pgfqpoint{0.000000in}{-0.027778in}}{\pgfqpoint{0.000000in}{0.000000in}}{%
\pgfpathmoveto{\pgfqpoint{0.000000in}{0.000000in}}%
\pgfpathlineto{\pgfqpoint{0.000000in}{-0.027778in}}%
\pgfusepath{stroke,fill}%
}%
\begin{pgfscope}%
\pgfsys@transformshift{1.533851in}{0.487346in}%
\pgfsys@useobject{currentmarker}{}%
\end{pgfscope}%
\end{pgfscope}%
\begin{pgfscope}%
\pgfsetbuttcap%
\pgfsetroundjoin%
\definecolor{currentfill}{rgb}{0.000000,0.000000,0.000000}%
\pgfsetfillcolor{currentfill}%
\pgfsetlinewidth{0.602250pt}%
\definecolor{currentstroke}{rgb}{0.000000,0.000000,0.000000}%
\pgfsetstrokecolor{currentstroke}%
\pgfsetdash{}{0pt}%
\pgfsys@defobject{currentmarker}{\pgfqpoint{0.000000in}{-0.027778in}}{\pgfqpoint{0.000000in}{0.000000in}}{%
\pgfpathmoveto{\pgfqpoint{0.000000in}{0.000000in}}%
\pgfpathlineto{\pgfqpoint{0.000000in}{-0.027778in}}%
\pgfusepath{stroke,fill}%
}%
\begin{pgfscope}%
\pgfsys@transformshift{1.606755in}{0.487346in}%
\pgfsys@useobject{currentmarker}{}%
\end{pgfscope}%
\end{pgfscope}%
\begin{pgfscope}%
\pgfsetbuttcap%
\pgfsetroundjoin%
\definecolor{currentfill}{rgb}{0.000000,0.000000,0.000000}%
\pgfsetfillcolor{currentfill}%
\pgfsetlinewidth{0.602250pt}%
\definecolor{currentstroke}{rgb}{0.000000,0.000000,0.000000}%
\pgfsetstrokecolor{currentstroke}%
\pgfsetdash{}{0pt}%
\pgfsys@defobject{currentmarker}{\pgfqpoint{0.000000in}{-0.027778in}}{\pgfqpoint{0.000000in}{0.000000in}}{%
\pgfpathmoveto{\pgfqpoint{0.000000in}{0.000000in}}%
\pgfpathlineto{\pgfqpoint{0.000000in}{-0.027778in}}%
\pgfusepath{stroke,fill}%
}%
\begin{pgfscope}%
\pgfsys@transformshift{1.666322in}{0.487346in}%
\pgfsys@useobject{currentmarker}{}%
\end{pgfscope}%
\end{pgfscope}%
\begin{pgfscope}%
\pgfsetbuttcap%
\pgfsetroundjoin%
\definecolor{currentfill}{rgb}{0.000000,0.000000,0.000000}%
\pgfsetfillcolor{currentfill}%
\pgfsetlinewidth{0.602250pt}%
\definecolor{currentstroke}{rgb}{0.000000,0.000000,0.000000}%
\pgfsetstrokecolor{currentstroke}%
\pgfsetdash{}{0pt}%
\pgfsys@defobject{currentmarker}{\pgfqpoint{0.000000in}{-0.027778in}}{\pgfqpoint{0.000000in}{0.000000in}}{%
\pgfpathmoveto{\pgfqpoint{0.000000in}{0.000000in}}%
\pgfpathlineto{\pgfqpoint{0.000000in}{-0.027778in}}%
\pgfusepath{stroke,fill}%
}%
\begin{pgfscope}%
\pgfsys@transformshift{1.716685in}{0.487346in}%
\pgfsys@useobject{currentmarker}{}%
\end{pgfscope}%
\end{pgfscope}%
\begin{pgfscope}%
\pgfsetbuttcap%
\pgfsetroundjoin%
\definecolor{currentfill}{rgb}{0.000000,0.000000,0.000000}%
\pgfsetfillcolor{currentfill}%
\pgfsetlinewidth{0.602250pt}%
\definecolor{currentstroke}{rgb}{0.000000,0.000000,0.000000}%
\pgfsetstrokecolor{currentstroke}%
\pgfsetdash{}{0pt}%
\pgfsys@defobject{currentmarker}{\pgfqpoint{0.000000in}{-0.027778in}}{\pgfqpoint{0.000000in}{0.000000in}}{%
\pgfpathmoveto{\pgfqpoint{0.000000in}{0.000000in}}%
\pgfpathlineto{\pgfqpoint{0.000000in}{-0.027778in}}%
\pgfusepath{stroke,fill}%
}%
\begin{pgfscope}%
\pgfsys@transformshift{1.760311in}{0.487346in}%
\pgfsys@useobject{currentmarker}{}%
\end{pgfscope}%
\end{pgfscope}%
\begin{pgfscope}%
\pgfsetbuttcap%
\pgfsetroundjoin%
\definecolor{currentfill}{rgb}{0.000000,0.000000,0.000000}%
\pgfsetfillcolor{currentfill}%
\pgfsetlinewidth{0.602250pt}%
\definecolor{currentstroke}{rgb}{0.000000,0.000000,0.000000}%
\pgfsetstrokecolor{currentstroke}%
\pgfsetdash{}{0pt}%
\pgfsys@defobject{currentmarker}{\pgfqpoint{0.000000in}{-0.027778in}}{\pgfqpoint{0.000000in}{0.000000in}}{%
\pgfpathmoveto{\pgfqpoint{0.000000in}{0.000000in}}%
\pgfpathlineto{\pgfqpoint{0.000000in}{-0.027778in}}%
\pgfusepath{stroke,fill}%
}%
\begin{pgfscope}%
\pgfsys@transformshift{1.798792in}{0.487346in}%
\pgfsys@useobject{currentmarker}{}%
\end{pgfscope}%
\end{pgfscope}%
\begin{pgfscope}%
\pgfsetbuttcap%
\pgfsetroundjoin%
\definecolor{currentfill}{rgb}{0.000000,0.000000,0.000000}%
\pgfsetfillcolor{currentfill}%
\pgfsetlinewidth{0.602250pt}%
\definecolor{currentstroke}{rgb}{0.000000,0.000000,0.000000}%
\pgfsetstrokecolor{currentstroke}%
\pgfsetdash{}{0pt}%
\pgfsys@defobject{currentmarker}{\pgfqpoint{0.000000in}{-0.027778in}}{\pgfqpoint{0.000000in}{0.000000in}}{%
\pgfpathmoveto{\pgfqpoint{0.000000in}{0.000000in}}%
\pgfpathlineto{\pgfqpoint{0.000000in}{-0.027778in}}%
\pgfusepath{stroke,fill}%
}%
\begin{pgfscope}%
\pgfsys@transformshift{2.059675in}{0.487346in}%
\pgfsys@useobject{currentmarker}{}%
\end{pgfscope}%
\end{pgfscope}%
\begin{pgfscope}%
\pgfsetbuttcap%
\pgfsetroundjoin%
\definecolor{currentfill}{rgb}{0.000000,0.000000,0.000000}%
\pgfsetfillcolor{currentfill}%
\pgfsetlinewidth{0.602250pt}%
\definecolor{currentstroke}{rgb}{0.000000,0.000000,0.000000}%
\pgfsetstrokecolor{currentstroke}%
\pgfsetdash{}{0pt}%
\pgfsys@defobject{currentmarker}{\pgfqpoint{0.000000in}{-0.027778in}}{\pgfqpoint{0.000000in}{0.000000in}}{%
\pgfpathmoveto{\pgfqpoint{0.000000in}{0.000000in}}%
\pgfpathlineto{\pgfqpoint{0.000000in}{-0.027778in}}%
\pgfusepath{stroke,fill}%
}%
\begin{pgfscope}%
\pgfsys@transformshift{2.192145in}{0.487346in}%
\pgfsys@useobject{currentmarker}{}%
\end{pgfscope}%
\end{pgfscope}%
\begin{pgfscope}%
\definecolor{textcolor}{rgb}{0.000000,0.000000,0.000000}%
\pgfsetstrokecolor{textcolor}%
\pgfsetfillcolor{textcolor}%
\pgftext[x=1.312238in,y=0.211111in,,top]{\color{textcolor}\rmfamily\fontsize{8.000000}{9.600000}\selectfont Runtime, seconds (log-scale)}%
\end{pgfscope}%
\begin{pgfscope}%
\pgfsetbuttcap%
\pgfsetroundjoin%
\definecolor{currentfill}{rgb}{0.000000,0.000000,0.000000}%
\pgfsetfillcolor{currentfill}%
\pgfsetlinewidth{0.803000pt}%
\definecolor{currentstroke}{rgb}{0.000000,0.000000,0.000000}%
\pgfsetstrokecolor{currentstroke}%
\pgfsetdash{}{0pt}%
\pgfsys@defobject{currentmarker}{\pgfqpoint{-0.048611in}{0.000000in}}{\pgfqpoint{-0.000000in}{0.000000in}}{%
\pgfpathmoveto{\pgfqpoint{-0.000000in}{0.000000in}}%
\pgfpathlineto{\pgfqpoint{-0.048611in}{0.000000in}}%
\pgfusepath{stroke,fill}%
}%
\begin{pgfscope}%
\pgfsys@transformshift{0.420988in}{0.530031in}%
\pgfsys@useobject{currentmarker}{}%
\end{pgfscope}%
\end{pgfscope}%
\begin{pgfscope}%
\definecolor{textcolor}{rgb}{0.000000,0.000000,0.000000}%
\pgfsetstrokecolor{textcolor}%
\pgfsetfillcolor{textcolor}%
\pgftext[x=0.254321in, y=0.481806in, left, base]{\color{textcolor}\rmfamily\fontsize{10.000000}{12.000000}\selectfont \(\displaystyle {2}\)}%
\end{pgfscope}%
\begin{pgfscope}%
\pgfsetbuttcap%
\pgfsetroundjoin%
\definecolor{currentfill}{rgb}{0.000000,0.000000,0.000000}%
\pgfsetfillcolor{currentfill}%
\pgfsetlinewidth{0.803000pt}%
\definecolor{currentstroke}{rgb}{0.000000,0.000000,0.000000}%
\pgfsetstrokecolor{currentstroke}%
\pgfsetdash{}{0pt}%
\pgfsys@defobject{currentmarker}{\pgfqpoint{-0.048611in}{0.000000in}}{\pgfqpoint{-0.000000in}{0.000000in}}{%
\pgfpathmoveto{\pgfqpoint{-0.000000in}{0.000000in}}%
\pgfpathlineto{\pgfqpoint{-0.048611in}{0.000000in}}%
\pgfusepath{stroke,fill}%
}%
\begin{pgfscope}%
\pgfsys@transformshift{0.420988in}{1.011131in}%
\pgfsys@useobject{currentmarker}{}%
\end{pgfscope}%
\end{pgfscope}%
\begin{pgfscope}%
\definecolor{textcolor}{rgb}{0.000000,0.000000,0.000000}%
\pgfsetstrokecolor{textcolor}%
\pgfsetfillcolor{textcolor}%
\pgftext[x=0.254321in, y=0.962906in, left, base]{\color{textcolor}\rmfamily\fontsize{10.000000}{12.000000}\selectfont \(\displaystyle {4}\)}%
\end{pgfscope}%
\begin{pgfscope}%
\pgfsetbuttcap%
\pgfsetroundjoin%
\definecolor{currentfill}{rgb}{0.000000,0.000000,0.000000}%
\pgfsetfillcolor{currentfill}%
\pgfsetlinewidth{0.803000pt}%
\definecolor{currentstroke}{rgb}{0.000000,0.000000,0.000000}%
\pgfsetstrokecolor{currentstroke}%
\pgfsetdash{}{0pt}%
\pgfsys@defobject{currentmarker}{\pgfqpoint{-0.048611in}{0.000000in}}{\pgfqpoint{-0.000000in}{0.000000in}}{%
\pgfpathmoveto{\pgfqpoint{-0.000000in}{0.000000in}}%
\pgfpathlineto{\pgfqpoint{-0.048611in}{0.000000in}}%
\pgfusepath{stroke,fill}%
}%
\begin{pgfscope}%
\pgfsys@transformshift{0.420988in}{1.492230in}%
\pgfsys@useobject{currentmarker}{}%
\end{pgfscope}%
\end{pgfscope}%
\begin{pgfscope}%
\definecolor{textcolor}{rgb}{0.000000,0.000000,0.000000}%
\pgfsetstrokecolor{textcolor}%
\pgfsetfillcolor{textcolor}%
\pgftext[x=0.254321in, y=1.444005in, left, base]{\color{textcolor}\rmfamily\fontsize{10.000000}{12.000000}\selectfont \(\displaystyle {6}\)}%
\end{pgfscope}%
\begin{pgfscope}%
\definecolor{textcolor}{rgb}{0.000000,0.000000,0.000000}%
\pgfsetstrokecolor{textcolor}%
\pgfsetfillcolor{textcolor}%
\pgftext[x=0.198766in,y=1.218846in,,bottom,rotate=90.000000]{\color{textcolor}\rmfamily\fontsize{8.000000}{9.600000}\selectfont Subjective score}%
\end{pgfscope}%
\begin{pgfscope}%
\pgfpathrectangle{\pgfqpoint{0.420988in}{0.487346in}}{\pgfqpoint{1.782500in}{1.463000in}}%
\pgfusepath{clip}%
\pgfsetbuttcap%
\pgfsetroundjoin%
\definecolor{currentfill}{rgb}{1.000000,0.498039,0.054902}%
\pgfsetfillcolor{currentfill}%
\pgfsetlinewidth{1.505625pt}%
\definecolor{currentstroke}{rgb}{1.000000,0.498039,0.054902}%
\pgfsetstrokecolor{currentstroke}%
\pgfsetdash{}{0pt}%
\pgfsys@defobject{currentmarker}{\pgfqpoint{-0.021960in}{-0.021960in}}{\pgfqpoint{0.021960in}{0.021960in}}{%
\pgfpathmoveto{\pgfqpoint{-0.021960in}{-0.021960in}}%
\pgfpathlineto{\pgfqpoint{0.021960in}{0.021960in}}%
\pgfpathmoveto{\pgfqpoint{-0.021960in}{0.021960in}}%
\pgfpathlineto{\pgfqpoint{0.021960in}{-0.021960in}}%
\pgfusepath{stroke,fill}%
}%
\begin{pgfscope}%
\pgfsys@transformshift{0.924751in}{1.404911in}%
\pgfsys@useobject{currentmarker}{}%
\end{pgfscope}%
\begin{pgfscope}%
\pgfsys@transformshift{1.667626in}{1.883846in}%
\pgfsys@useobject{currentmarker}{}%
\end{pgfscope}%
\begin{pgfscope}%
\pgfsys@transformshift{0.834256in}{1.777523in}%
\pgfsys@useobject{currentmarker}{}%
\end{pgfscope}%
\begin{pgfscope}%
\pgfsys@transformshift{2.122465in}{1.267557in}%
\pgfsys@useobject{currentmarker}{}%
\end{pgfscope}%
\begin{pgfscope}%
\pgfsys@transformshift{1.545879in}{1.720031in}%
\pgfsys@useobject{currentmarker}{}%
\end{pgfscope}%
\begin{pgfscope}%
\pgfsys@transformshift{1.647493in}{1.524945in}%
\pgfsys@useobject{currentmarker}{}%
\end{pgfscope}%
\begin{pgfscope}%
\pgfsys@transformshift{1.700417in}{1.097488in}%
\pgfsys@useobject{currentmarker}{}%
\end{pgfscope}%
\begin{pgfscope}%
\pgfsys@transformshift{1.213184in}{1.613708in}%
\pgfsys@useobject{currentmarker}{}%
\end{pgfscope}%
\begin{pgfscope}%
\pgfsys@transformshift{2.106188in}{1.749138in}%
\pgfsys@useobject{currentmarker}{}%
\end{pgfscope}%
\begin{pgfscope}%
\pgfsys@transformshift{0.773380in}{1.335873in}%
\pgfsys@useobject{currentmarker}{}%
\end{pgfscope}%
\begin{pgfscope}%
\pgfsys@transformshift{0.751651in}{1.256492in}%
\pgfsys@useobject{currentmarker}{}%
\end{pgfscope}%
\begin{pgfscope}%
\pgfsys@transformshift{0.860941in}{1.387832in}%
\pgfsys@useobject{currentmarker}{}%
\end{pgfscope}%
\begin{pgfscope}%
\pgfsys@transformshift{1.197788in}{1.218725in}%
\pgfsys@useobject{currentmarker}{}%
\end{pgfscope}%
\begin{pgfscope}%
\pgfsys@transformshift{1.263765in}{1.204773in}%
\pgfsys@useobject{currentmarker}{}%
\end{pgfscope}%
\begin{pgfscope}%
\pgfsys@transformshift{1.312255in}{1.360890in}%
\pgfsys@useobject{currentmarker}{}%
\end{pgfscope}%
\begin{pgfscope}%
\pgfsys@transformshift{1.194804in}{1.378931in}%
\pgfsys@useobject{currentmarker}{}%
\end{pgfscope}%
\begin{pgfscope}%
\pgfsys@transformshift{1.039167in}{1.492230in}%
\pgfsys@useobject{currentmarker}{}%
\end{pgfscope}%
\begin{pgfscope}%
\pgfsys@transformshift{2.094872in}{1.686835in}%
\pgfsys@useobject{currentmarker}{}%
\end{pgfscope}%
\end{pgfscope}%
\begin{pgfscope}%
\pgfsetrectcap%
\pgfsetmiterjoin%
\pgfsetlinewidth{0.803000pt}%
\definecolor{currentstroke}{rgb}{0.000000,0.000000,0.000000}%
\pgfsetstrokecolor{currentstroke}%
\pgfsetdash{}{0pt}%
\pgfpathmoveto{\pgfqpoint{0.420988in}{0.487346in}}%
\pgfpathlineto{\pgfqpoint{0.420988in}{1.950346in}}%
\pgfusepath{stroke}%
\end{pgfscope}%
\begin{pgfscope}%
\pgfsetrectcap%
\pgfsetmiterjoin%
\pgfsetlinewidth{0.803000pt}%
\definecolor{currentstroke}{rgb}{0.000000,0.000000,0.000000}%
\pgfsetstrokecolor{currentstroke}%
\pgfsetdash{}{0pt}%
\pgfpathmoveto{\pgfqpoint{2.203488in}{0.487346in}}%
\pgfpathlineto{\pgfqpoint{2.203488in}{1.950346in}}%
\pgfusepath{stroke}%
\end{pgfscope}%
\begin{pgfscope}%
\pgfsetrectcap%
\pgfsetmiterjoin%
\pgfsetlinewidth{0.803000pt}%
\definecolor{currentstroke}{rgb}{0.000000,0.000000,0.000000}%
\pgfsetstrokecolor{currentstroke}%
\pgfsetdash{}{0pt}%
\pgfpathmoveto{\pgfqpoint{0.420988in}{0.487346in}}%
\pgfpathlineto{\pgfqpoint{2.203488in}{0.487346in}}%
\pgfusepath{stroke}%
\end{pgfscope}%
\begin{pgfscope}%
\pgfsetrectcap%
\pgfsetmiterjoin%
\pgfsetlinewidth{0.803000pt}%
\definecolor{currentstroke}{rgb}{0.000000,0.000000,0.000000}%
\pgfsetstrokecolor{currentstroke}%
\pgfsetdash{}{0pt}%
\pgfpathmoveto{\pgfqpoint{0.420988in}{1.950346in}}%
\pgfpathlineto{\pgfqpoint{2.203488in}{1.950346in}}%
\pgfusepath{stroke}%
\end{pgfscope}%
\begin{pgfscope}%
\pgfpathrectangle{\pgfqpoint{0.420988in}{0.487346in}}{\pgfqpoint{1.782500in}{1.463000in}}%
\pgfusepath{clip}%
\pgfsetbuttcap%
\pgfsetroundjoin%
\definecolor{currentfill}{rgb}{0.121569,0.466667,0.705882}%
\pgfsetfillcolor{currentfill}%
\pgfsetlinewidth{1.003750pt}%
\definecolor{currentstroke}{rgb}{0.121569,0.466667,0.705882}%
\pgfsetstrokecolor{currentstroke}%
\pgfsetdash{}{0pt}%
\pgfsys@defobject{currentmarker}{\pgfqpoint{-0.021960in}{-0.021960in}}{\pgfqpoint{0.021960in}{0.021960in}}{%
\pgfpathmoveto{\pgfqpoint{0.000000in}{-0.021960in}}%
\pgfpathcurveto{\pgfqpoint{0.005824in}{-0.021960in}}{\pgfqpoint{0.011410in}{-0.019646in}}{\pgfqpoint{0.015528in}{-0.015528in}}%
\pgfpathcurveto{\pgfqpoint{0.019646in}{-0.011410in}}{\pgfqpoint{0.021960in}{-0.005824in}}{\pgfqpoint{0.021960in}{0.000000in}}%
\pgfpathcurveto{\pgfqpoint{0.021960in}{0.005824in}}{\pgfqpoint{0.019646in}{0.011410in}}{\pgfqpoint{0.015528in}{0.015528in}}%
\pgfpathcurveto{\pgfqpoint{0.011410in}{0.019646in}}{\pgfqpoint{0.005824in}{0.021960in}}{\pgfqpoint{0.000000in}{0.021960in}}%
\pgfpathcurveto{\pgfqpoint{-0.005824in}{0.021960in}}{\pgfqpoint{-0.011410in}{0.019646in}}{\pgfqpoint{-0.015528in}{0.015528in}}%
\pgfpathcurveto{\pgfqpoint{-0.019646in}{0.011410in}}{\pgfqpoint{-0.021960in}{0.005824in}}{\pgfqpoint{-0.021960in}{0.000000in}}%
\pgfpathcurveto{\pgfqpoint{-0.021960in}{-0.005824in}}{\pgfqpoint{-0.019646in}{-0.011410in}}{\pgfqpoint{-0.015528in}{-0.015528in}}%
\pgfpathcurveto{\pgfqpoint{-0.011410in}{-0.019646in}}{\pgfqpoint{-0.005824in}{-0.021960in}}{\pgfqpoint{0.000000in}{-0.021960in}}%
\pgfpathlineto{\pgfqpoint{0.000000in}{-0.021960in}}%
\pgfpathclose%
\pgfusepath{stroke,fill}%
}%
\begin{pgfscope}%
\pgfsys@transformshift{0.502011in}{0.883158in}%
\pgfsys@useobject{currentmarker}{}%
\end{pgfscope}%
\begin{pgfscope}%
\pgfsys@transformshift{0.687578in}{0.553846in}%
\pgfsys@useobject{currentmarker}{}%
\end{pgfscope}%
\begin{pgfscope}%
\pgfsys@transformshift{0.935124in}{0.695048in}%
\pgfsys@useobject{currentmarker}{}%
\end{pgfscope}%
\begin{pgfscope}%
\pgfsys@transformshift{1.966555in}{1.074155in}%
\pgfsys@useobject{currentmarker}{}%
\end{pgfscope}%
\begin{pgfscope}%
\pgfsys@transformshift{1.077978in}{1.345976in}%
\pgfsys@useobject{currentmarker}{}%
\end{pgfscope}%
\begin{pgfscope}%
\pgfsys@transformshift{1.074997in}{0.938244in}%
\pgfsys@useobject{currentmarker}{}%
\end{pgfscope}%
\begin{pgfscope}%
\pgfsys@transformshift{1.421955in}{1.320478in}%
\pgfsys@useobject{currentmarker}{}%
\end{pgfscope}%
\begin{pgfscope}%
\pgfsys@transformshift{1.375025in}{1.203330in}%
\pgfsys@useobject{currentmarker}{}%
\end{pgfscope}%
\begin{pgfscope}%
\pgfsys@transformshift{0.708664in}{0.707797in}%
\pgfsys@useobject{currentmarker}{}%
\end{pgfscope}%
\begin{pgfscope}%
\pgfsys@transformshift{0.999756in}{0.844670in}%
\pgfsys@useobject{currentmarker}{}%
\end{pgfscope}%
\end{pgfscope}%
\begin{pgfscope}%
\definecolor{textcolor}{rgb}{0.000000,0.000000,0.000000}%
\pgfsetstrokecolor{textcolor}%
\pgfsetfillcolor{textcolor}%
\pgftext[x=0.925172in,y=1.424155in,left,base]{\color{textcolor}\rmfamily\fontsize{4.000000}{4.800000}\selectfont COMISR}%
\end{pgfscope}%
\begin{pgfscope}%
\definecolor{textcolor}{rgb}{0.000000,0.000000,0.000000}%
\pgfsetstrokecolor{textcolor}%
\pgfsetfillcolor{textcolor}%
\pgftext[x=1.673048in,y=1.903090in,left,base]{\color{textcolor}\rmfamily\fontsize{4.000000}{4.800000}\selectfont VRT}%
\end{pgfscope}%
\begin{pgfscope}%
\definecolor{textcolor}{rgb}{0.000000,0.000000,0.000000}%
\pgfsetstrokecolor{textcolor}%
\pgfsetfillcolor{textcolor}%
\pgftext[x=0.834812in,y=1.801578in,left,base]{\color{textcolor}\rmfamily\fontsize{4.000000}{4.800000}\selectfont BasicVSR}%
\end{pgfscope}%
\begin{pgfscope}%
\definecolor{textcolor}{rgb}{0.000000,0.000000,0.000000}%
\pgfsetstrokecolor{textcolor}%
\pgfsetfillcolor{textcolor}%
\pgftext[x=0.539722in,y=0.904808in,left,base]{\color{textcolor}\rmfamily\fontsize{4.000000}{4.800000}\selectfont SRMD}%
\end{pgfscope}%
\begin{pgfscope}%
\definecolor{textcolor}{rgb}{0.000000,0.000000,0.000000}%
\pgfsetstrokecolor{textcolor}%
\pgfsetfillcolor{textcolor}%
\pgftext[x=1.936991in,y=1.298828in,left,base]{\color{textcolor}\rmfamily\fontsize{4.000000}{4.800000}\selectfont D3Dnet}%
\end{pgfscope}%
\begin{pgfscope}%
\definecolor{textcolor}{rgb}{0.000000,0.000000,0.000000}%
\pgfsetstrokecolor{textcolor}%
\pgfsetfillcolor{textcolor}%
\pgftext[x=1.545942in,y=1.739275in,left,base]{\color{textcolor}\rmfamily\fontsize{4.000000}{4.800000}\selectfont DBVSR}%
\end{pgfscope}%
\begin{pgfscope}%
\definecolor{textcolor}{rgb}{0.000000,0.000000,0.000000}%
\pgfsetstrokecolor{textcolor}%
\pgfsetfillcolor{textcolor}%
\pgftext[x=1.658875in,y=1.556217in,left,base]{\color{textcolor}\rmfamily\fontsize{4.000000}{4.800000}\selectfont DynaVSR-R}%
\end{pgfscope}%
\begin{pgfscope}%
\definecolor{textcolor}{rgb}{0.000000,0.000000,0.000000}%
\pgfsetstrokecolor{textcolor}%
\pgfsetfillcolor{textcolor}%
\pgftext[x=1.700457in,y=1.116732in,left,base]{\color{textcolor}\rmfamily\fontsize{4.000000}{4.800000}\selectfont DynaVSR-V}%
\end{pgfscope}%
\begin{pgfscope}%
\definecolor{textcolor}{rgb}{0.000000,0.000000,0.000000}%
\pgfsetstrokecolor{textcolor}%
\pgfsetfillcolor{textcolor}%
\pgftext[x=0.688448in,y=0.573090in,left,base]{\color{textcolor}\rmfamily\fontsize{4.000000}{4.800000}\selectfont ESPCN}%
\end{pgfscope}%
\begin{pgfscope}%
\definecolor{textcolor}{rgb}{0.000000,0.000000,0.000000}%
\pgfsetstrokecolor{textcolor}%
\pgfsetfillcolor{textcolor}%
\pgftext[x=0.960079in,y=0.642127in,left,base]{\color{textcolor}\rmfamily\fontsize{4.000000}{4.800000}\selectfont GFPGAN}%
\end{pgfscope}%
\begin{pgfscope}%
\definecolor{textcolor}{rgb}{0.000000,0.000000,0.000000}%
\pgfsetstrokecolor{textcolor}%
\pgfsetfillcolor{textcolor}%
\pgftext[x=1.919956in,y=0.997179in,left,base]{\color{textcolor}\rmfamily\fontsize{4.000000}{4.800000}\selectfont HCFlow}%
\end{pgfscope}%
\begin{pgfscope}%
\definecolor{textcolor}{rgb}{0.000000,0.000000,0.000000}%
\pgfsetstrokecolor{textcolor}%
\pgfsetfillcolor{textcolor}%
\pgftext[x=1.213358in,y=1.637763in,left,base]{\color{textcolor}\rmfamily\fontsize{4.000000}{4.800000}\selectfont LGFN}%
\end{pgfscope}%
\begin{pgfscope}%
\definecolor{textcolor}{rgb}{0.000000,0.000000,0.000000}%
\pgfsetstrokecolor{textcolor}%
\pgfsetfillcolor{textcolor}%
\pgftext[x=1.944560in,y=1.780409in,left,base]{\color{textcolor}\rmfamily\fontsize{4.000000}{4.800000}\selectfont RBPN}%
\end{pgfscope}%
\begin{pgfscope}%
\definecolor{textcolor}{rgb}{0.000000,0.000000,0.000000}%
\pgfsetstrokecolor{textcolor}%
\pgfsetfillcolor{textcolor}%
\pgftext[x=0.440671in,y=1.319035in,left,base]{\color{textcolor}\rmfamily\fontsize{4.000000}{4.800000}\selectfont RRN-10L}%
\end{pgfscope}%
\begin{pgfscope}%
\definecolor{textcolor}{rgb}{0.000000,0.000000,0.000000}%
\pgfsetstrokecolor{textcolor}%
\pgfsetfillcolor{textcolor}%
\pgftext[x=0.428626in,y=1.239653in,left,base]{\color{textcolor}\rmfamily\fontsize{4.000000}{4.800000}\selectfont RRN-5L}%
\end{pgfscope}%
\begin{pgfscope}%
\definecolor{textcolor}{rgb}{0.000000,0.000000,0.000000}%
\pgfsetstrokecolor{textcolor}%
\pgfsetfillcolor{textcolor}%
\pgftext[x=0.641829in,y=1.411887in,left,base]{\color{textcolor}\rmfamily\fontsize{4.000000}{4.800000}\selectfont RSDN}%
\end{pgfscope}%
\begin{pgfscope}%
\definecolor{textcolor}{rgb}{0.000000,0.000000,0.000000}%
\pgfsetstrokecolor{textcolor}%
\pgfsetfillcolor{textcolor}%
\pgftext[x=0.835505in,y=1.273811in,left,base]{\color{textcolor}\rmfamily\fontsize{4.000000}{4.800000}\selectfont Real-ESRGAN}%
\end{pgfscope}%
\begin{pgfscope}%
\definecolor{textcolor}{rgb}{0.000000,0.000000,0.000000}%
\pgfsetstrokecolor{textcolor}%
\pgfsetfillcolor{textcolor}%
\pgftext[x=1.085087in,y=0.964705in,left,base]{\color{textcolor}\rmfamily\fontsize{4.000000}{4.800000}\selectfont Real-ESRnet}%
\end{pgfscope}%
\begin{pgfscope}%
\definecolor{textcolor}{rgb}{0.000000,0.000000,0.000000}%
\pgfsetstrokecolor{textcolor}%
\pgfsetfillcolor{textcolor}%
\pgftext[x=1.444275in,y=1.344533in,left,base]{\color{textcolor}\rmfamily\fontsize{4.000000}{4.800000}\selectfont RealSR}%
\end{pgfscope}%
\begin{pgfscope}%
\definecolor{textcolor}{rgb}{0.000000,0.000000,0.000000}%
\pgfsetstrokecolor{textcolor}%
\pgfsetfillcolor{textcolor}%
\pgftext[x=0.874001in,y=1.093639in,left,base]{\color{textcolor}\rmfamily\fontsize{4.000000}{4.800000}\selectfont SOF-VSR-BD}%
\end{pgfscope}%
\begin{pgfscope}%
\definecolor{textcolor}{rgb}{0.000000,0.000000,0.000000}%
\pgfsetstrokecolor{textcolor}%
\pgfsetfillcolor{textcolor}%
\pgftext[x=1.269465in,y=1.236045in,left,base]{\color{textcolor}\rmfamily\fontsize{4.000000}{4.800000}\selectfont SOF-VSR-BI}%
\end{pgfscope}%
\begin{pgfscope}%
\definecolor{textcolor}{rgb}{0.000000,0.000000,0.000000}%
\pgfsetstrokecolor{textcolor}%
\pgfsetfillcolor{textcolor}%
\pgftext[x=1.375132in,y=1.126354in,left,base]{\color{textcolor}\rmfamily\fontsize{4.000000}{4.800000}\selectfont SwinIR}%
\end{pgfscope}%
\begin{pgfscope}%
\definecolor{textcolor}{rgb}{0.000000,0.000000,0.000000}%
\pgfsetstrokecolor{textcolor}%
\pgfsetfillcolor{textcolor}%
\pgftext[x=1.343072in,y=1.392162in,left,base]{\color{textcolor}\rmfamily\fontsize{4.000000}{4.800000}\selectfont TDAN}%
\end{pgfscope}%
\begin{pgfscope}%
\definecolor{textcolor}{rgb}{0.000000,0.000000,0.000000}%
\pgfsetstrokecolor{textcolor}%
\pgfsetfillcolor{textcolor}%
\pgftext[x=1.217256in,y=1.417419in,left,base]{\color{textcolor}\rmfamily\fontsize{4.000000}{4.800000}\selectfont TGA}%
\end{pgfscope}%
\begin{pgfscope}%
\definecolor{textcolor}{rgb}{0.000000,0.000000,0.000000}%
\pgfsetstrokecolor{textcolor}%
\pgfsetfillcolor{textcolor}%
\pgftext[x=0.861453in,y=1.523502in,left,base]{\color{textcolor}\rmfamily\fontsize{4.000000}{4.800000}\selectfont TMNet}%
\end{pgfscope}%
\begin{pgfscope}%
\definecolor{textcolor}{rgb}{0.000000,0.000000,0.000000}%
\pgfsetstrokecolor{textcolor}%
\pgfsetfillcolor{textcolor}%
\pgftext[x=1.857626in,y=1.609859in,left,base]{\color{textcolor}\rmfamily\fontsize{4.000000}{4.800000}\selectfont iSeeBetter}%
\end{pgfscope}%
\begin{pgfscope}%
\definecolor{textcolor}{rgb}{0.000000,0.000000,0.000000}%
\pgfsetstrokecolor{textcolor}%
\pgfsetfillcolor{textcolor}%
\pgftext[x=0.724410in,y=0.731852in,left,base]{\color{textcolor}\rmfamily\fontsize{4.000000}{4.800000}\selectfont waifu2x-anime}%
\end{pgfscope}%
\begin{pgfscope}%
\definecolor{textcolor}{rgb}{0.000000,0.000000,0.000000}%
\pgfsetstrokecolor{textcolor}%
\pgfsetfillcolor{textcolor}%
\pgftext[x=1.012409in,y=0.867523in,left,base]{\color{textcolor}\rmfamily\fontsize{4.000000}{4.800000}\selectfont waifu2x-cunet}%
\end{pgfscope}%
\begin{pgfscope}%
\pgfsetbuttcap%
\pgfsetmiterjoin%
\definecolor{currentfill}{rgb}{1.000000,1.000000,1.000000}%
\pgfsetfillcolor{currentfill}%
\pgfsetfillopacity{0.800000}%
\pgfsetlinewidth{1.003750pt}%
\definecolor{currentstroke}{rgb}{0.800000,0.800000,0.800000}%
\pgfsetstrokecolor{currentstroke}%
\pgfsetstrokeopacity{0.800000}%
\pgfsetdash{}{0pt}%
\pgfpathmoveto{\pgfqpoint{1.075873in}{0.522068in}}%
\pgfpathlineto{\pgfqpoint{2.154877in}{0.522068in}}%
\pgfpathquadraticcurveto{\pgfqpoint{2.168766in}{0.522068in}}{\pgfqpoint{2.168766in}{0.535957in}}%
\pgfpathlineto{\pgfqpoint{2.168766in}{0.819521in}}%
\pgfpathquadraticcurveto{\pgfqpoint{2.168766in}{0.833410in}}{\pgfqpoint{2.154877in}{0.833410in}}%
\pgfpathlineto{\pgfqpoint{1.075873in}{0.833410in}}%
\pgfpathquadraticcurveto{\pgfqpoint{1.061984in}{0.833410in}}{\pgfqpoint{1.061984in}{0.819521in}}%
\pgfpathlineto{\pgfqpoint{1.061984in}{0.535957in}}%
\pgfpathquadraticcurveto{\pgfqpoint{1.061984in}{0.522068in}}{\pgfqpoint{1.075873in}{0.522068in}}%
\pgfpathlineto{\pgfqpoint{1.075873in}{0.522068in}}%
\pgfpathclose%
\pgfusepath{stroke,fill}%
\end{pgfscope}%
\begin{pgfscope}%
\pgfsetbuttcap%
\pgfsetroundjoin%
\definecolor{currentfill}{rgb}{0.121569,0.466667,0.705882}%
\pgfsetfillcolor{currentfill}%
\pgfsetlinewidth{1.003750pt}%
\definecolor{currentstroke}{rgb}{0.121569,0.466667,0.705882}%
\pgfsetstrokecolor{currentstroke}%
\pgfsetdash{}{0pt}%
\pgfsys@defobject{currentmarker}{\pgfqpoint{-0.021960in}{-0.021960in}}{\pgfqpoint{0.021960in}{0.021960in}}{%
\pgfpathmoveto{\pgfqpoint{0.000000in}{-0.021960in}}%
\pgfpathcurveto{\pgfqpoint{0.005824in}{-0.021960in}}{\pgfqpoint{0.011410in}{-0.019646in}}{\pgfqpoint{0.015528in}{-0.015528in}}%
\pgfpathcurveto{\pgfqpoint{0.019646in}{-0.011410in}}{\pgfqpoint{0.021960in}{-0.005824in}}{\pgfqpoint{0.021960in}{0.000000in}}%
\pgfpathcurveto{\pgfqpoint{0.021960in}{0.005824in}}{\pgfqpoint{0.019646in}{0.011410in}}{\pgfqpoint{0.015528in}{0.015528in}}%
\pgfpathcurveto{\pgfqpoint{0.011410in}{0.019646in}}{\pgfqpoint{0.005824in}{0.021960in}}{\pgfqpoint{0.000000in}{0.021960in}}%
\pgfpathcurveto{\pgfqpoint{-0.005824in}{0.021960in}}{\pgfqpoint{-0.011410in}{0.019646in}}{\pgfqpoint{-0.015528in}{0.015528in}}%
\pgfpathcurveto{\pgfqpoint{-0.019646in}{0.011410in}}{\pgfqpoint{-0.021960in}{0.005824in}}{\pgfqpoint{-0.021960in}{0.000000in}}%
\pgfpathcurveto{\pgfqpoint{-0.021960in}{-0.005824in}}{\pgfqpoint{-0.019646in}{-0.011410in}}{\pgfqpoint{-0.015528in}{-0.015528in}}%
\pgfpathcurveto{\pgfqpoint{-0.011410in}{-0.019646in}}{\pgfqpoint{-0.005824in}{-0.021960in}}{\pgfqpoint{0.000000in}{-0.021960in}}%
\pgfpathlineto{\pgfqpoint{0.000000in}{-0.021960in}}%
\pgfpathclose%
\pgfusepath{stroke,fill}%
}%
\begin{pgfscope}%
\pgfsys@transformshift{1.159206in}{0.775250in}%
\pgfsys@useobject{currentmarker}{}%
\end{pgfscope}%
\end{pgfscope}%
\begin{pgfscope}%
\definecolor{textcolor}{rgb}{0.000000,0.000000,0.000000}%
\pgfsetstrokecolor{textcolor}%
\pgfsetfillcolor{textcolor}%
\pgftext[x=1.284206in,y=0.757021in,left,base]{\color{textcolor}\rmfamily\fontsize{5.000000}{6.000000}\selectfont Image SR}%
\end{pgfscope}%
\begin{pgfscope}%
\pgfsetbuttcap%
\pgfsetroundjoin%
\definecolor{currentfill}{rgb}{1.000000,0.498039,0.054902}%
\pgfsetfillcolor{currentfill}%
\pgfsetlinewidth{1.505625pt}%
\definecolor{currentstroke}{rgb}{1.000000,0.498039,0.054902}%
\pgfsetstrokecolor{currentstroke}%
\pgfsetdash{}{0pt}%
\pgfsys@defobject{currentmarker}{\pgfqpoint{-0.021960in}{-0.021960in}}{\pgfqpoint{0.021960in}{0.021960in}}{%
\pgfpathmoveto{\pgfqpoint{-0.021960in}{-0.021960in}}%
\pgfpathlineto{\pgfqpoint{0.021960in}{0.021960in}}%
\pgfpathmoveto{\pgfqpoint{-0.021960in}{0.021960in}}%
\pgfpathlineto{\pgfqpoint{0.021960in}{-0.021960in}}%
\pgfusepath{stroke,fill}%
}%
\begin{pgfscope}%
\pgfsys@transformshift{1.159206in}{0.678414in}%
\pgfsys@useobject{currentmarker}{}%
\end{pgfscope}%
\end{pgfscope}%
\begin{pgfscope}%
\definecolor{textcolor}{rgb}{0.000000,0.000000,0.000000}%
\pgfsetstrokecolor{textcolor}%
\pgfsetfillcolor{textcolor}%
\pgftext[x=1.284206in,y=0.660185in,left,base]{\color{textcolor}\rmfamily\fontsize{5.000000}{6.000000}\selectfont Video SR}%
\end{pgfscope}%
\begin{pgfscope}%
\pgfsetbuttcap%
\pgfsetroundjoin%
\pgfsetlinewidth{1.003750pt}%
\definecolor{currentstroke}{rgb}{0.600000,0.600000,0.600000}%
\pgfsetstrokecolor{currentstroke}%
\pgfsetdash{{3.700000pt}{1.600000pt}}{0.000000pt}%
\pgfpathmoveto{\pgfqpoint{1.089762in}{0.587654in}}%
\pgfpathlineto{\pgfqpoint{1.159206in}{0.587654in}}%
\pgfpathlineto{\pgfqpoint{1.228650in}{0.587654in}}%
\pgfusepath{stroke}%
\end{pgfscope}%
\begin{pgfscope}%
\definecolor{textcolor}{rgb}{0.000000,0.000000,0.000000}%
\pgfsetstrokecolor{textcolor}%
\pgfsetfillcolor{textcolor}%
\pgftext[x=1.284206in,y=0.563348in,left,base]{\color{textcolor}\rmfamily\fontsize{5.000000}{6.000000}\selectfont Pareto-optimal front}%
\end{pgfscope}%
\end{pgfpicture}%
\makeatother%
\endgroup%

%% file: metric2.pgf
\begingroup%
\makeatletter%
\begin{pgfpicture}%
\pgfpathrectangle{\pgfpointorigin}{\pgfqpoint{3.338170in}{2.946258in}}%
\pgfusepath{use as bounding box, clip}%
\begin{pgfscope}%
\pgfsetbuttcap%
\pgfsetmiterjoin%
\pgfsetlinewidth{0.000000pt}%
\definecolor{currentstroke}{rgb}{1.000000,1.000000,1.000000}%
\pgfsetstrokecolor{currentstroke}%
\pgfsetstrokeopacity{0.000000}%
\pgfsetdash{}{0pt}%
\pgfpathmoveto{\pgfqpoint{0.000000in}{0.000000in}}%
\pgfpathlineto{\pgfqpoint{3.338170in}{0.000000in}}%
\pgfpathlineto{\pgfqpoint{3.338170in}{2.946258in}}%
\pgfpathlineto{\pgfqpoint{0.000000in}{2.946258in}}%
\pgfpathclose%
\pgfusepath{}%
\end{pgfscope}%
\begin{pgfscope}%
\pgfsetbuttcap%
\pgfsetmiterjoin%
\definecolor{currentfill}{rgb}{1.000000,1.000000,1.000000}%
\pgfsetfillcolor{currentfill}%
\pgfsetlinewidth{0.000000pt}%
\definecolor{currentstroke}{rgb}{0.000000,0.000000,0.000000}%
\pgfsetstrokecolor{currentstroke}%
\pgfsetstrokeopacity{0.000000}%
\pgfsetdash{}{0pt}%
\pgfpathmoveto{\pgfqpoint{0.279012in}{1.833591in}}%
\pgfpathlineto{\pgfqpoint{1.659286in}{1.833591in}}%
\pgfpathlineto{\pgfqpoint{1.659286in}{2.523727in}}%
\pgfpathlineto{\pgfqpoint{0.279012in}{2.523727in}}%
\pgfpathclose%
\pgfusepath{fill}%
\end{pgfscope}%
\begin{pgfscope}%
\pgfpathrectangle{\pgfqpoint{0.279012in}{1.833591in}}{\pgfqpoint{1.380273in}{0.690137in}}%
\pgfusepath{clip}%
\pgfsys@transformshift{0.279012in}{1.833591in}%
\pgftext[left,bottom]{\includegraphics[interpolate=true,width=1.388889in,height=0.694444in]{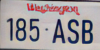}}%
\end{pgfscope}%
\begin{pgfscope}%
\definecolor{textcolor}{rgb}{0.000000,0.000000,0.000000}%
\pgfsetstrokecolor{textcolor}%
\pgfsetfillcolor{textcolor}%
\pgftext[x=0.223457in,y=2.178659in,,bottom,rotate=90.000000]{\color{textcolor}\rmfamily\fontsize{10.000000}{12.000000}\selectfont Input}%
\end{pgfscope}%
\begin{pgfscope}%
\pgfsetrectcap%
\pgfsetmiterjoin%
\pgfsetlinewidth{0.803000pt}%
\definecolor{currentstroke}{rgb}{0.000000,0.000000,0.000000}%
\pgfsetstrokecolor{currentstroke}%
\pgfsetdash{}{0pt}%
\pgfpathmoveto{\pgfqpoint{0.279012in}{1.833591in}}%
\pgfpathlineto{\pgfqpoint{0.279012in}{2.523727in}}%
\pgfusepath{stroke}%
\end{pgfscope}%
\begin{pgfscope}%
\pgfsetrectcap%
\pgfsetmiterjoin%
\pgfsetlinewidth{0.803000pt}%
\definecolor{currentstroke}{rgb}{0.000000,0.000000,0.000000}%
\pgfsetstrokecolor{currentstroke}%
\pgfsetdash{}{0pt}%
\pgfpathmoveto{\pgfqpoint{1.659286in}{1.833591in}}%
\pgfpathlineto{\pgfqpoint{1.659286in}{2.523727in}}%
\pgfusepath{stroke}%
\end{pgfscope}%
\begin{pgfscope}%
\pgfsetrectcap%
\pgfsetmiterjoin%
\pgfsetlinewidth{0.803000pt}%
\definecolor{currentstroke}{rgb}{0.000000,0.000000,0.000000}%
\pgfsetstrokecolor{currentstroke}%
\pgfsetdash{}{0pt}%
\pgfpathmoveto{\pgfqpoint{0.279012in}{1.833591in}}%
\pgfpathlineto{\pgfqpoint{1.659286in}{1.833591in}}%
\pgfusepath{stroke}%
\end{pgfscope}%
\begin{pgfscope}%
\pgfsetrectcap%
\pgfsetmiterjoin%
\pgfsetlinewidth{0.803000pt}%
\definecolor{currentstroke}{rgb}{0.000000,0.000000,0.000000}%
\pgfsetstrokecolor{currentstroke}%
\pgfsetdash{}{0pt}%
\pgfpathmoveto{\pgfqpoint{0.279012in}{2.523727in}}%
\pgfpathlineto{\pgfqpoint{1.659286in}{2.523727in}}%
\pgfusepath{stroke}%
\end{pgfscope}%
\begin{pgfscope}%
\definecolor{textcolor}{rgb}{0.000000,0.000000,0.000000}%
\pgfsetstrokecolor{textcolor}%
\pgfsetfillcolor{textcolor}%
\pgftext[x=0.743454in, y=2.749807in, left, base]{\color{textcolor}\rmfamily\fontsize{10.000000}{12.000000}\selectfont RealSR}%
\end{pgfscope}%
\begin{pgfscope}%
\definecolor{textcolor}{rgb}{0.000000,0.000000,0.000000}%
\pgfsetstrokecolor{textcolor}%
\pgfsetfillcolor{textcolor}%
\pgftext[x=0.417450in, y=2.607061in, left, base]{\color{textcolor}\rmfamily\fontsize{10.000000}{12.000000}\selectfont ERQAv2.0 = 0.89}%
\end{pgfscope}%
\begin{pgfscope}%
\pgfsetbuttcap%
\pgfsetmiterjoin%
\definecolor{currentfill}{rgb}{1.000000,1.000000,1.000000}%
\pgfsetfillcolor{currentfill}%
\pgfsetlinewidth{0.000000pt}%
\definecolor{currentstroke}{rgb}{0.000000,0.000000,0.000000}%
\pgfsetstrokecolor{currentstroke}%
\pgfsetstrokeopacity{0.000000}%
\pgfsetdash{}{0pt}%
\pgfpathmoveto{\pgfqpoint{1.857897in}{1.833591in}}%
\pgfpathlineto{\pgfqpoint{3.238170in}{1.833591in}}%
\pgfpathlineto{\pgfqpoint{3.238170in}{2.523727in}}%
\pgfpathlineto{\pgfqpoint{1.857897in}{2.523727in}}%
\pgfpathclose%
\pgfusepath{fill}%
\end{pgfscope}%
\begin{pgfscope}%
\pgfpathrectangle{\pgfqpoint{1.857897in}{1.833591in}}{\pgfqpoint{1.380273in}{0.690137in}}%
\pgfusepath{clip}%
\pgfsys@transformshift{1.857897in}{1.833591in}%
\pgftext[left,bottom]{\includegraphics[interpolate=true,width=1.388889in,height=0.694444in]{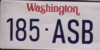}}%
\end{pgfscope}%
\begin{pgfscope}%
\pgfsetrectcap%
\pgfsetmiterjoin%
\pgfsetlinewidth{0.803000pt}%
\definecolor{currentstroke}{rgb}{0.000000,0.000000,0.000000}%
\pgfsetstrokecolor{currentstroke}%
\pgfsetdash{}{0pt}%
\pgfpathmoveto{\pgfqpoint{1.857897in}{1.833591in}}%
\pgfpathlineto{\pgfqpoint{1.857897in}{2.523727in}}%
\pgfusepath{stroke}%
\end{pgfscope}%
\begin{pgfscope}%
\pgfsetrectcap%
\pgfsetmiterjoin%
\pgfsetlinewidth{0.803000pt}%
\definecolor{currentstroke}{rgb}{0.000000,0.000000,0.000000}%
\pgfsetstrokecolor{currentstroke}%
\pgfsetdash{}{0pt}%
\pgfpathmoveto{\pgfqpoint{3.238170in}{1.833591in}}%
\pgfpathlineto{\pgfqpoint{3.238170in}{2.523727in}}%
\pgfusepath{stroke}%
\end{pgfscope}%
\begin{pgfscope}%
\pgfsetrectcap%
\pgfsetmiterjoin%
\pgfsetlinewidth{0.803000pt}%
\definecolor{currentstroke}{rgb}{0.000000,0.000000,0.000000}%
\pgfsetstrokecolor{currentstroke}%
\pgfsetdash{}{0pt}%
\pgfpathmoveto{\pgfqpoint{1.857897in}{1.833591in}}%
\pgfpathlineto{\pgfqpoint{3.238170in}{1.833591in}}%
\pgfusepath{stroke}%
\end{pgfscope}%
\begin{pgfscope}%
\pgfsetrectcap%
\pgfsetmiterjoin%
\pgfsetlinewidth{0.803000pt}%
\definecolor{currentstroke}{rgb}{0.000000,0.000000,0.000000}%
\pgfsetstrokecolor{currentstroke}%
\pgfsetdash{}{0pt}%
\pgfpathmoveto{\pgfqpoint{1.857897in}{2.523727in}}%
\pgfpathlineto{\pgfqpoint{3.238170in}{2.523727in}}%
\pgfusepath{stroke}%
\end{pgfscope}%
\begin{pgfscope}%
\definecolor{textcolor}{rgb}{0.000000,0.000000,0.000000}%
\pgfsetstrokecolor{textcolor}%
\pgfsetfillcolor{textcolor}%
\pgftext[x=2.304013in, y=2.749807in, left, base]{\color{textcolor}\rmfamily\fontsize{10.000000}{12.000000}\selectfont DBVSR}%
\end{pgfscope}%
\begin{pgfscope}%
\definecolor{textcolor}{rgb}{0.000000,0.000000,0.000000}%
\pgfsetstrokecolor{textcolor}%
\pgfsetfillcolor{textcolor}%
\pgftext[x=1.996335in, y=2.607061in, left, base]{\color{textcolor}\rmfamily\fontsize{10.000000}{12.000000}\selectfont ERQAv2.0 = 0.94}%
\end{pgfscope}%
\begin{pgfscope}%
\pgfsetbuttcap%
\pgfsetmiterjoin%
\definecolor{currentfill}{rgb}{1.000000,1.000000,1.000000}%
\pgfsetfillcolor{currentfill}%
\pgfsetlinewidth{0.000000pt}%
\definecolor{currentstroke}{rgb}{0.000000,0.000000,0.000000}%
\pgfsetstrokecolor{currentstroke}%
\pgfsetstrokeopacity{0.000000}%
\pgfsetdash{}{0pt}%
\pgfpathmoveto{\pgfqpoint{0.279012in}{0.991101in}}%
\pgfpathlineto{\pgfqpoint{1.659286in}{0.991101in}}%
\pgfpathlineto{\pgfqpoint{1.659286in}{1.681238in}}%
\pgfpathlineto{\pgfqpoint{0.279012in}{1.681238in}}%
\pgfpathclose%
\pgfusepath{fill}%
\end{pgfscope}%
\begin{pgfscope}%
\pgfpathrectangle{\pgfqpoint{0.279012in}{0.991101in}}{\pgfqpoint{1.380273in}{0.690137in}}%
\pgfusepath{clip}%
\pgfsys@transformshift{0.279012in}{0.991101in}%
\pgftext[left,bottom]{\includegraphics[interpolate=true,width=1.388889in,height=0.694444in]{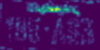}}%
\end{pgfscope}%
\begin{pgfscope}%
\definecolor{textcolor}{rgb}{0.000000,0.000000,0.000000}%
\pgfsetstrokecolor{textcolor}%
\pgfsetfillcolor{textcolor}%
\pgftext[x=0.223457in,y=1.336169in,,bottom,rotate=90.000000]{\color{textcolor}\rmfamily\fontsize{10.000000}{12.000000}\selectfont Heat map}%
\end{pgfscope}%
\begin{pgfscope}%
\pgfsetrectcap%
\pgfsetmiterjoin%
\pgfsetlinewidth{0.803000pt}%
\definecolor{currentstroke}{rgb}{0.000000,0.000000,0.000000}%
\pgfsetstrokecolor{currentstroke}%
\pgfsetdash{}{0pt}%
\pgfpathmoveto{\pgfqpoint{0.279012in}{0.991101in}}%
\pgfpathlineto{\pgfqpoint{0.279012in}{1.681238in}}%
\pgfusepath{stroke}%
\end{pgfscope}%
\begin{pgfscope}%
\pgfsetrectcap%
\pgfsetmiterjoin%
\pgfsetlinewidth{0.803000pt}%
\definecolor{currentstroke}{rgb}{0.000000,0.000000,0.000000}%
\pgfsetstrokecolor{currentstroke}%
\pgfsetdash{}{0pt}%
\pgfpathmoveto{\pgfqpoint{1.659286in}{0.991101in}}%
\pgfpathlineto{\pgfqpoint{1.659286in}{1.681238in}}%
\pgfusepath{stroke}%
\end{pgfscope}%
\begin{pgfscope}%
\pgfsetrectcap%
\pgfsetmiterjoin%
\pgfsetlinewidth{0.803000pt}%
\definecolor{currentstroke}{rgb}{0.000000,0.000000,0.000000}%
\pgfsetstrokecolor{currentstroke}%
\pgfsetdash{}{0pt}%
\pgfpathmoveto{\pgfqpoint{0.279012in}{0.991101in}}%
\pgfpathlineto{\pgfqpoint{1.659286in}{0.991101in}}%
\pgfusepath{stroke}%
\end{pgfscope}%
\begin{pgfscope}%
\pgfsetrectcap%
\pgfsetmiterjoin%
\pgfsetlinewidth{0.803000pt}%
\definecolor{currentstroke}{rgb}{0.000000,0.000000,0.000000}%
\pgfsetstrokecolor{currentstroke}%
\pgfsetdash{}{0pt}%
\pgfpathmoveto{\pgfqpoint{0.279012in}{1.681238in}}%
\pgfpathlineto{\pgfqpoint{1.659286in}{1.681238in}}%
\pgfusepath{stroke}%
\end{pgfscope}%
\begin{pgfscope}%
\pgfsetbuttcap%
\pgfsetmiterjoin%
\definecolor{currentfill}{rgb}{1.000000,1.000000,1.000000}%
\pgfsetfillcolor{currentfill}%
\pgfsetlinewidth{0.000000pt}%
\definecolor{currentstroke}{rgb}{0.000000,0.000000,0.000000}%
\pgfsetstrokecolor{currentstroke}%
\pgfsetstrokeopacity{0.000000}%
\pgfsetdash{}{0pt}%
\pgfpathmoveto{\pgfqpoint{1.857897in}{0.991101in}}%
\pgfpathlineto{\pgfqpoint{3.238170in}{0.991101in}}%
\pgfpathlineto{\pgfqpoint{3.238170in}{1.681238in}}%
\pgfpathlineto{\pgfqpoint{1.857897in}{1.681238in}}%
\pgfpathclose%
\pgfusepath{fill}%
\end{pgfscope}%
\begin{pgfscope}%
\pgfpathrectangle{\pgfqpoint{1.857897in}{0.991101in}}{\pgfqpoint{1.380273in}{0.690137in}}%
\pgfusepath{clip}%
\pgfsys@transformshift{1.857897in}{0.991101in}%
\pgftext[left,bottom]{\includegraphics[interpolate=true,width=1.388889in,height=0.694444in]{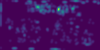}}%
\end{pgfscope}%
\begin{pgfscope}%
\pgfsetrectcap%
\pgfsetmiterjoin%
\pgfsetlinewidth{0.803000pt}%
\definecolor{currentstroke}{rgb}{0.000000,0.000000,0.000000}%
\pgfsetstrokecolor{currentstroke}%
\pgfsetdash{}{0pt}%
\pgfpathmoveto{\pgfqpoint{1.857897in}{0.991101in}}%
\pgfpathlineto{\pgfqpoint{1.857897in}{1.681238in}}%
\pgfusepath{stroke}%
\end{pgfscope}%
\begin{pgfscope}%
\pgfsetrectcap%
\pgfsetmiterjoin%
\pgfsetlinewidth{0.803000pt}%
\definecolor{currentstroke}{rgb}{0.000000,0.000000,0.000000}%
\pgfsetstrokecolor{currentstroke}%
\pgfsetdash{}{0pt}%
\pgfpathmoveto{\pgfqpoint{3.238170in}{0.991101in}}%
\pgfpathlineto{\pgfqpoint{3.238170in}{1.681238in}}%
\pgfusepath{stroke}%
\end{pgfscope}%
\begin{pgfscope}%
\pgfsetrectcap%
\pgfsetmiterjoin%
\pgfsetlinewidth{0.803000pt}%
\definecolor{currentstroke}{rgb}{0.000000,0.000000,0.000000}%
\pgfsetstrokecolor{currentstroke}%
\pgfsetdash{}{0pt}%
\pgfpathmoveto{\pgfqpoint{1.857897in}{0.991101in}}%
\pgfpathlineto{\pgfqpoint{3.238170in}{0.991101in}}%
\pgfusepath{stroke}%
\end{pgfscope}%
\begin{pgfscope}%
\pgfsetrectcap%
\pgfsetmiterjoin%
\pgfsetlinewidth{0.803000pt}%
\definecolor{currentstroke}{rgb}{0.000000,0.000000,0.000000}%
\pgfsetstrokecolor{currentstroke}%
\pgfsetdash{}{0pt}%
\pgfpathmoveto{\pgfqpoint{1.857897in}{1.681238in}}%
\pgfpathlineto{\pgfqpoint{3.238170in}{1.681238in}}%
\pgfusepath{stroke}%
\end{pgfscope}%
\begin{pgfscope}%
\pgfsetbuttcap%
\pgfsetmiterjoin%
\definecolor{currentfill}{rgb}{1.000000,1.000000,1.000000}%
\pgfsetfillcolor{currentfill}%
\pgfsetlinewidth{0.000000pt}%
\definecolor{currentstroke}{rgb}{0.000000,0.000000,0.000000}%
\pgfsetstrokecolor{currentstroke}%
\pgfsetstrokeopacity{0.000000}%
\pgfsetdash{}{0pt}%
\pgfpathmoveto{\pgfqpoint{0.279012in}{0.148611in}}%
\pgfpathlineto{\pgfqpoint{1.659286in}{0.148611in}}%
\pgfpathlineto{\pgfqpoint{1.659286in}{0.838748in}}%
\pgfpathlineto{\pgfqpoint{0.279012in}{0.838748in}}%
\pgfpathclose%
\pgfusepath{fill}%
\end{pgfscope}%
\begin{pgfscope}%
\pgfpathrectangle{\pgfqpoint{0.279012in}{0.148611in}}{\pgfqpoint{1.380273in}{0.690137in}}%
\pgfusepath{clip}%
\pgfsys@transformshift{0.279012in}{0.148611in}%
\pgftext[left,bottom]{\includegraphics[interpolate=true,width=1.388889in,height=0.694444in]{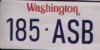}}%
\end{pgfscope}%
\begin{pgfscope}%
\definecolor{textcolor}{rgb}{0.000000,0.000000,0.000000}%
\pgfsetstrokecolor{textcolor}%
\pgfsetfillcolor{textcolor}%
\pgftext[x=0.223457in,y=0.493679in,,bottom,rotate=90.000000]{\color{textcolor}\rmfamily\fontsize{10.000000}{12.000000}\selectfont GT}%
\end{pgfscope}%
\begin{pgfscope}%
\pgfsetrectcap%
\pgfsetmiterjoin%
\pgfsetlinewidth{0.803000pt}%
\definecolor{currentstroke}{rgb}{0.000000,0.000000,0.000000}%
\pgfsetstrokecolor{currentstroke}%
\pgfsetdash{}{0pt}%
\pgfpathmoveto{\pgfqpoint{0.279012in}{0.148611in}}%
\pgfpathlineto{\pgfqpoint{0.279012in}{0.838748in}}%
\pgfusepath{stroke}%
\end{pgfscope}%
\begin{pgfscope}%
\pgfsetrectcap%
\pgfsetmiterjoin%
\pgfsetlinewidth{0.803000pt}%
\definecolor{currentstroke}{rgb}{0.000000,0.000000,0.000000}%
\pgfsetstrokecolor{currentstroke}%
\pgfsetdash{}{0pt}%
\pgfpathmoveto{\pgfqpoint{1.659286in}{0.148611in}}%
\pgfpathlineto{\pgfqpoint{1.659286in}{0.838748in}}%
\pgfusepath{stroke}%
\end{pgfscope}%
\begin{pgfscope}%
\pgfsetrectcap%
\pgfsetmiterjoin%
\pgfsetlinewidth{0.803000pt}%
\definecolor{currentstroke}{rgb}{0.000000,0.000000,0.000000}%
\pgfsetstrokecolor{currentstroke}%
\pgfsetdash{}{0pt}%
\pgfpathmoveto{\pgfqpoint{0.279012in}{0.148611in}}%
\pgfpathlineto{\pgfqpoint{1.659286in}{0.148611in}}%
\pgfusepath{stroke}%
\end{pgfscope}%
\begin{pgfscope}%
\pgfsetrectcap%
\pgfsetmiterjoin%
\pgfsetlinewidth{0.803000pt}%
\definecolor{currentstroke}{rgb}{0.000000,0.000000,0.000000}%
\pgfsetstrokecolor{currentstroke}%
\pgfsetdash{}{0pt}%
\pgfpathmoveto{\pgfqpoint{0.279012in}{0.838748in}}%
\pgfpathlineto{\pgfqpoint{1.659286in}{0.838748in}}%
\pgfusepath{stroke}%
\end{pgfscope}%
\begin{pgfscope}%
\pgfsetbuttcap%
\pgfsetmiterjoin%
\definecolor{currentfill}{rgb}{1.000000,1.000000,1.000000}%
\pgfsetfillcolor{currentfill}%
\pgfsetlinewidth{0.000000pt}%
\definecolor{currentstroke}{rgb}{0.000000,0.000000,0.000000}%
\pgfsetstrokecolor{currentstroke}%
\pgfsetstrokeopacity{0.000000}%
\pgfsetdash{}{0pt}%
\pgfpathmoveto{\pgfqpoint{1.857897in}{0.148611in}}%
\pgfpathlineto{\pgfqpoint{3.238170in}{0.148611in}}%
\pgfpathlineto{\pgfqpoint{3.238170in}{0.838748in}}%
\pgfpathlineto{\pgfqpoint{1.857897in}{0.838748in}}%
\pgfpathclose%
\pgfusepath{fill}%
\end{pgfscope}%
\begin{pgfscope}%
\pgfpathrectangle{\pgfqpoint{1.857897in}{0.148611in}}{\pgfqpoint{1.380273in}{0.690137in}}%
\pgfusepath{clip}%
\pgfsys@transformshift{1.857897in}{0.148611in}%
\pgftext[left,bottom]{\includegraphics[interpolate=true,width=1.388889in,height=0.694444in]{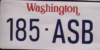}}%
\end{pgfscope}%
\begin{pgfscope}%
\pgfsetrectcap%
\pgfsetmiterjoin%
\pgfsetlinewidth{0.803000pt}%
\definecolor{currentstroke}{rgb}{0.000000,0.000000,0.000000}%
\pgfsetstrokecolor{currentstroke}%
\pgfsetdash{}{0pt}%
\pgfpathmoveto{\pgfqpoint{1.857897in}{0.148611in}}%
\pgfpathlineto{\pgfqpoint{1.857897in}{0.838748in}}%
\pgfusepath{stroke}%
\end{pgfscope}%
\begin{pgfscope}%
\pgfsetrectcap%
\pgfsetmiterjoin%
\pgfsetlinewidth{0.803000pt}%
\definecolor{currentstroke}{rgb}{0.000000,0.000000,0.000000}%
\pgfsetstrokecolor{currentstroke}%
\pgfsetdash{}{0pt}%
\pgfpathmoveto{\pgfqpoint{3.238170in}{0.148611in}}%
\pgfpathlineto{\pgfqpoint{3.238170in}{0.838748in}}%
\pgfusepath{stroke}%
\end{pgfscope}%
\begin{pgfscope}%
\pgfsetrectcap%
\pgfsetmiterjoin%
\pgfsetlinewidth{0.803000pt}%
\definecolor{currentstroke}{rgb}{0.000000,0.000000,0.000000}%
\pgfsetstrokecolor{currentstroke}%
\pgfsetdash{}{0pt}%
\pgfpathmoveto{\pgfqpoint{1.857897in}{0.148611in}}%
\pgfpathlineto{\pgfqpoint{3.238170in}{0.148611in}}%
\pgfusepath{stroke}%
\end{pgfscope}%
\begin{pgfscope}%
\pgfsetrectcap%
\pgfsetmiterjoin%
\pgfsetlinewidth{0.803000pt}%
\definecolor{currentstroke}{rgb}{0.000000,0.000000,0.000000}%
\pgfsetstrokecolor{currentstroke}%
\pgfsetdash{}{0pt}%
\pgfpathmoveto{\pgfqpoint{1.857897in}{0.838748in}}%
\pgfpathlineto{\pgfqpoint{3.238170in}{0.838748in}}%
\pgfusepath{stroke}%
\end{pgfscope}%
\end{pgfpicture}%
\makeatother%
\endgroup%

%% file: img/difference_noise_and_without.pgf
\begingroup%
\makeatletter%
\begin{pgfpicture}%
\pgfpathrectangle{\pgfpointorigin}{\pgfqpoint{3.194028in}{2.600648in}}%
\pgfusepath{use as bounding box, clip}%
\begin{pgfscope}%
\pgfsetbuttcap%
\pgfsetmiterjoin%
\definecolor{currentfill}{rgb}{1.000000,1.000000,1.000000}%
\pgfsetfillcolor{currentfill}%
\pgfsetlinewidth{0.000000pt}%
\definecolor{currentstroke}{rgb}{1.000000,1.000000,1.000000}%
\pgfsetstrokecolor{currentstroke}%
\pgfsetstrokeopacity{0.000000}%
\pgfsetdash{}{0pt}%
\pgfpathmoveto{\pgfqpoint{0.000000in}{0.000000in}}%
\pgfpathlineto{\pgfqpoint{3.194028in}{0.000000in}}%
\pgfpathlineto{\pgfqpoint{3.194028in}{2.600648in}}%
\pgfpathlineto{\pgfqpoint{0.000000in}{2.600648in}}%
\pgfpathclose%
\pgfusepath{fill}%
\end{pgfscope}%
\begin{pgfscope}%
\pgfsetbuttcap%
\pgfsetmiterjoin%
\definecolor{currentfill}{rgb}{1.000000,1.000000,1.000000}%
\pgfsetfillcolor{currentfill}%
\pgfsetlinewidth{0.000000pt}%
\definecolor{currentstroke}{rgb}{0.000000,0.000000,0.000000}%
\pgfsetstrokecolor{currentstroke}%
\pgfsetstrokeopacity{0.000000}%
\pgfsetdash{}{0pt}%
\pgfpathmoveto{\pgfqpoint{0.791528in}{0.335648in}}%
\pgfpathlineto{\pgfqpoint{3.194028in}{0.335648in}}%
\pgfpathlineto{\pgfqpoint{3.194028in}{2.600648in}}%
\pgfpathlineto{\pgfqpoint{0.791528in}{2.600648in}}%
\pgfpathclose%
\pgfusepath{fill}%
\end{pgfscope}%
\begin{pgfscope}%
\pgfpathrectangle{\pgfqpoint{0.791528in}{0.335648in}}{\pgfqpoint{2.402500in}{2.265000in}}%
\pgfusepath{clip}%
\pgfsetroundcap%
\pgfsetroundjoin%
\pgfsetlinewidth{1.003750pt}%
\definecolor{currentstroke}{rgb}{0.800000,0.800000,0.800000}%
\pgfsetstrokecolor{currentstroke}%
\pgfsetdash{}{0pt}%
\pgfpathmoveto{\pgfqpoint{0.791528in}{0.335648in}}%
\pgfpathlineto{\pgfqpoint{0.791528in}{2.600648in}}%
\pgfusepath{stroke}%
\end{pgfscope}%
\begin{pgfscope}%
\definecolor{textcolor}{rgb}{0.150000,0.150000,0.150000}%
\pgfsetstrokecolor{textcolor}%
\pgfsetfillcolor{textcolor}%
\pgftext[x=0.791528in,y=0.203703in,,top]{\color{textcolor}\sffamily\fontsize{6.000000}{7.200000}\selectfont 0.0}%
\end{pgfscope}%
\begin{pgfscope}%
\pgfpathrectangle{\pgfqpoint{0.791528in}{0.335648in}}{\pgfqpoint{2.402500in}{2.265000in}}%
\pgfusepath{clip}%
\pgfsetroundcap%
\pgfsetroundjoin%
\pgfsetlinewidth{1.003750pt}%
\definecolor{currentstroke}{rgb}{0.800000,0.800000,0.800000}%
\pgfsetstrokecolor{currentstroke}%
\pgfsetdash{}{0pt}%
\pgfpathmoveto{\pgfqpoint{1.167585in}{0.335648in}}%
\pgfpathlineto{\pgfqpoint{1.167585in}{2.600648in}}%
\pgfusepath{stroke}%
\end{pgfscope}%
\begin{pgfscope}%
\definecolor{textcolor}{rgb}{0.150000,0.150000,0.150000}%
\pgfsetstrokecolor{textcolor}%
\pgfsetfillcolor{textcolor}%
\pgftext[x=1.167585in,y=0.203703in,,top]{\color{textcolor}\sffamily\fontsize{6.000000}{7.200000}\selectfont 0.05}%
\end{pgfscope}%
\begin{pgfscope}%
\pgfpathrectangle{\pgfqpoint{0.791528in}{0.335648in}}{\pgfqpoint{2.402500in}{2.265000in}}%
\pgfusepath{clip}%
\pgfsetroundcap%
\pgfsetroundjoin%
\pgfsetlinewidth{1.003750pt}%
\definecolor{currentstroke}{rgb}{0.800000,0.800000,0.800000}%
\pgfsetstrokecolor{currentstroke}%
\pgfsetdash{}{0pt}%
\pgfpathmoveto{\pgfqpoint{1.543641in}{0.335648in}}%
\pgfpathlineto{\pgfqpoint{1.543641in}{2.600648in}}%
\pgfusepath{stroke}%
\end{pgfscope}%
\begin{pgfscope}%
\definecolor{textcolor}{rgb}{0.150000,0.150000,0.150000}%
\pgfsetstrokecolor{textcolor}%
\pgfsetfillcolor{textcolor}%
\pgftext[x=1.543641in,y=0.203703in,,top]{\color{textcolor}\sffamily\fontsize{6.000000}{7.200000}\selectfont 0.1}%
\end{pgfscope}%
\begin{pgfscope}%
\pgfpathrectangle{\pgfqpoint{0.791528in}{0.335648in}}{\pgfqpoint{2.402500in}{2.265000in}}%
\pgfusepath{clip}%
\pgfsetroundcap%
\pgfsetroundjoin%
\pgfsetlinewidth{1.003750pt}%
\definecolor{currentstroke}{rgb}{0.800000,0.800000,0.800000}%
\pgfsetstrokecolor{currentstroke}%
\pgfsetdash{}{0pt}%
\pgfpathmoveto{\pgfqpoint{1.919698in}{0.335648in}}%
\pgfpathlineto{\pgfqpoint{1.919698in}{2.600648in}}%
\pgfusepath{stroke}%
\end{pgfscope}%
\begin{pgfscope}%
\definecolor{textcolor}{rgb}{0.150000,0.150000,0.150000}%
\pgfsetstrokecolor{textcolor}%
\pgfsetfillcolor{textcolor}%
\pgftext[x=1.919698in,y=0.203703in,,top]{\color{textcolor}\sffamily\fontsize{6.000000}{7.200000}\selectfont 0.15}%
\end{pgfscope}%
\begin{pgfscope}%
\pgfpathrectangle{\pgfqpoint{0.791528in}{0.335648in}}{\pgfqpoint{2.402500in}{2.265000in}}%
\pgfusepath{clip}%
\pgfsetroundcap%
\pgfsetroundjoin%
\pgfsetlinewidth{1.003750pt}%
\definecolor{currentstroke}{rgb}{0.800000,0.800000,0.800000}%
\pgfsetstrokecolor{currentstroke}%
\pgfsetdash{}{0pt}%
\pgfpathmoveto{\pgfqpoint{2.295755in}{0.335648in}}%
\pgfpathlineto{\pgfqpoint{2.295755in}{2.600648in}}%
\pgfusepath{stroke}%
\end{pgfscope}%
\begin{pgfscope}%
\definecolor{textcolor}{rgb}{0.150000,0.150000,0.150000}%
\pgfsetstrokecolor{textcolor}%
\pgfsetfillcolor{textcolor}%
\pgftext[x=2.295755in,y=0.203703in,,top]{\color{textcolor}\sffamily\fontsize{6.000000}{7.200000}\selectfont 0.2}%
\end{pgfscope}%
\begin{pgfscope}%
\pgfpathrectangle{\pgfqpoint{0.791528in}{0.335648in}}{\pgfqpoint{2.402500in}{2.265000in}}%
\pgfusepath{clip}%
\pgfsetroundcap%
\pgfsetroundjoin%
\pgfsetlinewidth{1.003750pt}%
\definecolor{currentstroke}{rgb}{0.800000,0.800000,0.800000}%
\pgfsetstrokecolor{currentstroke}%
\pgfsetdash{}{0pt}%
\pgfpathmoveto{\pgfqpoint{2.671811in}{0.335648in}}%
\pgfpathlineto{\pgfqpoint{2.671811in}{2.600648in}}%
\pgfusepath{stroke}%
\end{pgfscope}%
\begin{pgfscope}%
\definecolor{textcolor}{rgb}{0.150000,0.150000,0.150000}%
\pgfsetstrokecolor{textcolor}%
\pgfsetfillcolor{textcolor}%
\pgftext[x=2.671811in,y=0.203703in,,top]{\color{textcolor}\sffamily\fontsize{6.000000}{7.200000}\selectfont 0.25}%
\end{pgfscope}%
\begin{pgfscope}%
\pgfpathrectangle{\pgfqpoint{0.791528in}{0.335648in}}{\pgfqpoint{2.402500in}{2.265000in}}%
\pgfusepath{clip}%
\pgfsetroundcap%
\pgfsetroundjoin%
\pgfsetlinewidth{1.003750pt}%
\definecolor{currentstroke}{rgb}{0.800000,0.800000,0.800000}%
\pgfsetstrokecolor{currentstroke}%
\pgfsetdash{}{0pt}%
\pgfpathmoveto{\pgfqpoint{3.047868in}{0.335648in}}%
\pgfpathlineto{\pgfqpoint{3.047868in}{2.600648in}}%
\pgfusepath{stroke}%
\end{pgfscope}%
\begin{pgfscope}%
\definecolor{textcolor}{rgb}{0.150000,0.150000,0.150000}%
\pgfsetstrokecolor{textcolor}%
\pgfsetfillcolor{textcolor}%
\pgftext[x=3.047868in,y=0.203703in,,top]{\color{textcolor}\sffamily\fontsize{6.000000}{7.200000}\selectfont 0.3}%
\end{pgfscope}%
\begin{pgfscope}%
\definecolor{textcolor}{rgb}{0.150000,0.150000,0.150000}%
\pgfsetstrokecolor{textcolor}%
\pgfsetfillcolor{textcolor}%
\pgftext[x=1.992778in,y=0.074074in,,top]{\color{textcolor}\sffamily\fontsize{6.000000}{7.200000}\selectfont ERQAv2.0}%
\end{pgfscope}%
\begin{pgfscope}%
\definecolor{textcolor}{rgb}{0.150000,0.150000,0.150000}%
\pgfsetstrokecolor{textcolor}%
\pgfsetfillcolor{textcolor}%
\pgftext[x=0.481051in, y=2.529768in, left, base]{\color{textcolor}\sffamily\fontsize{6.000000}{7.200000}\selectfont TGA}%
\end{pgfscope}%
\begin{pgfscope}%
\definecolor{textcolor}{rgb}{0.150000,0.150000,0.150000}%
\pgfsetstrokecolor{textcolor}%
\pgfsetfillcolor{textcolor}%
\pgftext[x=0.426653in, y=2.445879in, left, base]{\color{textcolor}\sffamily\fontsize{6.000000}{7.200000}\selectfont RSDN}%
\end{pgfscope}%
\begin{pgfscope}%
\definecolor{textcolor}{rgb}{0.150000,0.150000,0.150000}%
\pgfsetstrokecolor{textcolor}%
\pgfsetfillcolor{textcolor}%
\pgftext[x=0.319446in, y=2.361990in, left, base]{\color{textcolor}\sffamily\fontsize{6.000000}{7.200000}\selectfont ESRGAN}%
\end{pgfscope}%
\begin{pgfscope}%
\definecolor{textcolor}{rgb}{0.150000,0.150000,0.150000}%
\pgfsetstrokecolor{textcolor}%
\pgfsetfillcolor{textcolor}%
\pgftext[x=0.366467in, y=2.278102in, left, base]{\color{textcolor}\sffamily\fontsize{6.000000}{7.200000}\selectfont HCFlow}%
\end{pgfscope}%
\begin{pgfscope}%
\definecolor{textcolor}{rgb}{0.150000,0.150000,0.150000}%
\pgfsetstrokecolor{textcolor}%
\pgfsetfillcolor{textcolor}%
\pgftext[x=0.214266in, y=2.194213in, left, base]{\color{textcolor}\sffamily\fontsize{6.000000}{7.200000}\selectfont DynaVSR-R}%
\end{pgfscope}%
\begin{pgfscope}%
\definecolor{textcolor}{rgb}{0.150000,0.150000,0.150000}%
\pgfsetstrokecolor{textcolor}%
\pgfsetfillcolor{textcolor}%
\pgftext[x=0.212530in, y=2.110324in, left, base]{\color{textcolor}\sffamily\fontsize{6.000000}{7.200000}\selectfont DynaVSR-V}%
\end{pgfscope}%
\begin{pgfscope}%
\definecolor{textcolor}{rgb}{0.150000,0.150000,0.150000}%
\pgfsetstrokecolor{textcolor}%
\pgfsetfillcolor{textcolor}%
\pgftext[x=0.317999in, y=2.026435in, left, base]{\color{textcolor}\sffamily\fontsize{6.000000}{7.200000}\selectfont DUF-28L}%
\end{pgfscope}%
\begin{pgfscope}%
\definecolor{textcolor}{rgb}{0.150000,0.150000,0.150000}%
\pgfsetstrokecolor{textcolor}%
\pgfsetfillcolor{textcolor}%
\pgftext[x=0.439240in, y=1.942546in, left, base]{\color{textcolor}\sffamily\fontsize{6.000000}{7.200000}\selectfont LGFN}%
\end{pgfscope}%
\begin{pgfscope}%
\definecolor{textcolor}{rgb}{0.150000,0.150000,0.150000}%
\pgfsetstrokecolor{textcolor}%
\pgfsetfillcolor{textcolor}%
\pgftext[x=0.371096in, y=1.858657in, left, base]{\color{textcolor}\sffamily\fontsize{6.000000}{7.200000}\selectfont DBVSR}%
\end{pgfscope}%
\begin{pgfscope}%
\definecolor{textcolor}{rgb}{0.150000,0.150000,0.150000}%
\pgfsetstrokecolor{textcolor}%
\pgfsetfillcolor{textcolor}%
\pgftext[x=0.424193in, y=1.774768in, left, base]{\color{textcolor}\sffamily\fontsize{6.000000}{7.200000}\selectfont RBPN}%
\end{pgfscope}%
\begin{pgfscope}%
\definecolor{textcolor}{rgb}{0.150000,0.150000,0.150000}%
\pgfsetstrokecolor{textcolor}%
\pgfsetfillcolor{textcolor}%
\pgftext[x=0.278647in, y=1.690879in, left, base]{\color{textcolor}\sffamily\fontsize{6.000000}{7.200000}\selectfont iSeeBetter}%
\end{pgfscope}%
\begin{pgfscope}%
\definecolor{textcolor}{rgb}{0.150000,0.150000,0.150000}%
\pgfsetstrokecolor{textcolor}%
\pgfsetfillcolor{textcolor}%
\pgftext[x=0.388024in, y=1.606990in, left, base]{\color{textcolor}\sffamily\fontsize{6.000000}{7.200000}\selectfont TMNet}%
\end{pgfscope}%
\begin{pgfscope}%
\definecolor{textcolor}{rgb}{0.150000,0.150000,0.150000}%
\pgfsetstrokecolor{textcolor}%
\pgfsetfillcolor{textcolor}%
\pgftext[x=0.317999in, y=1.523102in, left, base]{\color{textcolor}\sffamily\fontsize{6.000000}{7.200000}\selectfont DUF-16L}%
\end{pgfscope}%
\begin{pgfscope}%
\definecolor{textcolor}{rgb}{0.150000,0.150000,0.150000}%
\pgfsetstrokecolor{textcolor}%
\pgfsetfillcolor{textcolor}%
\pgftext[x=0.146845in, y=1.439213in, left, base]{\color{textcolor}\sffamily\fontsize{6.000000}{7.200000}\selectfont SOF-VSR-BD}%
\end{pgfscope}%
\begin{pgfscope}%
\definecolor{textcolor}{rgb}{0.150000,0.150000,0.150000}%
\pgfsetstrokecolor{textcolor}%
\pgfsetfillcolor{textcolor}%
\pgftext[x=0.381658in, y=1.355324in, left, base]{\color{textcolor}\sffamily\fontsize{6.000000}{7.200000}\selectfont ESPCN}%
\end{pgfscope}%
\begin{pgfscope}%
\definecolor{textcolor}{rgb}{0.150000,0.150000,0.150000}%
\pgfsetstrokecolor{textcolor}%
\pgfsetfillcolor{textcolor}%
\pgftext[x=0.186197in, y=1.271435in, left, base]{\color{textcolor}\sffamily\fontsize{6.000000}{7.200000}\selectfont SOF-VSR-BI}%
\end{pgfscope}%
\begin{pgfscope}%
\definecolor{textcolor}{rgb}{0.150000,0.150000,0.150000}%
\pgfsetstrokecolor{textcolor}%
\pgfsetfillcolor{textcolor}%
\pgftext[x=0.370373in, y=1.187546in, left, base]{\color{textcolor}\sffamily\fontsize{6.000000}{7.200000}\selectfont D3Dnet}%
\end{pgfscope}%
\begin{pgfscope}%
\definecolor{textcolor}{rgb}{0.150000,0.150000,0.150000}%
\pgfsetstrokecolor{textcolor}%
\pgfsetfillcolor{textcolor}%
\pgftext[x=0.316264in, y=1.103657in, left, base]{\color{textcolor}\sffamily\fontsize{6.000000}{7.200000}\selectfont RRN-10L}%
\end{pgfscope}%
\begin{pgfscope}%
\definecolor{textcolor}{rgb}{0.150000,0.150000,0.150000}%
\pgfsetstrokecolor{textcolor}%
\pgfsetfillcolor{textcolor}%
\pgftext[x=0.360535in, y=1.019768in, left, base]{\color{textcolor}\sffamily\fontsize{6.000000}{7.200000}\selectfont RRN-5L}%
\end{pgfscope}%
\begin{pgfscope}%
\definecolor{textcolor}{rgb}{0.150000,0.150000,0.150000}%
\pgfsetstrokecolor{textcolor}%
\pgfsetfillcolor{textcolor}%
\pgftext[x=0.193143in, y=0.935879in, left, base]{\color{textcolor}\sffamily\fontsize{6.000000}{7.200000}\selectfont Real-ESRnet}%
\end{pgfscope}%
\begin{pgfscope}%
\definecolor{textcolor}{rgb}{0.150000,0.150000,0.150000}%
\pgfsetstrokecolor{textcolor}%
\pgfsetfillcolor{textcolor}%
\pgftext[x=0.148438in, y=0.851990in, left, base]{\color{textcolor}\sffamily\fontsize{6.000000}{7.200000}\selectfont waifu2x-cunet}%
\end{pgfscope}%
\begin{pgfscope}%
\definecolor{textcolor}{rgb}{0.150000,0.150000,0.150000}%
\pgfsetstrokecolor{textcolor}%
\pgfsetfillcolor{textcolor}%
\pgftext[x=0.411896in, y=0.768102in, left, base]{\color{textcolor}\sffamily\fontsize{6.000000}{7.200000}\selectfont SRMD}%
\end{pgfscope}%
\begin{pgfscope}%
\definecolor{textcolor}{rgb}{0.150000,0.150000,0.150000}%
\pgfsetstrokecolor{textcolor}%
\pgfsetfillcolor{textcolor}%
\pgftext[x=0.129629in, y=0.684213in, left, base]{\color{textcolor}\sffamily\fontsize{6.000000}{7.200000}\selectfont Real-ESRGAN}%
\end{pgfscope}%
\begin{pgfscope}%
\definecolor{textcolor}{rgb}{0.150000,0.150000,0.150000}%
\pgfsetstrokecolor{textcolor}%
\pgfsetfillcolor{textcolor}%
\pgftext[x=0.392798in, y=0.600324in, left, base]{\color{textcolor}\sffamily\fontsize{6.000000}{7.200000}\selectfont RealSR}%
\end{pgfscope}%
\begin{pgfscope}%
\definecolor{textcolor}{rgb}{0.150000,0.150000,0.150000}%
\pgfsetstrokecolor{textcolor}%
\pgfsetfillcolor{textcolor}%
\pgftext[x=0.401190in, y=0.516435in, left, base]{\color{textcolor}\sffamily\fontsize{6.000000}{7.200000}\selectfont SwinIR}%
\end{pgfscope}%
\begin{pgfscope}%
\definecolor{textcolor}{rgb}{0.150000,0.150000,0.150000}%
\pgfsetstrokecolor{textcolor}%
\pgfsetfillcolor{textcolor}%
\pgftext[x=0.131511in, y=0.432546in, left, base]{\color{textcolor}\sffamily\fontsize{6.000000}{7.200000}\selectfont waifu2x-anime}%
\end{pgfscope}%
\begin{pgfscope}%
\definecolor{textcolor}{rgb}{0.150000,0.150000,0.150000}%
\pgfsetstrokecolor{textcolor}%
\pgfsetfillcolor{textcolor}%
\pgftext[x=0.312791in, y=0.348657in, left, base]{\color{textcolor}\sffamily\fontsize{6.000000}{7.200000}\selectfont GFPGAN}%
\end{pgfscope}%
\begin{pgfscope}%
\definecolor{textcolor}{rgb}{0.150000,0.150000,0.150000}%
\pgfsetstrokecolor{textcolor}%
\pgfsetfillcolor{textcolor}%
\pgftext[x=0.074074in,y=1.468148in,,bottom,rotate=90.000000]{\color{textcolor}\sffamily\fontsize{6.000000}{7.200000}\selectfont Model}%
\end{pgfscope}%
\begin{pgfscope}%
\pgfpathrectangle{\pgfqpoint{0.791528in}{0.335648in}}{\pgfqpoint{2.402500in}{2.265000in}}%
\pgfusepath{clip}%
\pgfsetbuttcap%
\pgfsetmiterjoin%
\definecolor{currentfill}{rgb}{0.347059,0.458824,0.641176}%
\pgfsetfillcolor{currentfill}%
\pgfsetlinewidth{1.003750pt}%
\definecolor{currentstroke}{rgb}{1.000000,1.000000,1.000000}%
\pgfsetstrokecolor{currentstroke}%
\pgfsetdash{}{0pt}%
\pgfpathmoveto{\pgfqpoint{0.791528in}{2.592259in}}%
\pgfpathlineto{\pgfqpoint{3.079624in}{2.592259in}}%
\pgfpathlineto{\pgfqpoint{3.079624in}{2.525148in}}%
\pgfpathlineto{\pgfqpoint{0.791528in}{2.525148in}}%
\pgfpathclose%
\pgfusepath{stroke,fill}%
\end{pgfscope}%
\begin{pgfscope}%
\pgfpathrectangle{\pgfqpoint{0.791528in}{0.335648in}}{\pgfqpoint{2.402500in}{2.265000in}}%
\pgfusepath{clip}%
\pgfsetbuttcap%
\pgfsetmiterjoin%
\definecolor{currentfill}{rgb}{0.347059,0.458824,0.641176}%
\pgfsetfillcolor{currentfill}%
\pgfsetlinewidth{1.003750pt}%
\definecolor{currentstroke}{rgb}{1.000000,1.000000,1.000000}%
\pgfsetstrokecolor{currentstroke}%
\pgfsetdash{}{0pt}%
\pgfpathmoveto{\pgfqpoint{0.791528in}{2.508370in}}%
\pgfpathlineto{\pgfqpoint{2.860174in}{2.508370in}}%
\pgfpathlineto{\pgfqpoint{2.860174in}{2.441259in}}%
\pgfpathlineto{\pgfqpoint{0.791528in}{2.441259in}}%
\pgfpathclose%
\pgfusepath{stroke,fill}%
\end{pgfscope}%
\begin{pgfscope}%
\pgfpathrectangle{\pgfqpoint{0.791528in}{0.335648in}}{\pgfqpoint{2.402500in}{2.265000in}}%
\pgfusepath{clip}%
\pgfsetbuttcap%
\pgfsetmiterjoin%
\definecolor{currentfill}{rgb}{0.347059,0.458824,0.641176}%
\pgfsetfillcolor{currentfill}%
\pgfsetlinewidth{1.003750pt}%
\definecolor{currentstroke}{rgb}{1.000000,1.000000,1.000000}%
\pgfsetstrokecolor{currentstroke}%
\pgfsetdash{}{0pt}%
\pgfpathmoveto{\pgfqpoint{0.791528in}{2.424481in}}%
\pgfpathlineto{\pgfqpoint{2.290880in}{2.424481in}}%
\pgfpathlineto{\pgfqpoint{2.290880in}{2.357370in}}%
\pgfpathlineto{\pgfqpoint{0.791528in}{2.357370in}}%
\pgfpathclose%
\pgfusepath{stroke,fill}%
\end{pgfscope}%
\begin{pgfscope}%
\pgfpathrectangle{\pgfqpoint{0.791528in}{0.335648in}}{\pgfqpoint{2.402500in}{2.265000in}}%
\pgfusepath{clip}%
\pgfsetbuttcap%
\pgfsetmiterjoin%
\definecolor{currentfill}{rgb}{0.347059,0.458824,0.641176}%
\pgfsetfillcolor{currentfill}%
\pgfsetlinewidth{1.003750pt}%
\definecolor{currentstroke}{rgb}{1.000000,1.000000,1.000000}%
\pgfsetstrokecolor{currentstroke}%
\pgfsetdash{}{0pt}%
\pgfpathmoveto{\pgfqpoint{0.791528in}{2.340592in}}%
\pgfpathlineto{\pgfqpoint{2.141655in}{2.340592in}}%
\pgfpathlineto{\pgfqpoint{2.141655in}{2.273481in}}%
\pgfpathlineto{\pgfqpoint{0.791528in}{2.273481in}}%
\pgfpathclose%
\pgfusepath{stroke,fill}%
\end{pgfscope}%
\begin{pgfscope}%
\pgfpathrectangle{\pgfqpoint{0.791528in}{0.335648in}}{\pgfqpoint{2.402500in}{2.265000in}}%
\pgfusepath{clip}%
\pgfsetbuttcap%
\pgfsetmiterjoin%
\definecolor{currentfill}{rgb}{0.347059,0.458824,0.641176}%
\pgfsetfillcolor{currentfill}%
\pgfsetlinewidth{1.003750pt}%
\definecolor{currentstroke}{rgb}{1.000000,1.000000,1.000000}%
\pgfsetstrokecolor{currentstroke}%
\pgfsetdash{}{0pt}%
\pgfpathmoveto{\pgfqpoint{0.791528in}{2.256703in}}%
\pgfpathlineto{\pgfqpoint{2.020398in}{2.256703in}}%
\pgfpathlineto{\pgfqpoint{2.020398in}{2.189592in}}%
\pgfpathlineto{\pgfqpoint{0.791528in}{2.189592in}}%
\pgfpathclose%
\pgfusepath{stroke,fill}%
\end{pgfscope}%
\begin{pgfscope}%
\pgfpathrectangle{\pgfqpoint{0.791528in}{0.335648in}}{\pgfqpoint{2.402500in}{2.265000in}}%
\pgfusepath{clip}%
\pgfsetbuttcap%
\pgfsetmiterjoin%
\definecolor{currentfill}{rgb}{0.347059,0.458824,0.641176}%
\pgfsetfillcolor{currentfill}%
\pgfsetlinewidth{1.003750pt}%
\definecolor{currentstroke}{rgb}{1.000000,1.000000,1.000000}%
\pgfsetstrokecolor{currentstroke}%
\pgfsetdash{}{0pt}%
\pgfpathmoveto{\pgfqpoint{0.791528in}{2.172814in}}%
\pgfpathlineto{\pgfqpoint{1.938083in}{2.172814in}}%
\pgfpathlineto{\pgfqpoint{1.938083in}{2.105703in}}%
\pgfpathlineto{\pgfqpoint{0.791528in}{2.105703in}}%
\pgfpathclose%
\pgfusepath{stroke,fill}%
\end{pgfscope}%
\begin{pgfscope}%
\pgfpathrectangle{\pgfqpoint{0.791528in}{0.335648in}}{\pgfqpoint{2.402500in}{2.265000in}}%
\pgfusepath{clip}%
\pgfsetbuttcap%
\pgfsetmiterjoin%
\definecolor{currentfill}{rgb}{0.347059,0.458824,0.641176}%
\pgfsetfillcolor{currentfill}%
\pgfsetlinewidth{1.003750pt}%
\definecolor{currentstroke}{rgb}{1.000000,1.000000,1.000000}%
\pgfsetstrokecolor{currentstroke}%
\pgfsetdash{}{0pt}%
\pgfpathmoveto{\pgfqpoint{0.791528in}{2.088926in}}%
\pgfpathlineto{\pgfqpoint{1.846715in}{2.088926in}}%
\pgfpathlineto{\pgfqpoint{1.846715in}{2.021814in}}%
\pgfpathlineto{\pgfqpoint{0.791528in}{2.021814in}}%
\pgfpathclose%
\pgfusepath{stroke,fill}%
\end{pgfscope}%
\begin{pgfscope}%
\pgfpathrectangle{\pgfqpoint{0.791528in}{0.335648in}}{\pgfqpoint{2.402500in}{2.265000in}}%
\pgfusepath{clip}%
\pgfsetbuttcap%
\pgfsetmiterjoin%
\definecolor{currentfill}{rgb}{0.347059,0.458824,0.641176}%
\pgfsetfillcolor{currentfill}%
\pgfsetlinewidth{1.003750pt}%
\definecolor{currentstroke}{rgb}{1.000000,1.000000,1.000000}%
\pgfsetstrokecolor{currentstroke}%
\pgfsetdash{}{0pt}%
\pgfpathmoveto{\pgfqpoint{0.791528in}{2.005037in}}%
\pgfpathlineto{\pgfqpoint{1.712728in}{2.005037in}}%
\pgfpathlineto{\pgfqpoint{1.712728in}{1.937926in}}%
\pgfpathlineto{\pgfqpoint{0.791528in}{1.937926in}}%
\pgfpathclose%
\pgfusepath{stroke,fill}%
\end{pgfscope}%
\begin{pgfscope}%
\pgfpathrectangle{\pgfqpoint{0.791528in}{0.335648in}}{\pgfqpoint{2.402500in}{2.265000in}}%
\pgfusepath{clip}%
\pgfsetbuttcap%
\pgfsetmiterjoin%
\definecolor{currentfill}{rgb}{0.347059,0.458824,0.641176}%
\pgfsetfillcolor{currentfill}%
\pgfsetlinewidth{1.003750pt}%
\definecolor{currentstroke}{rgb}{1.000000,1.000000,1.000000}%
\pgfsetstrokecolor{currentstroke}%
\pgfsetdash{}{0pt}%
\pgfpathmoveto{\pgfqpoint{0.791528in}{1.921148in}}%
\pgfpathlineto{\pgfqpoint{1.699580in}{1.921148in}}%
\pgfpathlineto{\pgfqpoint{1.699580in}{1.854037in}}%
\pgfpathlineto{\pgfqpoint{0.791528in}{1.854037in}}%
\pgfpathclose%
\pgfusepath{stroke,fill}%
\end{pgfscope}%
\begin{pgfscope}%
\pgfpathrectangle{\pgfqpoint{0.791528in}{0.335648in}}{\pgfqpoint{2.402500in}{2.265000in}}%
\pgfusepath{clip}%
\pgfsetbuttcap%
\pgfsetmiterjoin%
\definecolor{currentfill}{rgb}{0.347059,0.458824,0.641176}%
\pgfsetfillcolor{currentfill}%
\pgfsetlinewidth{1.003750pt}%
\definecolor{currentstroke}{rgb}{1.000000,1.000000,1.000000}%
\pgfsetstrokecolor{currentstroke}%
\pgfsetdash{}{0pt}%
\pgfpathmoveto{\pgfqpoint{0.791528in}{1.837259in}}%
\pgfpathlineto{\pgfqpoint{1.621917in}{1.837259in}}%
\pgfpathlineto{\pgfqpoint{1.621917in}{1.770148in}}%
\pgfpathlineto{\pgfqpoint{0.791528in}{1.770148in}}%
\pgfpathclose%
\pgfusepath{stroke,fill}%
\end{pgfscope}%
\begin{pgfscope}%
\pgfpathrectangle{\pgfqpoint{0.791528in}{0.335648in}}{\pgfqpoint{2.402500in}{2.265000in}}%
\pgfusepath{clip}%
\pgfsetbuttcap%
\pgfsetmiterjoin%
\definecolor{currentfill}{rgb}{0.347059,0.458824,0.641176}%
\pgfsetfillcolor{currentfill}%
\pgfsetlinewidth{1.003750pt}%
\definecolor{currentstroke}{rgb}{1.000000,1.000000,1.000000}%
\pgfsetstrokecolor{currentstroke}%
\pgfsetdash{}{0pt}%
\pgfpathmoveto{\pgfqpoint{0.791528in}{1.753370in}}%
\pgfpathlineto{\pgfqpoint{1.568294in}{1.753370in}}%
\pgfpathlineto{\pgfqpoint{1.568294in}{1.686259in}}%
\pgfpathlineto{\pgfqpoint{0.791528in}{1.686259in}}%
\pgfpathclose%
\pgfusepath{stroke,fill}%
\end{pgfscope}%
\begin{pgfscope}%
\pgfpathrectangle{\pgfqpoint{0.791528in}{0.335648in}}{\pgfqpoint{2.402500in}{2.265000in}}%
\pgfusepath{clip}%
\pgfsetbuttcap%
\pgfsetmiterjoin%
\definecolor{currentfill}{rgb}{0.347059,0.458824,0.641176}%
\pgfsetfillcolor{currentfill}%
\pgfsetlinewidth{1.003750pt}%
\definecolor{currentstroke}{rgb}{1.000000,1.000000,1.000000}%
\pgfsetstrokecolor{currentstroke}%
\pgfsetdash{}{0pt}%
\pgfpathmoveto{\pgfqpoint{0.791528in}{1.669481in}}%
\pgfpathlineto{\pgfqpoint{1.522889in}{1.669481in}}%
\pgfpathlineto{\pgfqpoint{1.522889in}{1.602370in}}%
\pgfpathlineto{\pgfqpoint{0.791528in}{1.602370in}}%
\pgfpathclose%
\pgfusepath{stroke,fill}%
\end{pgfscope}%
\begin{pgfscope}%
\pgfpathrectangle{\pgfqpoint{0.791528in}{0.335648in}}{\pgfqpoint{2.402500in}{2.265000in}}%
\pgfusepath{clip}%
\pgfsetbuttcap%
\pgfsetmiterjoin%
\definecolor{currentfill}{rgb}{0.347059,0.458824,0.641176}%
\pgfsetfillcolor{currentfill}%
\pgfsetlinewidth{1.003750pt}%
\definecolor{currentstroke}{rgb}{1.000000,1.000000,1.000000}%
\pgfsetstrokecolor{currentstroke}%
\pgfsetdash{}{0pt}%
\pgfpathmoveto{\pgfqpoint{0.791528in}{1.585592in}}%
\pgfpathlineto{\pgfqpoint{1.490576in}{1.585592in}}%
\pgfpathlineto{\pgfqpoint{1.490576in}{1.518481in}}%
\pgfpathlineto{\pgfqpoint{0.791528in}{1.518481in}}%
\pgfpathclose%
\pgfusepath{stroke,fill}%
\end{pgfscope}%
\begin{pgfscope}%
\pgfpathrectangle{\pgfqpoint{0.791528in}{0.335648in}}{\pgfqpoint{2.402500in}{2.265000in}}%
\pgfusepath{clip}%
\pgfsetbuttcap%
\pgfsetmiterjoin%
\definecolor{currentfill}{rgb}{0.347059,0.458824,0.641176}%
\pgfsetfillcolor{currentfill}%
\pgfsetlinewidth{1.003750pt}%
\definecolor{currentstroke}{rgb}{1.000000,1.000000,1.000000}%
\pgfsetstrokecolor{currentstroke}%
\pgfsetdash{}{0pt}%
\pgfpathmoveto{\pgfqpoint{0.791528in}{1.501703in}}%
\pgfpathlineto{\pgfqpoint{1.461466in}{1.501703in}}%
\pgfpathlineto{\pgfqpoint{1.461466in}{1.434592in}}%
\pgfpathlineto{\pgfqpoint{0.791528in}{1.434592in}}%
\pgfpathclose%
\pgfusepath{stroke,fill}%
\end{pgfscope}%
\begin{pgfscope}%
\pgfpathrectangle{\pgfqpoint{0.791528in}{0.335648in}}{\pgfqpoint{2.402500in}{2.265000in}}%
\pgfusepath{clip}%
\pgfsetbuttcap%
\pgfsetmiterjoin%
\definecolor{currentfill}{rgb}{0.347059,0.458824,0.641176}%
\pgfsetfillcolor{currentfill}%
\pgfsetlinewidth{1.003750pt}%
\definecolor{currentstroke}{rgb}{1.000000,1.000000,1.000000}%
\pgfsetstrokecolor{currentstroke}%
\pgfsetdash{}{0pt}%
\pgfpathmoveto{\pgfqpoint{0.791528in}{1.417814in}}%
\pgfpathlineto{\pgfqpoint{1.396562in}{1.417814in}}%
\pgfpathlineto{\pgfqpoint{1.396562in}{1.350703in}}%
\pgfpathlineto{\pgfqpoint{0.791528in}{1.350703in}}%
\pgfpathclose%
\pgfusepath{stroke,fill}%
\end{pgfscope}%
\begin{pgfscope}%
\pgfpathrectangle{\pgfqpoint{0.791528in}{0.335648in}}{\pgfqpoint{2.402500in}{2.265000in}}%
\pgfusepath{clip}%
\pgfsetbuttcap%
\pgfsetmiterjoin%
\definecolor{currentfill}{rgb}{0.347059,0.458824,0.641176}%
\pgfsetfillcolor{currentfill}%
\pgfsetlinewidth{1.003750pt}%
\definecolor{currentstroke}{rgb}{1.000000,1.000000,1.000000}%
\pgfsetstrokecolor{currentstroke}%
\pgfsetdash{}{0pt}%
\pgfpathmoveto{\pgfqpoint{0.791528in}{1.333926in}}%
\pgfpathlineto{\pgfqpoint{1.373023in}{1.333926in}}%
\pgfpathlineto{\pgfqpoint{1.373023in}{1.266814in}}%
\pgfpathlineto{\pgfqpoint{0.791528in}{1.266814in}}%
\pgfpathclose%
\pgfusepath{stroke,fill}%
\end{pgfscope}%
\begin{pgfscope}%
\pgfpathrectangle{\pgfqpoint{0.791528in}{0.335648in}}{\pgfqpoint{2.402500in}{2.265000in}}%
\pgfusepath{clip}%
\pgfsetbuttcap%
\pgfsetmiterjoin%
\definecolor{currentfill}{rgb}{0.347059,0.458824,0.641176}%
\pgfsetfillcolor{currentfill}%
\pgfsetlinewidth{1.003750pt}%
\definecolor{currentstroke}{rgb}{1.000000,1.000000,1.000000}%
\pgfsetstrokecolor{currentstroke}%
\pgfsetdash{}{0pt}%
\pgfpathmoveto{\pgfqpoint{0.791528in}{1.250037in}}%
\pgfpathlineto{\pgfqpoint{1.319122in}{1.250037in}}%
\pgfpathlineto{\pgfqpoint{1.319122in}{1.182926in}}%
\pgfpathlineto{\pgfqpoint{0.791528in}{1.182926in}}%
\pgfpathclose%
\pgfusepath{stroke,fill}%
\end{pgfscope}%
\begin{pgfscope}%
\pgfpathrectangle{\pgfqpoint{0.791528in}{0.335648in}}{\pgfqpoint{2.402500in}{2.265000in}}%
\pgfusepath{clip}%
\pgfsetbuttcap%
\pgfsetmiterjoin%
\definecolor{currentfill}{rgb}{0.347059,0.458824,0.641176}%
\pgfsetfillcolor{currentfill}%
\pgfsetlinewidth{1.003750pt}%
\definecolor{currentstroke}{rgb}{1.000000,1.000000,1.000000}%
\pgfsetstrokecolor{currentstroke}%
\pgfsetdash{}{0pt}%
\pgfpathmoveto{\pgfqpoint{0.791528in}{1.166148in}}%
\pgfpathlineto{\pgfqpoint{1.271766in}{1.166148in}}%
\pgfpathlineto{\pgfqpoint{1.271766in}{1.099037in}}%
\pgfpathlineto{\pgfqpoint{0.791528in}{1.099037in}}%
\pgfpathclose%
\pgfusepath{stroke,fill}%
\end{pgfscope}%
\begin{pgfscope}%
\pgfpathrectangle{\pgfqpoint{0.791528in}{0.335648in}}{\pgfqpoint{2.402500in}{2.265000in}}%
\pgfusepath{clip}%
\pgfsetbuttcap%
\pgfsetmiterjoin%
\definecolor{currentfill}{rgb}{0.347059,0.458824,0.641176}%
\pgfsetfillcolor{currentfill}%
\pgfsetlinewidth{1.003750pt}%
\definecolor{currentstroke}{rgb}{1.000000,1.000000,1.000000}%
\pgfsetstrokecolor{currentstroke}%
\pgfsetdash{}{0pt}%
\pgfpathmoveto{\pgfqpoint{0.791528in}{1.082259in}}%
\pgfpathlineto{\pgfqpoint{1.185413in}{1.082259in}}%
\pgfpathlineto{\pgfqpoint{1.185413in}{1.015148in}}%
\pgfpathlineto{\pgfqpoint{0.791528in}{1.015148in}}%
\pgfpathclose%
\pgfusepath{stroke,fill}%
\end{pgfscope}%
\begin{pgfscope}%
\pgfpathrectangle{\pgfqpoint{0.791528in}{0.335648in}}{\pgfqpoint{2.402500in}{2.265000in}}%
\pgfusepath{clip}%
\pgfsetbuttcap%
\pgfsetmiterjoin%
\definecolor{currentfill}{rgb}{0.347059,0.458824,0.641176}%
\pgfsetfillcolor{currentfill}%
\pgfsetlinewidth{1.003750pt}%
\definecolor{currentstroke}{rgb}{1.000000,1.000000,1.000000}%
\pgfsetstrokecolor{currentstroke}%
\pgfsetdash{}{0pt}%
\pgfpathmoveto{\pgfqpoint{0.791528in}{0.998370in}}%
\pgfpathlineto{\pgfqpoint{1.150593in}{0.998370in}}%
\pgfpathlineto{\pgfqpoint{1.150593in}{0.931259in}}%
\pgfpathlineto{\pgfqpoint{0.791528in}{0.931259in}}%
\pgfpathclose%
\pgfusepath{stroke,fill}%
\end{pgfscope}%
\begin{pgfscope}%
\pgfpathrectangle{\pgfqpoint{0.791528in}{0.335648in}}{\pgfqpoint{2.402500in}{2.265000in}}%
\pgfusepath{clip}%
\pgfsetbuttcap%
\pgfsetmiterjoin%
\definecolor{currentfill}{rgb}{0.347059,0.458824,0.641176}%
\pgfsetfillcolor{currentfill}%
\pgfsetlinewidth{1.003750pt}%
\definecolor{currentstroke}{rgb}{1.000000,1.000000,1.000000}%
\pgfsetstrokecolor{currentstroke}%
\pgfsetdash{}{0pt}%
\pgfpathmoveto{\pgfqpoint{0.791528in}{0.914481in}}%
\pgfpathlineto{\pgfqpoint{1.135829in}{0.914481in}}%
\pgfpathlineto{\pgfqpoint{1.135829in}{0.847370in}}%
\pgfpathlineto{\pgfqpoint{0.791528in}{0.847370in}}%
\pgfpathclose%
\pgfusepath{stroke,fill}%
\end{pgfscope}%
\begin{pgfscope}%
\pgfpathrectangle{\pgfqpoint{0.791528in}{0.335648in}}{\pgfqpoint{2.402500in}{2.265000in}}%
\pgfusepath{clip}%
\pgfsetbuttcap%
\pgfsetmiterjoin%
\definecolor{currentfill}{rgb}{0.347059,0.458824,0.641176}%
\pgfsetfillcolor{currentfill}%
\pgfsetlinewidth{1.003750pt}%
\definecolor{currentstroke}{rgb}{1.000000,1.000000,1.000000}%
\pgfsetstrokecolor{currentstroke}%
\pgfsetdash{}{0pt}%
\pgfpathmoveto{\pgfqpoint{0.791528in}{0.830592in}}%
\pgfpathlineto{\pgfqpoint{1.071482in}{0.830592in}}%
\pgfpathlineto{\pgfqpoint{1.071482in}{0.763481in}}%
\pgfpathlineto{\pgfqpoint{0.791528in}{0.763481in}}%
\pgfpathclose%
\pgfusepath{stroke,fill}%
\end{pgfscope}%
\begin{pgfscope}%
\pgfpathrectangle{\pgfqpoint{0.791528in}{0.335648in}}{\pgfqpoint{2.402500in}{2.265000in}}%
\pgfusepath{clip}%
\pgfsetbuttcap%
\pgfsetmiterjoin%
\definecolor{currentfill}{rgb}{0.347059,0.458824,0.641176}%
\pgfsetfillcolor{currentfill}%
\pgfsetlinewidth{1.003750pt}%
\definecolor{currentstroke}{rgb}{1.000000,1.000000,1.000000}%
\pgfsetstrokecolor{currentstroke}%
\pgfsetdash{}{0pt}%
\pgfpathmoveto{\pgfqpoint{0.791528in}{0.746703in}}%
\pgfpathlineto{\pgfqpoint{1.044183in}{0.746703in}}%
\pgfpathlineto{\pgfqpoint{1.044183in}{0.679592in}}%
\pgfpathlineto{\pgfqpoint{0.791528in}{0.679592in}}%
\pgfpathclose%
\pgfusepath{stroke,fill}%
\end{pgfscope}%
\begin{pgfscope}%
\pgfpathrectangle{\pgfqpoint{0.791528in}{0.335648in}}{\pgfqpoint{2.402500in}{2.265000in}}%
\pgfusepath{clip}%
\pgfsetbuttcap%
\pgfsetmiterjoin%
\definecolor{currentfill}{rgb}{0.347059,0.458824,0.641176}%
\pgfsetfillcolor{currentfill}%
\pgfsetlinewidth{1.003750pt}%
\definecolor{currentstroke}{rgb}{1.000000,1.000000,1.000000}%
\pgfsetstrokecolor{currentstroke}%
\pgfsetdash{}{0pt}%
\pgfpathmoveto{\pgfqpoint{0.791528in}{0.662814in}}%
\pgfpathlineto{\pgfqpoint{1.028583in}{0.662814in}}%
\pgfpathlineto{\pgfqpoint{1.028583in}{0.595703in}}%
\pgfpathlineto{\pgfqpoint{0.791528in}{0.595703in}}%
\pgfpathclose%
\pgfusepath{stroke,fill}%
\end{pgfscope}%
\begin{pgfscope}%
\pgfpathrectangle{\pgfqpoint{0.791528in}{0.335648in}}{\pgfqpoint{2.402500in}{2.265000in}}%
\pgfusepath{clip}%
\pgfsetbuttcap%
\pgfsetmiterjoin%
\definecolor{currentfill}{rgb}{0.347059,0.458824,0.641176}%
\pgfsetfillcolor{currentfill}%
\pgfsetlinewidth{1.003750pt}%
\definecolor{currentstroke}{rgb}{1.000000,1.000000,1.000000}%
\pgfsetstrokecolor{currentstroke}%
\pgfsetdash{}{0pt}%
\pgfpathmoveto{\pgfqpoint{0.791528in}{0.578926in}}%
\pgfpathlineto{\pgfqpoint{0.989306in}{0.578926in}}%
\pgfpathlineto{\pgfqpoint{0.989306in}{0.511814in}}%
\pgfpathlineto{\pgfqpoint{0.791528in}{0.511814in}}%
\pgfpathclose%
\pgfusepath{stroke,fill}%
\end{pgfscope}%
\begin{pgfscope}%
\pgfpathrectangle{\pgfqpoint{0.791528in}{0.335648in}}{\pgfqpoint{2.402500in}{2.265000in}}%
\pgfusepath{clip}%
\pgfsetbuttcap%
\pgfsetmiterjoin%
\definecolor{currentfill}{rgb}{0.347059,0.458824,0.641176}%
\pgfsetfillcolor{currentfill}%
\pgfsetlinewidth{1.003750pt}%
\definecolor{currentstroke}{rgb}{1.000000,1.000000,1.000000}%
\pgfsetstrokecolor{currentstroke}%
\pgfsetdash{}{0pt}%
\pgfpathmoveto{\pgfqpoint{0.791528in}{0.495037in}}%
\pgfpathlineto{\pgfqpoint{0.951812in}{0.495037in}}%
\pgfpathlineto{\pgfqpoint{0.951812in}{0.427926in}}%
\pgfpathlineto{\pgfqpoint{0.791528in}{0.427926in}}%
\pgfpathclose%
\pgfusepath{stroke,fill}%
\end{pgfscope}%
\begin{pgfscope}%
\pgfpathrectangle{\pgfqpoint{0.791528in}{0.335648in}}{\pgfqpoint{2.402500in}{2.265000in}}%
\pgfusepath{clip}%
\pgfsetbuttcap%
\pgfsetmiterjoin%
\definecolor{currentfill}{rgb}{0.347059,0.458824,0.641176}%
\pgfsetfillcolor{currentfill}%
\pgfsetlinewidth{1.003750pt}%
\definecolor{currentstroke}{rgb}{1.000000,1.000000,1.000000}%
\pgfsetstrokecolor{currentstroke}%
\pgfsetdash{}{0pt}%
\pgfpathmoveto{\pgfqpoint{0.791528in}{0.411148in}}%
\pgfpathlineto{\pgfqpoint{0.919109in}{0.411148in}}%
\pgfpathlineto{\pgfqpoint{0.919109in}{0.344037in}}%
\pgfpathlineto{\pgfqpoint{0.791528in}{0.344037in}}%
\pgfpathclose%
\pgfusepath{stroke,fill}%
\end{pgfscope}%
\begin{pgfscope}%
\pgfpathrectangle{\pgfqpoint{0.791528in}{0.335648in}}{\pgfqpoint{2.402500in}{2.265000in}}%
\pgfusepath{clip}%
\pgfsetroundcap%
\pgfsetroundjoin%
\pgfsetlinewidth{2.710125pt}%
\definecolor{currentstroke}{rgb}{0.260000,0.260000,0.260000}%
\pgfsetstrokecolor{currentstroke}%
\pgfsetdash{}{0pt}%
\pgfusepath{stroke}%
\end{pgfscope}%
\begin{pgfscope}%
\pgfpathrectangle{\pgfqpoint{0.791528in}{0.335648in}}{\pgfqpoint{2.402500in}{2.265000in}}%
\pgfusepath{clip}%
\pgfsetroundcap%
\pgfsetroundjoin%
\pgfsetlinewidth{2.710125pt}%
\definecolor{currentstroke}{rgb}{0.260000,0.260000,0.260000}%
\pgfsetstrokecolor{currentstroke}%
\pgfsetdash{}{0pt}%
\pgfusepath{stroke}%
\end{pgfscope}%
\begin{pgfscope}%
\pgfpathrectangle{\pgfqpoint{0.791528in}{0.335648in}}{\pgfqpoint{2.402500in}{2.265000in}}%
\pgfusepath{clip}%
\pgfsetroundcap%
\pgfsetroundjoin%
\pgfsetlinewidth{2.710125pt}%
\definecolor{currentstroke}{rgb}{0.260000,0.260000,0.260000}%
\pgfsetstrokecolor{currentstroke}%
\pgfsetdash{}{0pt}%
\pgfusepath{stroke}%
\end{pgfscope}%
\begin{pgfscope}%
\pgfpathrectangle{\pgfqpoint{0.791528in}{0.335648in}}{\pgfqpoint{2.402500in}{2.265000in}}%
\pgfusepath{clip}%
\pgfsetroundcap%
\pgfsetroundjoin%
\pgfsetlinewidth{2.710125pt}%
\definecolor{currentstroke}{rgb}{0.260000,0.260000,0.260000}%
\pgfsetstrokecolor{currentstroke}%
\pgfsetdash{}{0pt}%
\pgfusepath{stroke}%
\end{pgfscope}%
\begin{pgfscope}%
\pgfpathrectangle{\pgfqpoint{0.791528in}{0.335648in}}{\pgfqpoint{2.402500in}{2.265000in}}%
\pgfusepath{clip}%
\pgfsetroundcap%
\pgfsetroundjoin%
\pgfsetlinewidth{2.710125pt}%
\definecolor{currentstroke}{rgb}{0.260000,0.260000,0.260000}%
\pgfsetstrokecolor{currentstroke}%
\pgfsetdash{}{0pt}%
\pgfusepath{stroke}%
\end{pgfscope}%
\begin{pgfscope}%
\pgfpathrectangle{\pgfqpoint{0.791528in}{0.335648in}}{\pgfqpoint{2.402500in}{2.265000in}}%
\pgfusepath{clip}%
\pgfsetroundcap%
\pgfsetroundjoin%
\pgfsetlinewidth{2.710125pt}%
\definecolor{currentstroke}{rgb}{0.260000,0.260000,0.260000}%
\pgfsetstrokecolor{currentstroke}%
\pgfsetdash{}{0pt}%
\pgfusepath{stroke}%
\end{pgfscope}%
\begin{pgfscope}%
\pgfpathrectangle{\pgfqpoint{0.791528in}{0.335648in}}{\pgfqpoint{2.402500in}{2.265000in}}%
\pgfusepath{clip}%
\pgfsetroundcap%
\pgfsetroundjoin%
\pgfsetlinewidth{2.710125pt}%
\definecolor{currentstroke}{rgb}{0.260000,0.260000,0.260000}%
\pgfsetstrokecolor{currentstroke}%
\pgfsetdash{}{0pt}%
\pgfusepath{stroke}%
\end{pgfscope}%
\begin{pgfscope}%
\pgfpathrectangle{\pgfqpoint{0.791528in}{0.335648in}}{\pgfqpoint{2.402500in}{2.265000in}}%
\pgfusepath{clip}%
\pgfsetroundcap%
\pgfsetroundjoin%
\pgfsetlinewidth{2.710125pt}%
\definecolor{currentstroke}{rgb}{0.260000,0.260000,0.260000}%
\pgfsetstrokecolor{currentstroke}%
\pgfsetdash{}{0pt}%
\pgfusepath{stroke}%
\end{pgfscope}%
\begin{pgfscope}%
\pgfpathrectangle{\pgfqpoint{0.791528in}{0.335648in}}{\pgfqpoint{2.402500in}{2.265000in}}%
\pgfusepath{clip}%
\pgfsetroundcap%
\pgfsetroundjoin%
\pgfsetlinewidth{2.710125pt}%
\definecolor{currentstroke}{rgb}{0.260000,0.260000,0.260000}%
\pgfsetstrokecolor{currentstroke}%
\pgfsetdash{}{0pt}%
\pgfusepath{stroke}%
\end{pgfscope}%
\begin{pgfscope}%
\pgfpathrectangle{\pgfqpoint{0.791528in}{0.335648in}}{\pgfqpoint{2.402500in}{2.265000in}}%
\pgfusepath{clip}%
\pgfsetroundcap%
\pgfsetroundjoin%
\pgfsetlinewidth{2.710125pt}%
\definecolor{currentstroke}{rgb}{0.260000,0.260000,0.260000}%
\pgfsetstrokecolor{currentstroke}%
\pgfsetdash{}{0pt}%
\pgfusepath{stroke}%
\end{pgfscope}%
\begin{pgfscope}%
\pgfpathrectangle{\pgfqpoint{0.791528in}{0.335648in}}{\pgfqpoint{2.402500in}{2.265000in}}%
\pgfusepath{clip}%
\pgfsetroundcap%
\pgfsetroundjoin%
\pgfsetlinewidth{2.710125pt}%
\definecolor{currentstroke}{rgb}{0.260000,0.260000,0.260000}%
\pgfsetstrokecolor{currentstroke}%
\pgfsetdash{}{0pt}%
\pgfusepath{stroke}%
\end{pgfscope}%
\begin{pgfscope}%
\pgfpathrectangle{\pgfqpoint{0.791528in}{0.335648in}}{\pgfqpoint{2.402500in}{2.265000in}}%
\pgfusepath{clip}%
\pgfsetroundcap%
\pgfsetroundjoin%
\pgfsetlinewidth{2.710125pt}%
\definecolor{currentstroke}{rgb}{0.260000,0.260000,0.260000}%
\pgfsetstrokecolor{currentstroke}%
\pgfsetdash{}{0pt}%
\pgfusepath{stroke}%
\end{pgfscope}%
\begin{pgfscope}%
\pgfpathrectangle{\pgfqpoint{0.791528in}{0.335648in}}{\pgfqpoint{2.402500in}{2.265000in}}%
\pgfusepath{clip}%
\pgfsetroundcap%
\pgfsetroundjoin%
\pgfsetlinewidth{2.710125pt}%
\definecolor{currentstroke}{rgb}{0.260000,0.260000,0.260000}%
\pgfsetstrokecolor{currentstroke}%
\pgfsetdash{}{0pt}%
\pgfusepath{stroke}%
\end{pgfscope}%
\begin{pgfscope}%
\pgfpathrectangle{\pgfqpoint{0.791528in}{0.335648in}}{\pgfqpoint{2.402500in}{2.265000in}}%
\pgfusepath{clip}%
\pgfsetroundcap%
\pgfsetroundjoin%
\pgfsetlinewidth{2.710125pt}%
\definecolor{currentstroke}{rgb}{0.260000,0.260000,0.260000}%
\pgfsetstrokecolor{currentstroke}%
\pgfsetdash{}{0pt}%
\pgfusepath{stroke}%
\end{pgfscope}%
\begin{pgfscope}%
\pgfpathrectangle{\pgfqpoint{0.791528in}{0.335648in}}{\pgfqpoint{2.402500in}{2.265000in}}%
\pgfusepath{clip}%
\pgfsetroundcap%
\pgfsetroundjoin%
\pgfsetlinewidth{2.710125pt}%
\definecolor{currentstroke}{rgb}{0.260000,0.260000,0.260000}%
\pgfsetstrokecolor{currentstroke}%
\pgfsetdash{}{0pt}%
\pgfusepath{stroke}%
\end{pgfscope}%
\begin{pgfscope}%
\pgfpathrectangle{\pgfqpoint{0.791528in}{0.335648in}}{\pgfqpoint{2.402500in}{2.265000in}}%
\pgfusepath{clip}%
\pgfsetroundcap%
\pgfsetroundjoin%
\pgfsetlinewidth{2.710125pt}%
\definecolor{currentstroke}{rgb}{0.260000,0.260000,0.260000}%
\pgfsetstrokecolor{currentstroke}%
\pgfsetdash{}{0pt}%
\pgfusepath{stroke}%
\end{pgfscope}%
\begin{pgfscope}%
\pgfpathrectangle{\pgfqpoint{0.791528in}{0.335648in}}{\pgfqpoint{2.402500in}{2.265000in}}%
\pgfusepath{clip}%
\pgfsetroundcap%
\pgfsetroundjoin%
\pgfsetlinewidth{2.710125pt}%
\definecolor{currentstroke}{rgb}{0.260000,0.260000,0.260000}%
\pgfsetstrokecolor{currentstroke}%
\pgfsetdash{}{0pt}%
\pgfusepath{stroke}%
\end{pgfscope}%
\begin{pgfscope}%
\pgfpathrectangle{\pgfqpoint{0.791528in}{0.335648in}}{\pgfqpoint{2.402500in}{2.265000in}}%
\pgfusepath{clip}%
\pgfsetroundcap%
\pgfsetroundjoin%
\pgfsetlinewidth{2.710125pt}%
\definecolor{currentstroke}{rgb}{0.260000,0.260000,0.260000}%
\pgfsetstrokecolor{currentstroke}%
\pgfsetdash{}{0pt}%
\pgfusepath{stroke}%
\end{pgfscope}%
\begin{pgfscope}%
\pgfpathrectangle{\pgfqpoint{0.791528in}{0.335648in}}{\pgfqpoint{2.402500in}{2.265000in}}%
\pgfusepath{clip}%
\pgfsetroundcap%
\pgfsetroundjoin%
\pgfsetlinewidth{2.710125pt}%
\definecolor{currentstroke}{rgb}{0.260000,0.260000,0.260000}%
\pgfsetstrokecolor{currentstroke}%
\pgfsetdash{}{0pt}%
\pgfusepath{stroke}%
\end{pgfscope}%
\begin{pgfscope}%
\pgfpathrectangle{\pgfqpoint{0.791528in}{0.335648in}}{\pgfqpoint{2.402500in}{2.265000in}}%
\pgfusepath{clip}%
\pgfsetroundcap%
\pgfsetroundjoin%
\pgfsetlinewidth{2.710125pt}%
\definecolor{currentstroke}{rgb}{0.260000,0.260000,0.260000}%
\pgfsetstrokecolor{currentstroke}%
\pgfsetdash{}{0pt}%
\pgfusepath{stroke}%
\end{pgfscope}%
\begin{pgfscope}%
\pgfpathrectangle{\pgfqpoint{0.791528in}{0.335648in}}{\pgfqpoint{2.402500in}{2.265000in}}%
\pgfusepath{clip}%
\pgfsetroundcap%
\pgfsetroundjoin%
\pgfsetlinewidth{2.710125pt}%
\definecolor{currentstroke}{rgb}{0.260000,0.260000,0.260000}%
\pgfsetstrokecolor{currentstroke}%
\pgfsetdash{}{0pt}%
\pgfusepath{stroke}%
\end{pgfscope}%
\begin{pgfscope}%
\pgfpathrectangle{\pgfqpoint{0.791528in}{0.335648in}}{\pgfqpoint{2.402500in}{2.265000in}}%
\pgfusepath{clip}%
\pgfsetroundcap%
\pgfsetroundjoin%
\pgfsetlinewidth{2.710125pt}%
\definecolor{currentstroke}{rgb}{0.260000,0.260000,0.260000}%
\pgfsetstrokecolor{currentstroke}%
\pgfsetdash{}{0pt}%
\pgfusepath{stroke}%
\end{pgfscope}%
\begin{pgfscope}%
\pgfpathrectangle{\pgfqpoint{0.791528in}{0.335648in}}{\pgfqpoint{2.402500in}{2.265000in}}%
\pgfusepath{clip}%
\pgfsetroundcap%
\pgfsetroundjoin%
\pgfsetlinewidth{2.710125pt}%
\definecolor{currentstroke}{rgb}{0.260000,0.260000,0.260000}%
\pgfsetstrokecolor{currentstroke}%
\pgfsetdash{}{0pt}%
\pgfusepath{stroke}%
\end{pgfscope}%
\begin{pgfscope}%
\pgfpathrectangle{\pgfqpoint{0.791528in}{0.335648in}}{\pgfqpoint{2.402500in}{2.265000in}}%
\pgfusepath{clip}%
\pgfsetroundcap%
\pgfsetroundjoin%
\pgfsetlinewidth{2.710125pt}%
\definecolor{currentstroke}{rgb}{0.260000,0.260000,0.260000}%
\pgfsetstrokecolor{currentstroke}%
\pgfsetdash{}{0pt}%
\pgfusepath{stroke}%
\end{pgfscope}%
\begin{pgfscope}%
\pgfpathrectangle{\pgfqpoint{0.791528in}{0.335648in}}{\pgfqpoint{2.402500in}{2.265000in}}%
\pgfusepath{clip}%
\pgfsetroundcap%
\pgfsetroundjoin%
\pgfsetlinewidth{2.710125pt}%
\definecolor{currentstroke}{rgb}{0.260000,0.260000,0.260000}%
\pgfsetstrokecolor{currentstroke}%
\pgfsetdash{}{0pt}%
\pgfusepath{stroke}%
\end{pgfscope}%
\begin{pgfscope}%
\pgfpathrectangle{\pgfqpoint{0.791528in}{0.335648in}}{\pgfqpoint{2.402500in}{2.265000in}}%
\pgfusepath{clip}%
\pgfsetroundcap%
\pgfsetroundjoin%
\pgfsetlinewidth{2.710125pt}%
\definecolor{currentstroke}{rgb}{0.260000,0.260000,0.260000}%
\pgfsetstrokecolor{currentstroke}%
\pgfsetdash{}{0pt}%
\pgfusepath{stroke}%
\end{pgfscope}%
\begin{pgfscope}%
\pgfpathrectangle{\pgfqpoint{0.791528in}{0.335648in}}{\pgfqpoint{2.402500in}{2.265000in}}%
\pgfusepath{clip}%
\pgfsetroundcap%
\pgfsetroundjoin%
\pgfsetlinewidth{2.710125pt}%
\definecolor{currentstroke}{rgb}{0.260000,0.260000,0.260000}%
\pgfsetstrokecolor{currentstroke}%
\pgfsetdash{}{0pt}%
\pgfusepath{stroke}%
\end{pgfscope}%
\begin{pgfscope}%
\pgfsetrectcap%
\pgfsetmiterjoin%
\pgfsetlinewidth{1.254687pt}%
\definecolor{currentstroke}{rgb}{0.800000,0.800000,0.800000}%
\pgfsetstrokecolor{currentstroke}%
\pgfsetdash{}{0pt}%
\pgfpathmoveto{\pgfqpoint{0.791528in}{0.335648in}}%
\pgfpathlineto{\pgfqpoint{0.791528in}{2.600648in}}%
\pgfusepath{stroke}%
\end{pgfscope}%
\begin{pgfscope}%
\pgfsetrectcap%
\pgfsetmiterjoin%
\pgfsetlinewidth{1.254687pt}%
\definecolor{currentstroke}{rgb}{0.800000,0.800000,0.800000}%
\pgfsetstrokecolor{currentstroke}%
\pgfsetdash{}{0pt}%
\pgfpathmoveto{\pgfqpoint{3.194028in}{0.335648in}}%
\pgfpathlineto{\pgfqpoint{3.194028in}{2.600648in}}%
\pgfusepath{stroke}%
\end{pgfscope}%
\begin{pgfscope}%
\pgfsetrectcap%
\pgfsetmiterjoin%
\pgfsetlinewidth{1.254687pt}%
\definecolor{currentstroke}{rgb}{0.800000,0.800000,0.800000}%
\pgfsetstrokecolor{currentstroke}%
\pgfsetdash{}{0pt}%
\pgfpathmoveto{\pgfqpoint{0.791528in}{0.335648in}}%
\pgfpathlineto{\pgfqpoint{3.194028in}{0.335648in}}%
\pgfusepath{stroke}%
\end{pgfscope}%
\begin{pgfscope}%
\pgfsetrectcap%
\pgfsetmiterjoin%
\pgfsetlinewidth{1.254687pt}%
\definecolor{currentstroke}{rgb}{0.800000,0.800000,0.800000}%
\pgfsetstrokecolor{currentstroke}%
\pgfsetdash{}{0pt}%
\pgfpathmoveto{\pgfqpoint{0.791528in}{2.600648in}}%
\pgfpathlineto{\pgfqpoint{3.194028in}{2.600648in}}%
\pgfusepath{stroke}%
\end{pgfscope}%
\end{pgfpicture}%
\makeatother%
\endgroup%